\documentclass[acmtog,nonacm]{acmart}

\usepackage{booktabs} \usepackage{multirow}
\usepackage{xcolor}
\usepackage[ruled]{algorithm2e} 
\SetAlFnt{\small}
\SetAlCapFnt{\small}
\SetAlCapNameFnt{\small}
\SetAlCapHSkip{0pt}

\begin{document}
\title{Tele360: Real-Time Feed-Forward Human Reconstruction from Sparse Unposed Cameras}

\author{Hanzhang Tu}
\orcid{0009-0003-7555-9546}
\affiliation{ \institution{Tsinghua University}
 \city{Beijing}
 \country{China}}
\email{thz22@mails.tsinghua.edu.cn}
\author{Zhanfeng Liao}
\orcid{0009-0008-3337-3993}
\affiliation{ \institution{Tsinghua University}
 \city{Beijing}
 \country{China}}
\email{luckier003@gmail.com}
\author{Wei Min}
\orcid{0009-0006-0789-4635}
\affiliation{ \institution{Shadow AI}
 \city{Beijing}
 \country{China}}
\email{minwei@yingshen-ai.com}
\author{Jiajun Zhang}
\authornote{Jiajun Zhang and Yebin Liu are co-corresponding authors.}
\orcid{0009-0002-9712-9953}
\affiliation{ \institution{Tsinghua University}
 \city{Beijing}
 \country{China}}
\email{zhangjiajun@mail.tsinghua.edu.cn}
\author{Yebin Liu}
\authornotemark[1]
\orcid{0000-0003-3215-0225}
\affiliation{ \institution{Tsinghua University}
 \city{Beijing}
 \country{China}}
\email{liuyebin@mail.tsinghua.edu.cn}

\renewcommand\shortauthors{Tu, H. et al.}

\begin{abstract}
    Live free-viewpoint visualization of real humans is critical for immersive communication and interactive digital experiences.
Existing methods either rely on computationally expensive optimization or require calibrated cameras and low-resolution inputs, making real-time high-resolution deployment impractical.
In this work, we present Tele360, the first real-time feed-forward system for dynamic human reconstruction and live free-viewpoint visualization from sparse, unposed RGB streams.
Our system jointly estimates camera poses and reconstructs a dynamic 3D Gaussian representation for each time instance in a single forward pass.
To achieve this, we start by designing a lightweight sparsity-aware multi-view transformer backbone that tokenizes foreground human regions while preserving global context through a shared scene token. 
We then employ a fully transformer-based Gaussian decoder to mitigate convolution-induced over-smoothing while keeping decoding sparse and efficient. In addition, we introduce a hybrid feature pyramid that injects multi-scale appearance cues into geometry prediction. 
We further introduce a lightweight differentiable Levenberg-Marquardt camera refinement layer to enhance multi-view consistency and geometric alignment.
Moreover, to stabilize learning under sparse, unposed inputs, we transfer multi-view geometry priors from a large visual-geometry foundation model via teacher-student distillation. 
Finally, the predicted Gaussian maps are streamed with video codecs to remote devices for interactive free-viewpoint rendering. 
Together, these designs form a balanced pipeline between efficiency and fidelity.
Extensive experiments show that Tele360 achieves state-of-the-art visual quality on studio benchmarks while supporting real-time 2K input-to-rendering at over 25 FPS on a single consumer GPU. Additional captured sequences illustrate its performance across varied subjects, clothing, and motions under our multi-camera setup.
\end{abstract}




\begin{teaserfigure}
    \centering
    \includegraphics[width=1.0\linewidth]{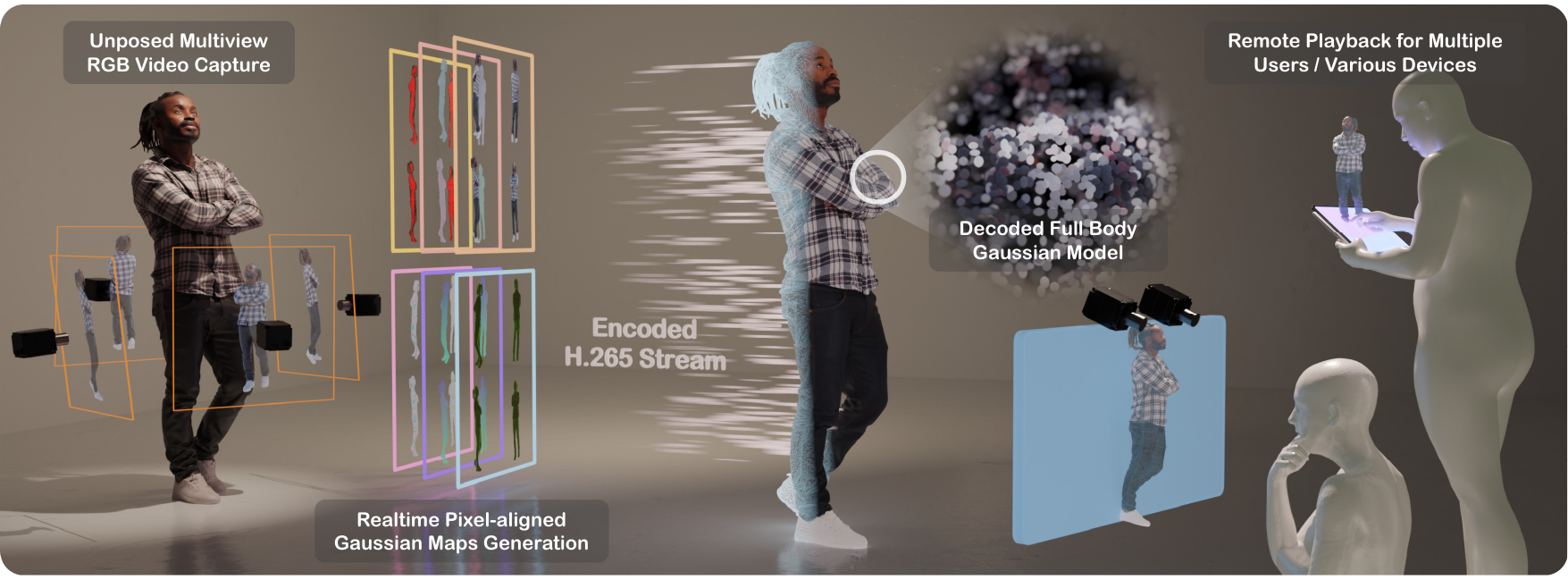}
    \caption{We present Tele360, a human telepresence system that uses only 4 to 6 unposed RGB cameras and enables real-time 3D reconstruction, streaming, and interactive viewing of dynamic humans represented by 3D Gaussians at 2K resolution across multiple endpoints.
    }
    \label{fig:teaser}
\end{teaserfigure}

\maketitle

\section{Introduction}
\label{sec:intro}
With the rapid growth of immersive communication and digital experiences, live free-viewpoint visualization of real humans is becoming a fundamental capability for applications such as AR/VR content creation~\cite{xu2024representing, wang2025freetimegs, yang2024real, li2024spacetime, liao2026sharptimegs, gao20257dgs}, telepresence~\cite{orts2016holoportation, guan2023metastream, zhou2023live4d, telealoha}, and digital avatars~\cite{hu2024gaussianavatar, xu2023avatarmav, chen2024meshavatar, chen2024monogaussianavatar, qian2024gaussianavatars, chen2025taoavatar}.
Delivering such experiences requires real-time reconstruction of dynamic humans from sparse, unposed multi-view streams.
Recent advances in efficient point-based rendering~\cite{3dgs} and visual geometry foundation models~\cite{vggt, mast3r, pi3, da3, spatialtrackerv2} have significantly advanced the field, but a deployable end-to-end system that meets these requirements has yet to be demonstrated.

Despite these advances, existing approaches still fall short of practical requirements and can be broadly categorized into two paradigms.
Optimization-based methods recover a dynamic representation via iterative fitting, 
either by training a 4D representation~\cite{yang2024real, li2024spacetime, wang2025freetimegs, liao2026sharptimegs} over long sequences or by optimizing a per-frame 3D representation~\cite{sun20243dgstream,yan2025instant,liu2024dynamics}.
While achieving high quality, they incur substantial computational cost and latency, and typically rely on known camera poses for multi-view consistency, making them unsuitable for live, interactive use.
Feed-forward methods~\cite{gps,zhou2024gps,lin2022efficient,xu2025depthsplat}, in contrast, predict a renderable representation in a single pass and are better aligned with low-latency requirements.
However, most of them assume camera poses are given when fusing sparse-view observations.
More recently, pose-agnostic variants~\cite{vggt,da3,keetha2025mapanything,anysplat,noposplat} have begun to relax this assumption by distilling or fine-tuning large visual geometry foundation models~\cite{vggt, mast3r}, yet deploying such heavy backbones in real time remains challenging, and they are restricted to relatively low input and rendering resolutions.

To address these challenges, we present Tele360, the first real-time feed-forward system for full-body dynamic human reconstruction that enables live 360$^\circ$ free-viewpoint visualization from sparse, unposed multi-view streams. Our system takes live 2K video from only 4–6 RGB cameras and runs fully online on a single consumer GPU, jointly estimating camera poses and reconstructing a dynamic 3D Gaussian representation at real-time frame rates. 
The predicted per-view Gaussian maps are encoded and streamed to a remote viewer for interactive 2K free-viewpoint rendering.

As the core of our system, we design a lightweight architecture guided by three principles: real-time efficiency, geometric fidelity, and robust generalization.
We begin by introducing a sparsity-aware multi-view transformer.
It tokenizes only the foreground human region, substantially reducing the token budget under high-resolution inputs.
As this foreground-only tokenization discards background cues and weakens the global context needed for stable camera pose estimation, we further incorporate a shared scene token that aggregates full-image information across all views. Together, these designs preserve the global structure required for reliable pose estimation while making real-time inference feasible.
Building on this compact backbone, we adopt a fully transformer-based decoder to keep decoding sparse and efficient, while also mitigating the tendency of convolutional DPT-style decoders~\cite{dpt} to oversmooth sharp depth discontinuities and introduce flying points near boundaries.
In addition, since ViT-based feature extraction is commonly performed at reduced resolution for efficiency, we introduce a lightweight CNN~\cite{edgenext} that extracts a multi-scale feature pyramid from the 2K inputs and complements the low-resolution DINO~\cite{simeoni2025dinov3} features to form hybrid features. The pyramid features are progressively injected into the decoder and the Gaussian attribute head in a coarse-to-fine manner, providing high-frequency visual cues for high-resolution prediction.
Since even small camera pose errors can lead to multi-view misalignment, duplicated surfaces, and blurred rendering, we further introduce a lightweight differentiable Levenberg-Marquardt camera self-calibration layer. It refines the camera poses using predicted geometry and cross-view pixel-aligned feature consistency in a feed-forward manner.
Finally, to improve generalization and training stability, we incorporate a teacher–student distillation scheme based on a large visual geometry foundation model~\cite{vggt}. By aligning the patch-wise similarity structure during training, our compact backbone inherits strong multi-view geometry and pose priors.

The predicted per-view Gaussian maps are pixel-aligned 2D representations, making them directly streamable with standard video codecs without additional parameterization or compression~\cite{v3,li2022streaming,wang2024videorf,dai20254d}. On the viewer side, the decoded maps can be used to instantiate 3D Gaussian primitives for each time step, enabling interactive 2K free-viewpoint rendering on devices such as tablets and autostereoscopic displays.
We evaluate our system on studio-captured benchmarks and additional sequences captured with our multi-camera setup, covering varied subjects, clothing, and motions. Our method runs fully online at over 25 FPS on a single consumer GPU, enabling real-time reconstruction, streaming, and interactive free-viewpoint 2K rendering with leading visual quality.
Contributions are summarized as follows:
\begin{itemize}
    \item We present Tele360, the first real-time feed-forward approach for full-body dynamic human reconstruction from sparse, unposed multi-view streams, enabling live 360$^\circ$ free-viewpoint visualization with 2K input-to-rendering at over 25 FPS on a single consumer GPU. 
    \item We introduce a 
        sparsity-aware multi-view transformer
    that balances efficiency and reconstruction fidelity, featuring (i) a hybrid CNN-ViT design to exploit high-resolution appearance cues, 
    (ii) 
        a reduced token budget through foreground-only attention,
    (iii) an efficient transformer-based Gaussian decoder that mitigates over-smoothing while preserving fast decoding, and (iv) an LM-based camera self-calibration layer for multi-view alignment.
    \item We build a deployable live system that integrates multi-view capture, real-time reconstruction, and codec-based streaming to support interactive remote viewing across devices such as tablets and autostereoscopic displays.
\end{itemize}

\section{Related Work}
\label{sec:relatedwork}
\paragraph{Optimization-based Dynamic Reconstruction.} 
Optimization-based pipelines remain a dominant approach for high-fidelity dynamic 3D/4D reconstruction from multi-view videos, using NeRF~\cite{mildenhall2020nerf} and, more recently, 3DGS~\cite{3dgs}.
They typically recover a sequence-specific representation by minimizing photometric and geometric inconsistencies across views and time, often optimizing a unified 4D model for an entire capture.
Earlier dynamic NeRF extensions represent scene evolution with neural scene-flow fields, as in NSFF~\cite{li2021nsff}, or with layered space-time radiance fields, as in ST-NeRF~\cite{zhang2021stnerf}.
Within this paradigm, scene dynamics are commonly modeled by explicit spatiotemporal representations~\cite{yang2024real,lee2024fully,luiten2024dynamic,xu2024representing,wang2025freetimegs,li2024spacetime,duan20244d,gao20257dgs,4k4d}, or by deformation fields (e.g., MLPs or low-rank K-planes) that drive time-varying Gaussian attributes~\cite{yang2024deformable,bae2024per,qingming2025modgs,guo2024motion,lu20243d,shaw2024swings,zhu2024motiongs,labe2024dgd,liang2025gaufre,xu2024grid4d,kim20244d}.
Related human-avatar methods parameterize 3D Gaussians on structured canonical maps: Animatable Gaussians~\cite{li2024animatable} learns pose-dependent front/back maps from a subject-specific template, while Vid2Avatar-Pro~\cite{guo2025vid2avatarpro} learns a cross-identity prior and personalizes it to each video.
These persistent avatar representations differ from our per-frame, per-view Gaussian maps, which are inferred directly from live inputs.
Several recent works further explore online streaming variants that process frames sequentially to update per-frame reconstructions~\cite{sun20243dgstream,yan2025instant,liu2024dynamics}.
Despite impressive fidelity, these methods require scene- or sequence-specific iterative optimization whose cost scales with video length and the number of views, and they typically assume calibrated or well-controlled capture, making real-time interactive deployment challenging. In contrast, our work targets real-time reconstruction from sparse, unposed streams with a feed-forward model, enabling low-latency free-viewpoint viewing.

\paragraph{Feed-Forward Dynamic Reconstruction.} 
Recent advances in feed-forward 3D reconstruction aim to replace per-scene optimization with generalizable models trained across diverse scenes.
These approaches can be broadly categorized into pose-aware and pose-free formulations.
Pose-aware methods assume known camera parameters and reconstruct geometry directly from calibrated multi-view inputs~\cite{gps,zhou2024gps,lin2022efficient,xu2025depthsplat}.
This calibrated setting is also shared by earlier image-based rendering methods, from geometry-guided Unstructured Lumigraph Rendering~\cite{buehler2001unstructured} to learned generalizable renderers such as IBRNet~\cite{wang2021ibrnet} and HumanNeRF~\cite{zhao2022humannerf}; these approaches primarily target view synthesis rather than explicit, streamable 3D reconstruction.
Despite their efficiency, pose-aware reconstruction methods rely on accurate calibration, limiting their applicability in unconstrained capture settings.
Pose-free methods instead jointly infer geometry and camera poses from uncalibrated images via end-to-end learning~\cite{vggt,wang2024dust3r,mast3r,da3,keetha2025mapanything,anysplat,noposplat}, improving usability in casual multi-view setups.
DUSt3R~\cite{wang2024dust3r} and MASt3R~\cite{mast3r} leverage Transformers to directly predict inter-view point maps, enabling feed-forward estimation of depth and relative pose. 
VGGT~\cite{vggt}, $\pi^3$~\cite{pi3}, and Depth Anything~3~\cite{da3} stack cascaded Transformer blocks to jointly infer camera poses, point trajectories, and scene geometry in a single forward pass, substantially improving both accuracy and efficiency. 
However, inference cost typically increases with the number of input views and image resolution, making real-time high-resolution deployment challenging.
Moreover, these models are primarily designed for static scene reconstruction and may exhibit cross-view inconsistencies or reduced texture fidelity when they are applied to dynamic, human-centric scenarios.
Recent work HiReFF~\cite{jiang2026hireff} applies pose-free feed-forward reconstruction to dynamic humans captured from sparse, uncalibrated video streams, while remaining below real-time throughput.

\paragraph{Streamable Volumetric Video.} 
Recent research on volumetric video streaming focuses on efficient capture, transmission, and playback across heterogeneous platforms.
End-to-end systems such as Holoportation~\cite{orts2016holoportation}, MetaStream~\cite{guan2023metastream}, Live4D~\cite{zhou2023live4d}, and Tele-Aloha~\cite{telealoha} integrate live capture, delivery, and interactive viewing under different sensing and calibration assumptions.
NeRF-based approaches~\cite{li2022streaming,wang2023neural2,wang2024videorf,wu2024tetrirf} compress neural radiance representations using video codecs or structured decompositions to enable real-time decoding and rendering.
Similarly, 3DGS-based methods~\cite{v3,dai20254d} adopt Gaussian representations and employ quantization or codec-based compression for streamable playback. 
However, the NeRF- and 3DGS-based compression pipelines above typically assume that volumetric assets have been reconstructed offline through scene-specific optimization, and thus decouple reconstruction from transmission.
Such designs are not suitable for live capture scenarios where reconstruction, encoding, and rendering must operate jointly under strict latency constraints.
In contrast, our method jointly supports real-time reconstruction, transmission, and playback of volumetric video. 

\paragraph{Volumetric Fusion for Dynamic 3D Reconstruction.} 
Volumetric fusion is a classical direction for template-free dynamic 3D reconstruction. 
DynamicFusion~\cite{newcombe2015dynamicfusion} pioneered real-time, template-free non-rigid reconstruction from a single RGB-D sensor by estimating a volumetric warp to a canonical model. 
Subsequent works improve robustness by incorporating strong regularization and motion priors such as physical constraints~\cite{slavcheva2017killingfusion, slavcheva2018sobolevfusion}, skeleton cues~\cite{yu2017bodyfusion}, parametric body models~\cite{yu2018doublefusion}, and learned correspondences~\cite{bozic2020deepdeform}, but remain vulnerable to occlusions and drift in invisible regions. 
To overcome this, Fusion4D~\cite{dou2016fusion4d} and Motion2Fusion~\cite{dou2017motion2fusion} extend non-rigid fusion to real-time multi-view RGB-D rigs, while Function4D~\cite{yu2021function4d} targets very sparse consumer RGB-D inputs by combining sliding-window fusion with implicit surface refinement.
Beyond template-free fusion, Drivable Avatar Clothing~\cite{xiang2023drivable} targets full-body telepresence by driving a preconstructed subject-specific avatar from sparse RGB-D observations and the motion of the body and face.
These systems, however, rely on depth sensors, calibrated capture, or preconstructed avatars, whereas our method operates on sparse, unposed RGB streams without subject-specific optimization.

\section{Method}
\label{sec:method}

\begin{figure*}[ht]
    \centering
    \includegraphics[width=1.0\linewidth]{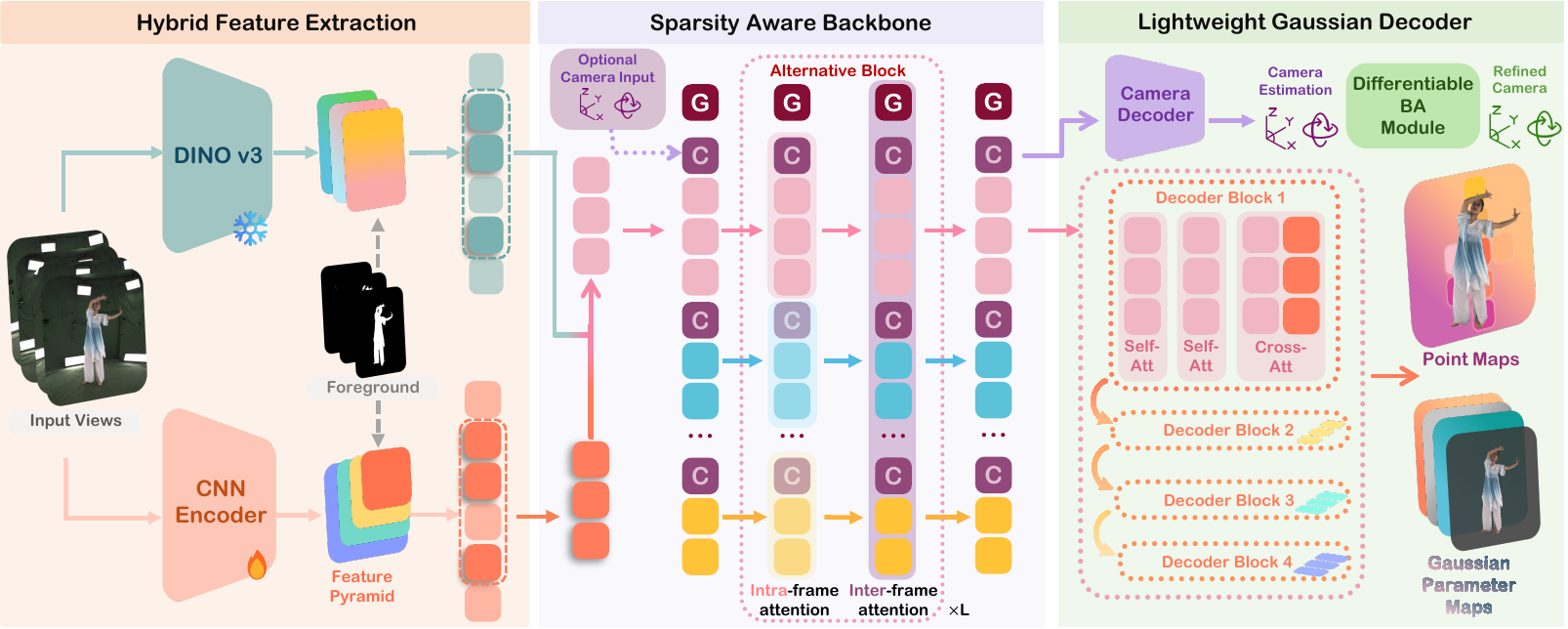}
        \caption{Pipeline overview. Given 4 to 6 uncalibrated RGB videos, we extract features with a frozen DINO~\cite{simeoni2025dinov3} and a lightweight CNN encoder, and predict a foreground mask to select foreground tokens. 
        The hybrid features are fed into our sparsity-aware backbone and lightweight decoder to predict camera parameters, depth maps, and Gaussian parameter maps. After the BA module, the final point maps used for Gaussian positions are obtained from the predicted depth maps and the refined camera poses. When calibrated parameters are available, the camera decoder and BA are bypassed. }
    \label{fig:pipeline}
    \end{figure*}
In this section, we present our feed-forward framework for real-time dynamic human reconstruction from sparse, unposed multi-view streams.
We first define the problem setting and the output representation in Sec.~\ref{subsec:task definition}. We then detail the proposed architecture and its key design choices in Sec.~\ref{subsec: architecture}. Finally, we describe the training setup and learning strategy in Sec.~\ref{subsec: training}. An overview of our pipeline is shown in Fig.~\ref{fig:pipeline}.

\subsection{Task Definition}
\label{subsec:task definition}
We consider real-time dynamic human reconstruction from sparse, unposed multi-view video streams. At each time step $t$, the input arrives as a set of synchronized 2K RGB frames $\mathcal{I}_t=\{I_t^{(i)}\}_{i=1}^{N}$ with $I_t^{(i)}\in\mathbb{R}^{H\times W\times 3}$, where $N$ denotes the number of input views (variable across setups, typically $N \in [4,6]$). Our goal is to learn a feed-forward network $f_\theta$ that maps the multi-view input at each time step to a renderable reconstruction and the corresponding camera parameters: $(\Pi_t,\mathcal{M}_t)=f_\theta(\mathcal{I}_t)$, where $\Pi_t=\{\Pi_t^{(i)}\}_{i=1}^{N}$ denotes the predicted per-view camera parameters and $\mathcal{M}_t=\{\mathcal{M}_t^{(i)}\}_{i=1}^{N}$ denotes the predicted per-view Gaussian maps. The model operates online and processes each $\mathcal{I}_t$ independently.
Although the physical cameras in our rig are nominally static, Tele360 predicts camera parameters for every frame. This design makes calibration optional, accommodates small changes caused by camera drift or vibration during live capture, and adds less than 0.5 ms of computation under the six-view setting.

We use compact parameterizations for the outputs. 
For each view $i$, we parameterize the camera as
$\Pi_t^{(i)}=\{q_t^{(i)},t_t^{(i)},f_t^{(i)}\}\in\mathbb{R}^{9}$,
where $q_t^{(i)}\in\mathbb{R}^{4}$ is a rotation quaternion, $t_t^{(i)}\in\mathbb{R}^{3}$ is a translation vector, and $f_t^{(i)}\in\mathbb{R}^{2}$ denotes FOV.
Each per-view Gaussian map $\mathcal{M}_t^{(i)}$ consists of a point map and an attribute map,
$\mathcal{M}_t^{(i)}=\big(P_t^{(i)}, A_t^{(i)}\big)$,
where $P_t^{(i)}\in\mathbb{R}^{h\times w\times 3}$ provides the 3D position and
$A_t^{(i)}\in\mathbb{R}^{h\times w\times C_a}$ stores Gaussian attributes (e.g., opacity, scale, rotation, and appearance).
In our implementation, the Gaussian maps are predicted at 1K resolution ($h \times w$).
Since the representation is pixel-aligned, each pixel corresponds to one anisotropic 3D Gaussian, and aggregating
$\{\mathcal{M}_t^{(i)}\}_{i=1}^{N}$ across views directly instantiates a set of 3D Gaussian primitives $\mathcal{G}_t$ for rendering.

\subsection{Architecture}
\label{subsec: architecture}
We now detail the architecture of Tele360. Our design balances two practical requirements: preserving fine-grained appearance from 2K inputs and enabling real-time inference with a lightweight model. Accordingly, the architecture comprises four key components: (1) hybrid feature extraction for multi-scale image features, (2) a sparsity-aware multi-view backbone for efficient cross-view reasoning under unposed capture, (3) a lightweight Gaussian decoder that predicts pixel-aligned Gaussian maps for rendering, and (4) an LM-based camera self-calibration layer for multi-view alignment.
\subsubsection{Hybrid Feature Extraction.}
 Recent visual-geometry foundation models build on ViT~\cite{vit} representations for multi-view reasoning. VGGT~\cite{vggt} and MASt3R~\cite{mast3r} are representative examples. 
  They often leverage self-supervised DINO features~\cite{simeoni2025dinov3}. However, both the computational cost and memory requirements of ViTs grow rapidly as input resolution increases. These models therefore process smaller inputs, typically $512 \times 512$. 
  Our human-centric setting faces the same constraint: naively increasing the ViT input resolution would compromise real-time inference.

 To exploit high-resolution visual cues without incurring excessive computation, we adopt a hybrid ViT–CNN design as shown in Fig.~\ref{fig:pipeline}. Specifically, we keep the DINO feature branch at a lower input resolution to keep the transformer token budget manageable, and introduce a lightweight yet effective CNN~\cite{edgenext} to process the 2K inputs and produce a multi-scale feature pyramid that captures fine-grained appearance details. 
 Let $I_{l}^{(i)}$ and $I_{h}^{(i)}$ denote the low-resolution (512) and high-resolution (2K) versions of view $i$, respectively. 
 We extract a ViT feature map using a frozen DINO encoder $\mathcal{E}_v$ and a multi-scale feature pyramid using a CNN encoder $\mathcal{E}_c$:
 \begin{equation}
\begin{aligned}
 F^{(i)}_v = \mathcal{E}_\text{v}(I_{l}^{(i)}), \{F^{(i)}_{c,s}\}_{s=1}^{S} = \mathcal{E}_\text{c}(I_h^{(i)}).
 \end{aligned}
\end{equation}
where $s$ indexes pyramid levels. We then fuse the ViT features with the coarsest pyramid level $F^{(i)}_{c,S}$ by first tokenizing the CNN feature map and aligning channel dimensions with an MLP:
\begin{equation}
    H^{(i)}=F_v^{(i)}+\mathrm{MLP}(\mathrm{Tok}(F_{c,S}^{(i)}))
\end{equation}
The resulting hybrid features are fed into the subsequent sparsity-aware transformer backbone for cross-view reasoning.
Note that the multi-scale pyramid features are also used in the Gaussian decoder and will be discussed later.

\subsubsection{Sparsity-Aware Backbone.}
In human-centric captures, informative content is largely concentrated in the foreground, so applying dense ViT attention over full images wastes computation on background regions and hinders real-time deployment. We therefore adopt a sparsity-aware multi-view transformer backbone that restricts attention to foreground tokens while preserving the global context needed for stable camera estimation via a shared scene token. Here, ``sparse'' refers to token sparsity, meaning that we attend only to tokens whose spatial locations fall inside the foreground mask rather than a dense ViT token grid.

Concretely, given the hybrid per-view features $H^{(i)}$ and the foreground mask $S^{(i)}$ predicted by Robust Video Matting~\cite{lin2021robust}, we select only the foreground tokens to form a sparse token set $\tilde{H}^{(i)}$. We augment each view with a learnable camera token $t^{(i)}_c$. The intra-frame token sequence is constructed as
\begin{equation}
    f^{(i)}_{\operatorname{intra-frame}}=[t^{(i)}_c,\ \tilde{H}^{(i)}],
\end{equation}
on which we apply standard self-attention for within-view reasoning. We then perform inter-frame fusion by concatenating tokens from all views, 
\begin{equation}
    f_{\operatorname{inter-frame}}=[t_s, t_c^{(1)},\tilde{H}^{(1)},\ldots,t_c^{(N)},\tilde{H}^{(N)}],
\end{equation}
and applying self-attention to enable multi-view interaction under a compact token budget. Here $t_s$ is a learnable scene token shared across all views that aggregates global context during inter-frame attention. It preserves global context without adding background tokens, helping stabilize camera pose estimation.

\subsubsection{Lightweight Gaussian Parameter Decoder.}
To keep the decoding stage efficient, we preserve token-level sparsity and avoid dense convolutional upsampling used in DPT~\cite{dpt} decoders. Instead, we adopt a lightweight transformer decoder that refines sparse queries and predicts pixel-aligned Gaussian maps. This design also mitigates convolution-induced over-smoothing and reduces flying points near sharp boundaries.

For each view $i$, we initialize a set of sparse decoder queries on the foreground region by combining the corresponding image patch content, UV positional encoding, and the per-view camera token $t_c^{(i)}$. Specifically, given the foreground patch indices $\Omega^{(i)}$ selected by the predicted mask on the high-resolution input $I_h^{(i)}$, we construct
\begin{equation}
Q^{(i)}=\mathrm{Extract}(I_h^{(i)};\Omega^{(i)})+\mathrm{UV}(\Omega^{(i)})+(t_c^{(i)}),
\qquad Q^{(i)}\in\mathbb{R}^{p\times d},
\end{equation}
where $p=|\Omega^{(i)}|$ is the number of foreground patches and $d$ is the token dimension. Here, $\mathrm{Extract}(I_h^{(i)};\Omega^{(i)})\in\mathbb{R}^{p\times d}$ extracts image content for the selected foreground regions, and $\mathrm{UV}(\Omega^{(i)})\in\mathbb{R}^{p\times d}$ encodes the normalized pixel coordinates.

Starting from $Q^{(i)}$, the decoder alternates two types of attention blocks. The first is self-attention (SA), where the foreground queries interact with each other to propagate intra-frame information. The second is cross-attention (CA), where the queries attend to multi-scale conditioning features from earlier stages of the network. Concretely, at each refinement stage $s$, we form the key/value tokens by fusing an intermediate transformer backbone feature $F^{(i)}_{\text{back},s}$ with a CNN feature-pyramid level $F^{(i)}_{\text{c},s}$, and use the fused tokens as $K,V$ in cross-attention:
\begin{equation}
K_s^{(i)},V_s^{(i)} \leftarrow F^{(i)}_{\text{back},s}+F^{(i)}_{\text{c},s}.
\end{equation}
We apply these CA updates in a coarse-to-fine schedule: backbone features are used from deeper to shallower layers, while CNN pyramid features are fused from shallower to deeper levels, so that the decoder gradually combines high-level geometric cues with fine-grained appearance details. In our implementation, we repeat the SA--SA--CA pattern for $S$ stages, each conditioned on the corresponding multi-scale features. 
After refinement, the updated queries are scattered back to their corresponding locations in a structured 2D feature map, which is fed into separate lightweight heads to predict the per-view depth map and Gaussian attribute maps.
The BA module then refines the camera poses, and the final point maps used for Gaussian positions are obtained from the predicted depth maps and the refined camera poses.

\subsubsection{Differentiable Camera BA Layer}
\label{sec:diff_camera_ba}

\begin{figure}[t!]
    \centering
    \includegraphics[width=1.0\linewidth]{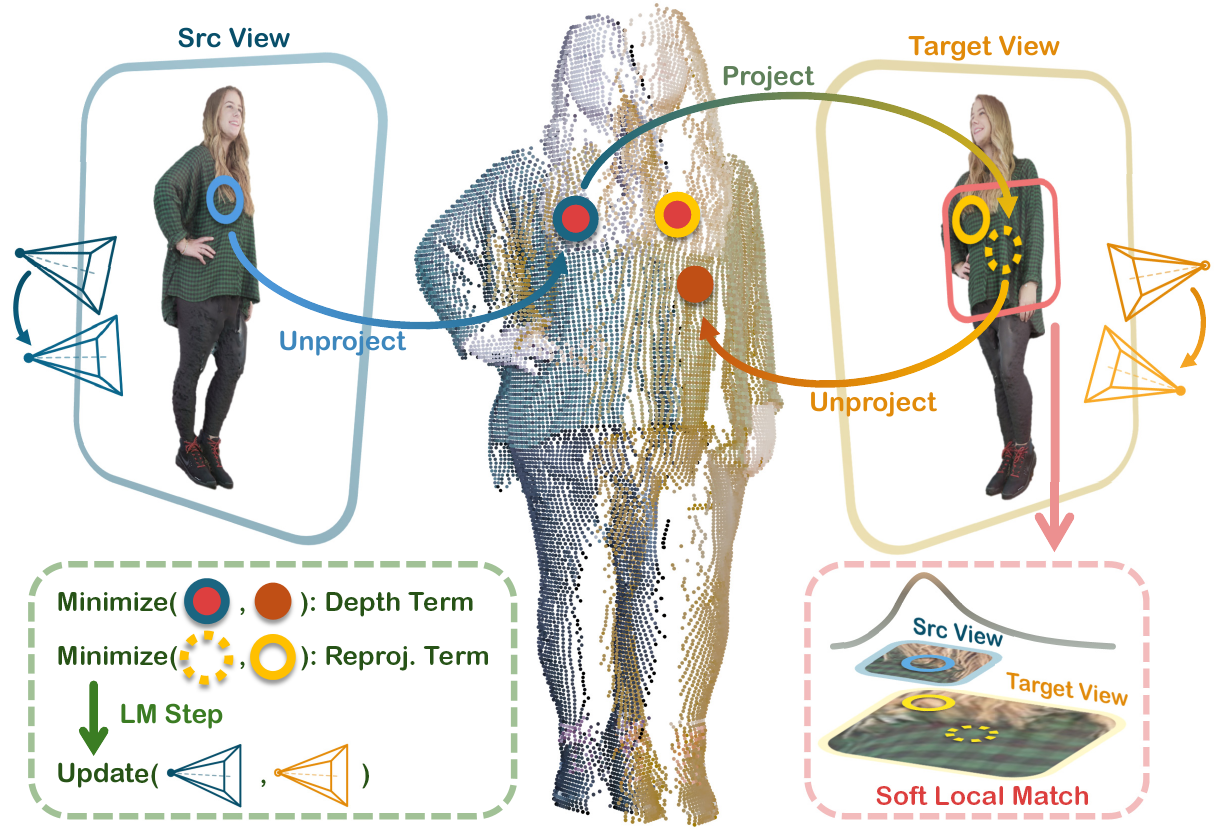}
    \caption{Differentiable Camera BA Layer. Given source depth and tentative cross-view correspondences, our BA layer jointly minimizes reprojection and depth-consistency residuals via LM, updating the camera poses to improve multi-view geometric alignment.}
    \label{fig:diff-ba-method}
    \end{figure}

The modules described above predict pixel-aligned Gaussian maps together with feed-forward camera parameters. 
Although such VGGT-style camera prediction provides a strong initialization, it is not explicitly optimized for the multi-view geometry of the current input instance. 
In our setting, even small pose errors may lead to misaligned per-view Gaussian maps, duplicated surfaces, and blurred novel-view rendering. 
To improve geometric consistency while preserving the feed-forward nature of the system, we append a lightweight \emph{camera-only differentiable bundle adjustment (BA)} layer after the main model.

Fig.~\ref{fig:diff-ba-method} illustrates the differentiable camera BA layer. 
Let $\mathbf{T}_i^0 \in SE(3)$ denote the initial camera-to-world pose predicted for view $i$, $\mathbf{A}_i$ the intrinsic matrix, $D_i$ the predicted depth map, and $F_i$ the dense feature map. 
We refine only the camera poses $\{\mathbf{T}_i\}_{i=1}^{N}$, while keeping the predicted depths/features as differentiable measurements. 
Let $\mathcal{V}_i$ denote the valid pixel domain of source view $i$, and let 
$\mathcal{U}_i \subseteq \mathcal{V}_i$ be a sparse anchor-pixel set sampled from this domain. 
For each anchor pixel $\mathbf{u}\in\mathcal{U}_i$, we back-project the point and transform it to a target view $j$:
\begin{equation}
\begin{aligned}
    \mathbf{x}_i(\mathbf{u}) &=
    D_i(\mathbf{u}) \mathbf{A}_i^{-1}\bar{\mathbf{u}}, \\
    \mathbf{x}_{ij}(\mathbf{u}) &=
    \mathbf{T}_j^{-1}\mathbf{T}_i \mathbf{x}_i(\mathbf{u}),
\end{aligned}
\end{equation}
where $\bar{\mathbf{u}}$ is the homogeneous image coordinate. The projected target location and its depth are given by:
\begin{equation}
\begin{aligned}
    \hat{\mathbf{u}}_{ij} &= \pi(\mathbf{A}_j\mathbf{x}_{ij}), \\
    \hat{z}_{ij} &= [\mathbf{x}_{ij}]_z .
\end{aligned}
\end{equation}

Since ground-truth correspondences are unavailable at test time, we obtain target observations through differentiable local matching. 
Around $\hat{\mathbf{u}}_{ij}$, we construct a small candidate window $\mathcal{N}_r(\hat{\mathbf{u}}_{ij})$ and compute a soft correspondence using feature similarity:
\begin{equation}
\begin{aligned}
    p_{ij}(\mathbf{v}\mid\mathbf{u}) &=
    \operatorname*{softmax}_{\mathbf{v}\in\mathcal{N}_r}
    \left(
    \frac{\langle \bar{F}_i(\mathbf{u}), \bar{F}_j(\mathbf{v}) \rangle}{\tau}
    \right), \\
    \tilde{\mathbf{u}}_{ij}(\mathbf{u}) &=
    \sum_{\mathbf{v}\in\mathcal{N}_r}
    p_{ij}(\mathbf{v}\mid\mathbf{u})\,\mathbf{v},
\end{aligned}
\end{equation}
where $\bar{F}$ denotes $\ell_2$-normalized features composed of the DINO feature $F_v$ and the CNN feature $F_{c,2}$, and $\tau$ is the matching temperature. This soft-argmax formulation keeps the correspondence estimation differentiable and provides sub-pixel target measurements for BA.

We then define a pose-only BA objective over directed view pairs $\mathcal{P} = \{(i,j)\in[N]\times[N]\mid i\neq j\}$, where $[N]=\{1,\dots,N\}$ denotes the set of view indices.
For each matched anchor, the residual contains a reprojection term and a depth-consistency term:
\begin{equation}
\begin{aligned}
    \mathbf{r}^{u}_{ij} &=
    \pi(\mathbf{A}_j\mathbf{x}_{ij}) - \tilde{\mathbf{u}}_{ij}, \\
    r^{z}_{ij} &= \frac{\hat{z}_{ij} - D_j(\tilde{\mathbf{u}}_{ij})}{\sigma_z}.
\end{aligned}
\end{equation}
The camera refinement is formulated as follows:
\begin{equation}
\begin{aligned}
\{\mathbf{T}_i^\star\}_{i=1}^N
= \operatorname*{argmin}_{\{\mathbf{T}_i\}}
&\sum_{(i,j)\in\mathcal{P}}
    \sum_{\mathbf{u}\in\mathcal{U}_i}
    \omega_{ij}(\mathbf{u})\,
    \rho\left(
        \|\mathbf{r}^{u}_{ij}(\mathbf{u})\|_2^2
        + \lambda_z |r^{z}_{ij}(\mathbf{u})|^2
    \right)
    \\
&+ \lambda_p
    \sum_i
    \left\|
        \operatorname{Log}_{\mathrm{SE}(3)}\!\left(
        \mathbf{T}_i(\mathbf{T}_i^0)^{-1}
        \right)^\vee
    \right\|_2^2 .
\end{aligned}
\end{equation}
Here, $\operatorname{Log}_{\mathrm{SE}(3)}(\cdot)$ is the logarithm map on the Lie group $SE(3)$. 
It maps the relative pose $\mathbf{T}_i(\mathbf{T}_i^0)^{-1}$ to a twist in the Lie algebra $\mathfrak{se}(3)$, and $(\cdot)^\vee$ converts it to a 6D vector consisting of rotational and translational components. 
$\omega_{ij}$ compactly absorbs matching confidence, local uncertainty, visibility/depth validity, and robust filtering, while the weak pose prior keeps the refined solution in the gauge of the feed-forward initialization.
$\rho(\cdot)$ is a robust penalty function applied to the combined residual energy, which downweights unreliable correspondences, occlusions, and depth outliers. 

We solve the above objective with unrolled Levenberg--Marquardt (LM) iterations. In all experiments, we sample $|\mathcal{U}_i|=2048$ anchor pixels per view, use a local matching-window radius of $r=3$, and unroll four LM iterations.
At each iteration, analytic Jacobian blocks are computed with respect to the right-multiplicative updates to the source and target camera poses, and sparse normal equations are accumulated over the view graph. 
We define the pose-prior residual as follows:
\begin{equation}
\begin{aligned}
\mathbf{e}_p &=
\left[
(\mathbf{e}_1^{p})^\top,\ldots,(\mathbf{e}_N^{p})^\top
\right]^\top,\\
\mathbf{e}_i^{p} &= \operatorname{Log}_{\mathrm{SE}(3)}
\left(
\mathbf{T}_i(\mathbf{T}_i^0)^{-1}
\right)^\vee
\in \mathbb{R}^6.
\end{aligned}
\end{equation}

Using a first-order identity approximation for the pose-prior Jacobian, the normal equations are given by:
\begin{equation}
\begin{aligned}
    \mathbf{H} &= \mathbf{J}^{\top}\mathbf{W}\mathbf{J} + \lambda_p\mathbf{I}, \\
    \mathbf{b} &= \mathbf{J}^{\top}\mathbf{W}\mathbf{r} + \lambda_p\mathbf{e}_p .
\end{aligned}
\end{equation}
The Jacobian $\mathbf{J}$ is the stacked Jacobian of all reprojection and depth residuals with respect to the stacked local pose increments $\boldsymbol{\delta}=[\boldsymbol{\delta}_1^\top,\ldots,\boldsymbol{\delta}_N^\top]^\top$. 
The stacked pose increment
$\boldsymbol{\delta}$
is obtained by the damped linear solve:
\begin{equation}
    \left(
    \mathbf{H}
    + \lambda_{\mathrm{lm}}\mathbf{H}_{\mathrm{diag}}
    \right)
    \boldsymbol{\delta}
    =
    -\mathbf{b},
\end{equation}
where $\mathbf{H}_{\mathrm{diag}}$ denotes the diagonal part of $\mathbf{H}$. Each camera pose is then updated using a right perturbation:
\begin{equation}
    \mathbf{T}_i \leftarrow
    \mathbf{T}_i \operatorname{Exp}_{\mathrm{SE}(3)}(\boldsymbol{\delta}_i),
\end{equation}
where $\operatorname{Exp}_{\mathrm{SE}(3)}(\cdot)$ is the Lie-group exponential map from $\mathfrak{se}(3)$ to $SE(3)$.
For each residual from pair $(i,j)$, only the Jacobian blocks with respect to $\boldsymbol{\delta}_i$ and $\boldsymbol{\delta}_j$ are non-zero.

Since only camera poses are optimized, the system dimension is $6N$, making the layer lightweight for sparse multi-view inputs.
This layer bridges feed-forward reconstruction and instance-specific test-time optimization. 
During inference, the network parameters remain fixed throughout the camera-refinement iterations, while the BA layer refines the camera poses using only the predicted depths, features, and multi-view consistency of the current frames. 
During training, all components including feature sampling, soft matching, residual construction, sparse normal equation accumulation, and the linear solve are differentiable. 
Therefore, the BA layer can be kept inside the computation graph and trained end-to-end with the main model, allowing the camera predictor and geometry/feature heads to learn outputs that are not only accurate in a feed-forward sense, but also well-conditioned for subsequent geometric refinement.

\subsection{Training} 
\label{subsec: training}
\subsubsection{Teacher-Student Distillation.}
To transfer strong priors from a large multi-view geometry model~\cite{vggt} into our lightweight model and improve generalization, we adopt a teacher-student distillation scheme. However, naive direct feature alignment can be overly restrictive, since our student differs in architecture and resolution, and it is specialized for human-centric reconstruction. Instead, we distill a relational prior by aligning the patch-wise similarity structure within each image.

Concretely, we extract $p$ patch features of dimension $d$ from an image and form $L_2$-normalized feature matrices $X_s, X_{\text{vggt}}\in\mathbb{R}^{p\times d}$ from our student and the frozen VGGT teacher, respectively. We then align their patch--patch similarity structures using a Gram loss~\cite{gramloss}:
\begin{equation}
    \mathcal{L}_{\text{Gram}}=\left\|X_sX_s^\top - X_{\text{vggt}}X_{\text{vggt}}^\top\right\|_F^2.
\end{equation}
By aligning this relational structure rather than raw features, the student inherits the teacher's patch-level consistency while retaining the flexibility to specialize in human-centric reconstruction.

\subsubsection{BA-aware Geometric Supervision.}
We supervise the differentiable camera BA layer by applying losses to the intermediate camera poses produced by the unrolled optimization. Let
$\tilde{\mathbf{T}}^{(\ell,i)}$ denote the refined pose of view $i$ after the
$\ell$-th BA iteration, and $\mathbf{T}^{(i)}$ denote the normalized ground-truth
pose. Taking the first view as the reference, we define:
\begin{equation}
    \small
    \mathcal{L}_{\text{BA}}
    =
    \sum_{\ell=1}^{L}\alpha_\ell
    \frac{1}{N-1}
    \sum_{i=2}^{N}
    \left\|
    \operatorname{Log}_{\mathrm{SE}(3)}
    \left(
    \tilde{\mathbf{T}}^{(\ell,i)}
    (\mathbf{T}^{(i)})^{-1}
    \right)^\vee
    \right\|_1 .
\end{equation}
This encourages each BA step to progressively correct the feed-forward camera prediction.

We also apply a geometry-guided feature consistency loss to make the features suitable for local matching. Using the ground-truth depth and cameras, each valid source pixel $\mathbf{u}$ in view $i$ is projected to its corresponding location $\mathbf{v}^{(i,j)}$ in view $j$. 
Around this location, we construct a local window $\{\mathbf{v}^{(i,j)}+\boldsymbol{\delta}^\mathbf{v}_m\}_{m=1}^{M}$ and compute matching probabilities by cosine similarity:
\begin{equation}
\small
p_m =
\operatorname{softmax}_m
\left(
\frac{
\langle
\bar{F}^{(i)}(\mathbf{u}),
\bar{F}^{(j)}(\mathbf{v}^{(i,j)}+\boldsymbol{\delta}^\mathbf{v}_m)
\rangle
}{\tau}
\right).
\end{equation}
We supervise it with a Gaussian soft target centered at the projected correspondence:
\begin{equation}
    \begin{aligned}
        \small
        y_m &=
        \operatorname{softmax}_m
        \left(
        -\frac{\|\boldsymbol{\delta}^\mathbf{v}_m\|_2^2}{2\sigma^2}
        \right), \\
        \small
        \mathcal{L}_{\text{GFC}}
        &=
        -\frac{1}{Z}
        \sum_{i,j,\mathbf{u}}
        w^{(i,j)}(\mathbf{u})
        \sum_{m=1}^{M}
        y_m \log p_m .
    \end{aligned}
\end{equation}
Here $w^{(i,j)}$ denotes the validity weight derived from mask, visibility and depth-consistency checks. The final BA-related objective is
\begin{equation}
    \small
    \mathcal{L}_{\text{BA-train}}
    =
    \lambda_{\text{BA}}\mathcal{L}_{\text{BA}}
    +
    \lambda_{\text{GFC}}\mathcal{L}_{\text{GFC}} .
\end{equation}
These losses jointly supervise the unrolled camera refinement and encourage
locally discriminative, cross-view consistent features.

\subsubsection{Training Objectives.}
We train the model with a combination of geometric supervision, camera supervision, rendering loss, BA-aware loss, and Gram-based distillation:
\begin{equation}
\begin{gathered}
\mathcal{L} = \lambda_{\mathrm{geo}}\mathcal{L}_{\text{geo}}
+ \lambda_{\mathrm{cam}}\mathcal{L}_{\text{cam}}
+ \lambda_{\mathrm{render}}\mathcal{L}_{\text{render}} \\
+ \mathcal{L}_{\text{BA-train}}
+ \lambda_{\mathrm{Gram}}\mathcal{L}_{\text{Gram}}, \\
\mathcal{L}_{\text{geo}} = \sum_{i=1}^{N}\Big(
\left\|C^{(i)}\odot(\tilde{P}^{(i)}-P^{(i)})\right\|_2
+ \\ 
\left\|C^{(i)}\odot(\nabla\tilde{P}^{(i)}-\nabla P^{(i)})\right\|_1
-\alpha \log C^{(i)}
\Big),\\
\mathcal{L}_{\text{cam}} = \sum_{i=1}^{N}\left\|\tilde{\mathbf{\Pi}}^{(i)}-\mathbf{\Pi}^{(i)}\right\|_1,\\
\mathcal{L}_{\text{render}} = \sum_{i=1}^{N}\Big(
\lambda_{i}\left\|\tilde{I}^{(i)}-I_h^{(i)}\right\|_1
+\lambda_{s}\mathcal{L}_s(\tilde{I}^{(i)}, I_h^{(i)})
+\lambda_{p}\mathcal{L}_p(\tilde{I}^{(i)}, I_h^{(i)})
\Big).
\end{gathered}
\end{equation}
Following the formulations in Secs.~\ref{subsec:task definition}--~\ref{subsec: architecture}, $\tilde{\cdot}$ denotes predictions and $\lambda$ denotes loss weights. $P^{(i)}$ is the point map of the $i$-th view, $\mathbf{\Pi}^{(i)}$ denotes the camera parameters and $I^{(i)}$ denotes the rendered high-resolution image for view $i$. $C^{(i)}$ is the per-pixel confidence map, $\nabla$ denotes the spatial gradient, $\alpha$ weights its regularization term, and $\mathcal{L}_s$ and $\mathcal{L}_p$ are SSIM loss~\cite{wang2004image} and perceptual loss~\cite{zhang2018the}, respectively.
We set the loss weights to $\lambda_{\mathrm{geo}}=2.0$, $\lambda_{\mathrm{cam}}=20.0$, $\lambda_{\mathrm{render}}=1.0$, $\lambda_{\mathrm{BA}}=1.0$, $\lambda_{\mathrm{GFC}}=0.01$, and $\lambda_{\mathrm{Gram}}=0.1$.

\subsubsection{Ground Truth Coordinate Normalization.}
Since 3D reconstruction is ambiguous up to a global similarity transform, we normalize the supervision into a canonical coordinate system. We first transform all camera parameters and 3D points into the coordinate frame of the first input camera so that its pose becomes identity. We then scale all camera translations (including those of the input and novel-view cameras) and the point map $P$ by the mean distance of the input camera centers to the origin.

\section{Real-Time Streaming System}
\label{sec:system}
\begin{figure}
    \centering
    \includegraphics[width=1.0\linewidth]{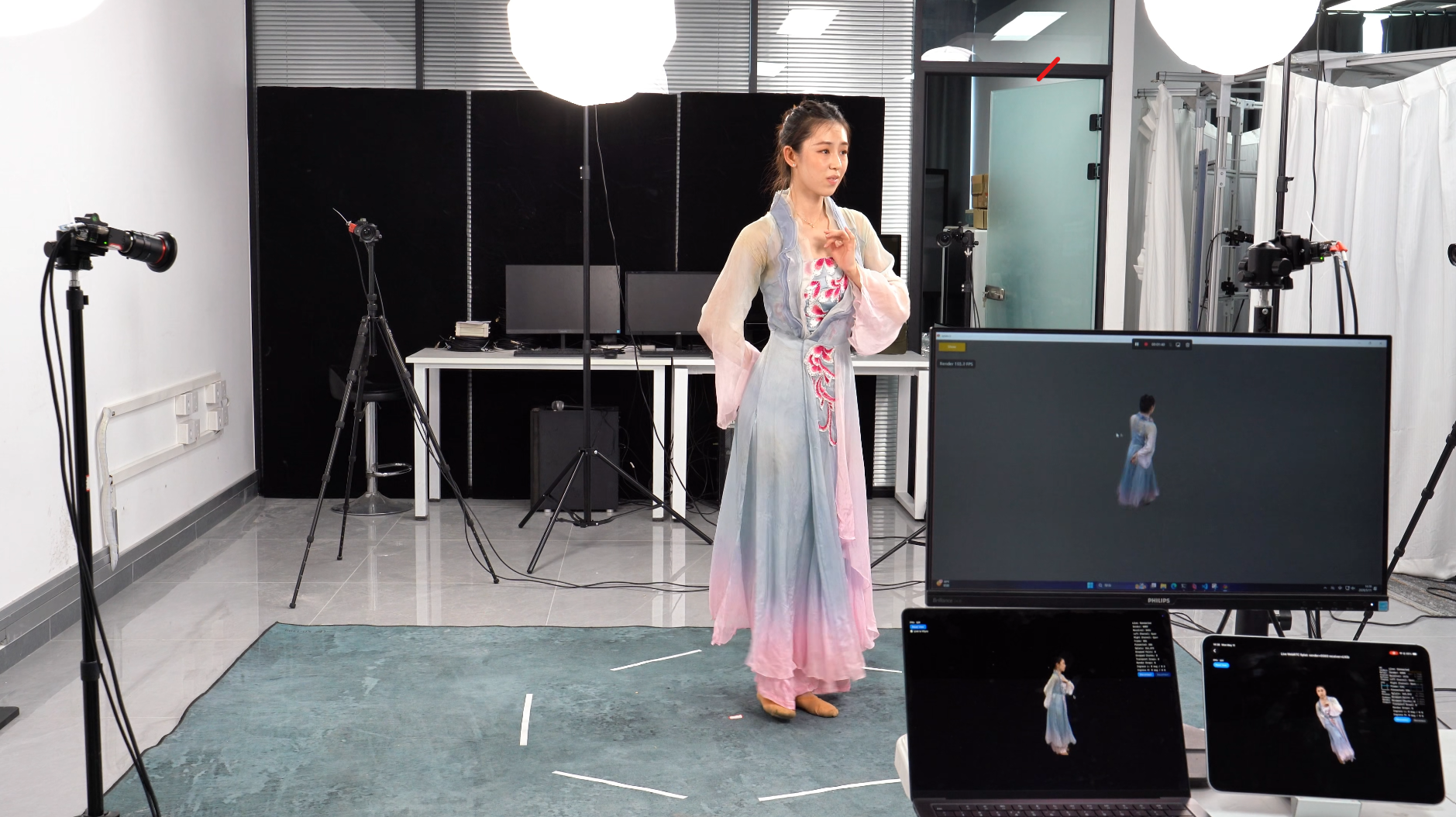}
    \caption{Photographs of our deployed system. Our system supports multiple users simultaneously viewing live 3DGS streams transmitted over the network and freely controlling the viewpoint.}
    \label{fig:systemoverview}
\end{figure}

An overview of our system is shown in Fig.~\ref{fig:systemoverview}. 
We design both the hardware and software stack with 
deployment and interactive viewing in mind.
Overall, the system comprises the following components: multi-view video capture, reconstruction, stream encoding, transmission, stream decoding, and endpoint display.
It enables remote free-viewpoint viewing of the captured performer across a range of client platforms, including autostereoscopic display-equipped PCs for glasses-free 3D experiences and mobile devices (e.g., tablets) for portable access.

\subsection{Multi-View Capture System}
Our system captures 4 to 6 video streams to provide 360$^\circ$ coverage of the performer. Each camera (BFS-U3-123S6C-C) records video at a resolution of $4096 \times 3000$ at $30 \operatorname{Hz}$. 
The cameras are approximately evenly spaced on a circular rig around the performer. Note that our system is not tied to specific camera hardware, requires no pre-calibration, and can be deployed with commodity RGB cameras.
Although the system does not require pre-calibration, optional cross-camera color calibration using a standard ColorChecker can be performed to improve appearance consistency across views.
\subsection{Data Encoding, Transmission and Decoding}
\label{subsec:trans}
Upon receiving new frames from the capture system, we convert the streams to the input format required by our reconstruction pipeline (2K) and run inference fully online. We leverage TensorRT to accelerate non-transformer components. For transformer layers, we adopt FlashAttention~\cite{dao2023flashattention2} and implement custom fused kernels via Triton. To maximize overall efficiency, we also implement the remaining computations in Triton.

To enable real-time, free-viewpoint rendering on remote clients, we design an efficient streaming pipeline that leverages standard hardware-accelerated video codecs.
Since our feed-forward reconstruction network outputs pixel-aligned Gaussian maps, the maps retain a 2D spatial structure rather than forming unordered Gaussian sets and can therefore be naturally reorganized into large image canvases for transmission.

\begin{figure}[t!]
    \centering
    \includegraphics[width=1.0\linewidth]{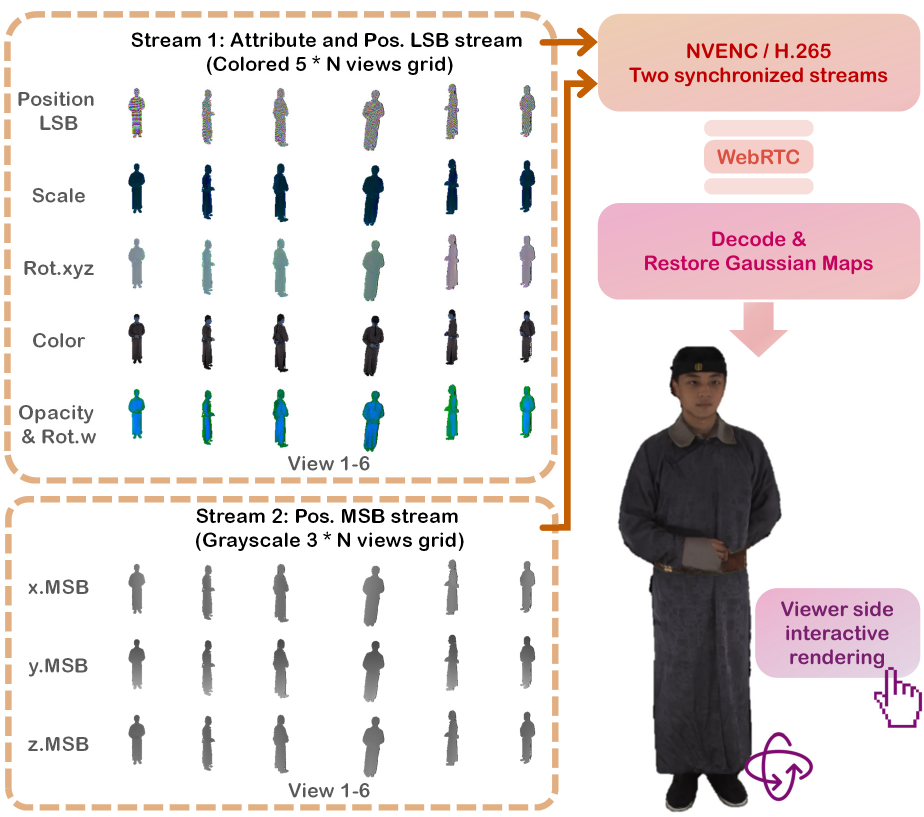}
        \caption{Codec-friendly streaming of our pixel-aligned Gaussian maps. We transmit Gaussian maps directly over the network at a controllable bandwidth, enabling users to interact in real time on their devices. This fixed-layout representation enables standard hardware video coding without serializing a variable-length Gaussian set.}
    \label{fig:supp-compress}
    \end{figure}

Our streaming pipeline is illustrated in Fig.~\ref{fig:supp-compress}. 
To preserve the geometric precision of the reconstructed 3D Gaussians while remaining compatible with 8-bit video encoders, we introduce a split-byte encoding scheme for the 3D point map. Specifically, the predicted float32 coordinates $(x,y,z)$ are first quantized to uint16. Each 16-bit value is then decomposed into a most significant byte (MSB) and a least significant byte (LSB). The MSBs (middle part of Fig.~\ref{fig:supp-compress}) of the three coordinate channels are packed into a synchronized single-channel grayscale stream, arranged as a $3 \times N_{\mathrm{view}}$ grid so that the three stacked rows correspond to the high-order bytes of the $X, Y$ and $Z$ coordinates, respectively. The LSBs are packed together with the remaining Gaussian attributes into a second synchronized RGB stream (left part of Fig.~\ref{fig:supp-compress}). Concretely, the tiled RGB canvas contains the remaining low-order bytes of the point map (Row 1), the Gaussian scale parameters (Row 2), the first three components of the quaternion $(q_x, q_y, q_z)$ (Row 3), the color map (Row 4), and the remaining scalar channels including $q_w$ and opacity $\alpha$ (Row 5, with one unused channel serving as alignment padding). 
By tiling the multi-view maps into a single cohesive frame (e.g., $6K \times 5K$ for the RGB stream and $6K \times 3K$ for the grayscale stream), we avoid the overhead of transmitting multiple separate videos.

We encode the resulting streams with H.265 using hardware-accelerated NVENC and transmit them to the remote viewer via WebRTC. On the receiver side, hardware decoders restore the Gaussian maps for interactive rendering with our highly efficient, cross-platform WebGPU viewer. In our measurements, the end-to-end system requires approximately 100 Mbit/s of bandwidth, which is feasible in common network environments.

\paragraph{System Throughput and Latency.}
Tele360 uses a multi-stage pipeline that overlaps capture, matting, reconstruction, encoding, transmission, reception, and rendering across consecutive frames. This design prioritizes steady-state throughput at the cost of increased end-to-end latency. Outside the reconstruction network, RVM is the only additional GPU workload and requires 2.9~ms for six views; capture and transmission run on CPUs, while H.265 encoding uses dedicated NVIDIA codec hardware, minimizing contention with reconstruction. The complete deployed system, including non-algorithmic components, occupies approximately 20~GB of GPU memory. The client-side reception, decoding, and rendering pipeline sustains 30~FPS, with a viewing delay that depends on the client device. We define end-to-end latency as the delay from a real-world action to its appearance on the client; it is approximately 500~ms on a MacBook Pro and 650~ms on an iPad Pro.

\section{Experiments}

\subsection{Network Architecture}
We provide additional details of the network architecture in this section. 
For hybrid feature extraction, we use a frozen DINOv3~\cite{simeoni2025dinov3} encoder and an EdgeNeXt~\cite{edgenext} encoder. Specifically, we use the ViT-B~\cite{vit} variant of DINOv3 and EdgeNeXt-xxs equipped with STDA blocks, which provides an efficient design for processing multiple high-resolution input images simultaneously.
The extracted features are then fed into a transformer backbone with 24 blocks for cross-view 3D reasoning.
Among them, 12 are intra-frame blocks and 12 are inter-frame blocks, arranged in an alternating manner. 
The feature dimension of the transformer backbone is set to 1024, with 16 attention heads.
RoPE~\cite{su2024roformer} with a frequency of 100 is applied to all blocks. We also employ QKNorm~\cite{henry2020query} and LayerScale~\cite{touvron2021going} initialization to improve training stability.
The decoder transformer follows a similar architecture to the backbone. It contains 12 blocks, which are organized into 4 groups, each consisting of two self-attention layers followed by one cross-attention layer.
For the four cross-attention layers, the key and value features are taken from the 5th, 11th, 17th, and 23rd blocks of the transformer backbone, respectively, together with the four levels of the CNN feature pyramid. 
Finally, a lightweight MLP maps the decoder outputs to the point predictions, while several shallow CNN heads are used to predict the Gaussian parameter maps.

\subsection{Experimental Settings}
\noindent\textbf{Datasets.}
Our method is trained on a large-scale dataset comprising 3126 human scans~\cite{yu2021function4d,han2023high,twindom} and approximately 1,000 multi-view video sequences from DNA-Rendering~\cite{cheng2023dna}. The video sequences contain 60 viewpoints and range from 150 to 225 frames. For the static scans, we utilize Blender to render images from 60 viewpoints uniformly surrounding the subjects. For the DNA-Rendering data, we refine the provided foreground segmentation masks and employ NeuS2~\cite{neus2} to obtain the ground-truth geometry for supervision. During training, 4 to 6 views are randomly sampled from a scene and fed into the network. To ensure optimal load balancing, we keep the number of processed images and token lengths roughly equivalent across GPUs for each batch.

\noindent\textbf{Baselines.}
We compare our method against methods from two main categories:
        \textit{Generalized feed-forward reconstruction methods.} Methods in this category do not require camera parameters and can recover the scene geometry and appearance from uncalibrated images in a feed-forward manner. Representative methods include NoPoSplat~\cite{noposplat}, AnySplat~\cite{anysplat} and Depth Anything 3~\cite{da3}. 
        \textit{Human-specific feed-forward sparse-view reconstruction methods.} These methods typically require accurate camera calibration and are designed for human reconstruction, including DoubleField~\cite{shao2022doublefield}, GHG~\cite{kwon2024generalizable} and GPS-Gaussian~\cite{gps}.
In addition, we compare against NeuS2~\cite{neus2} as a representative optimization-based method. It can recover geometry and appearance from sparse views relatively quickly. 
Input viewpoints are uniformly sampled from a circular camera array and kept identical across methods. 

\noindent\textbf{Evaluation Metrics.}
We assess the quality of novel-view rendering using the Peak Signal-to-Noise Ratio~(PSNR), Structural Similarity Index Measure~(SSIM)~\cite{wang2004image} and LPIPS~\cite{zhang2018the} metrics at $2048\times2048$ resolution. 
For models rendering at alternative resolutions (e.g., Depth Anything 3~\cite{da3}, AnySplat~\cite{anysplat}), we bilinearly upsample their outputs to $2048\times2048$ resolution. 
For methods that require camera parameters as input, we provide ground-truth poses; we explicitly mark this setting in all reported results.

We evaluate pose accuracy with Relative Rotation Error~(RRE) and Relative Translation Error~(RTE). 
RRE@K and RTE@K denote the percentage of camera pairs with errors below $K^\circ$, and AUC@5/10/30 measures the area under the pose error curve up to the corresponding thresholds. 
We also report the mean and median RRE/RTE.

To ensure a fair comparison on datasets with backgrounds, we follow each method's prescribed input setting: NoPoSplat~\cite{noposplat}, AnySplat~\cite{anysplat}, and Depth Anything~3~\cite{da3} receive the original full images, whereas GPS-Gaussian~\cite{gps} receives foreground-only images, consistent with its training protocol.
After reconstruction and before rendering, we apply the same foreground mask to the outputs of all methods to remove background Gaussians and render all results against a uniform black background.

\noindent\textbf{Implementation Details.} Our model is trained on a single server equipped with 8 NVIDIA A100 GPUs for a total of 300K iterations. We employ the AdamW optimizer with a learning rate of $5\times10^{-5}$ and weight decay of $0.05$. The learning rate is linearly warmed up from $10^{-8}$ to $5\times10^{-5}$ during the first $2\%$ of training, followed by cosine decay to $10^{-8}$. We use gradient accumulation over 4 steps and clip the gradient norm to 1.0.

\begin{figure*}[ht!]
    \centering
    \includegraphics[
        width=\textwidth,
        height=0.92\textheight,
        keepaspectratio
    ]{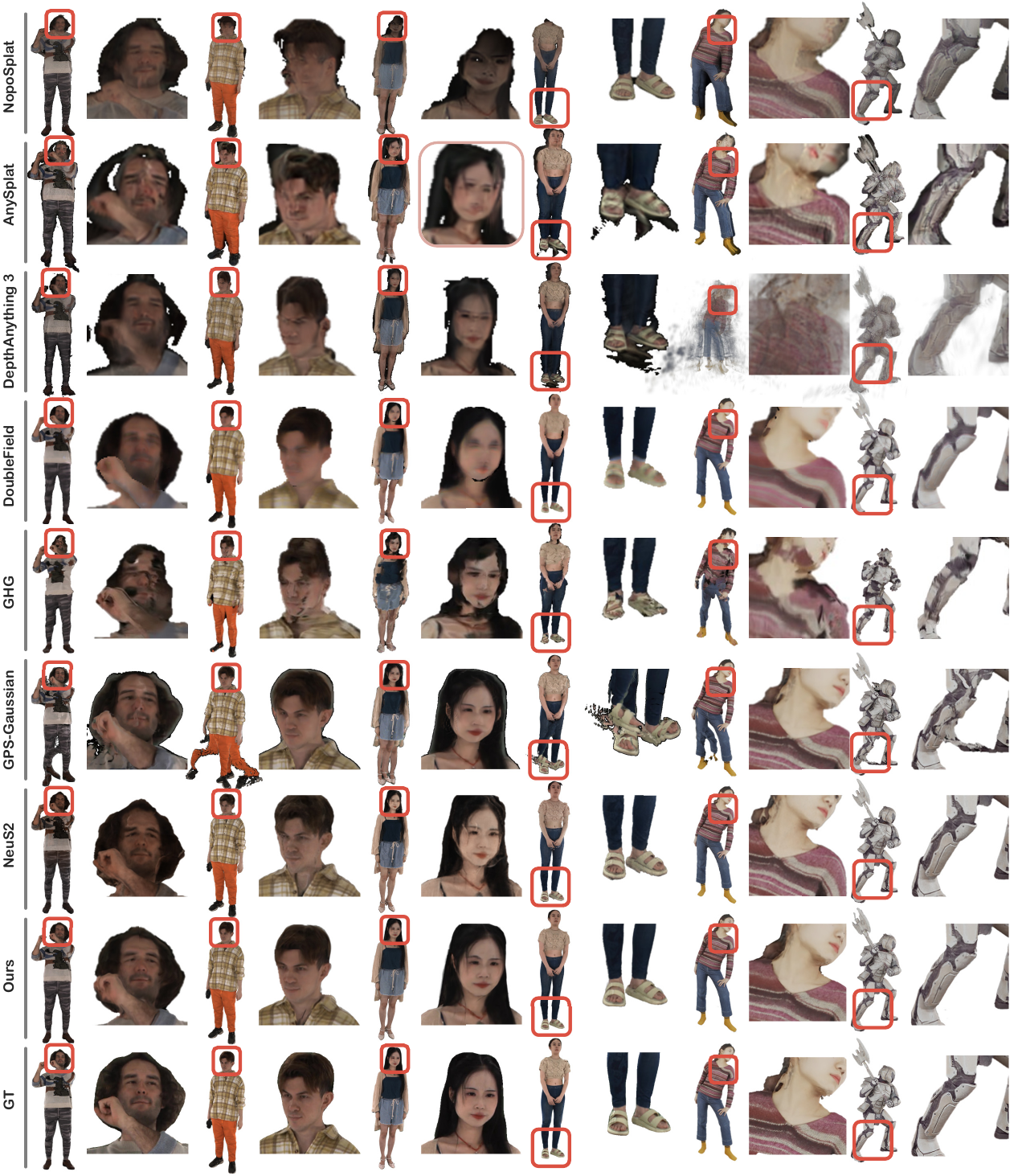}
        \caption{Qualitative comparisons. We compare our method with state-of-the-art baselines~\cite{noposplat,anysplat,da3,shao2022doublefield,kwon2024generalizable,gps,neus2}. Zoom-in patches highlight fine-grained details. While existing methods suffer from blurry textures or severe geometric artifacts, our method successfully reconstructs high-fidelity details (e.g., facial features and footwear), closely matching the ground truth.}
    \label{fig:comp-main}
    \end{figure*}
\begin{table*}[t!]
\centering
\caption{Quantitative comparison of novel view synthesis on the DNA-Rendering~\cite{cheng2023dna} dataset. $^*$ For two-view methods, adjacent views are paired as inputs, and timing is accumulated over all pairs. $^{\dagger}$GPS-Gaussian requires stereo rectification, which may fail given 6 views. ``Align'' denotes an evaluation-time camera alignment step used for NoPoSplat~\cite{noposplat} and AnySplat~\cite{anysplat} when reporting novel-view synthesis metrics.}
\label{tab:comp-rendering-main}
\begin{tabular}{@{}cccccccc@{}}
\toprule
Method                      & w/calib. & View No. &~Align~  & ~PSNR$\uparrow$~ & ~SSIM$\uparrow$~ & ~LPIPS$\downarrow$~ & ~Time$\downarrow$~ \\ \midrule
NoPoSplat~\cite{noposplat}  & No     & 6 & Yes  &  22.1677       &     0.8574     &    0.1433   &  \textasciitilde600 ms$^*$  \\
AnySplat~\cite{anysplat}    & No     & 6 & Yes  &  24.9883       &     0.9163     &    0.0781   &  \textasciitilde150 ms  \\
Depth Anything~3~\cite{da3}   & No     & 6 & No   &  21.1142       &     0.9115     &    0.0926   &  \textasciitilde300 ms  \\
DoubleField~\cite{shao2022doublefield}   & Yes     & 6 & No   &  29.4737       &     0.9596     &    0.0523   &  $\gg$1 s  \\
GHG~\cite{kwon2024generalizable}   & Yes     & 6 & No   &  26.2712       &     0.9458     &    0.0534   &  \textasciitilde150 ms  \\
GPS-Gaussian~\cite{gps}     & Yes    & 8$^{\dagger}$ & No   &  22.4023       &     0.8675     &    0.1312   &  \textasciitilde300 ms$^*$  \\
NeuS2~\cite{neus2}   & Yes     & 6 & No   &  30.6106       &     0.9683     &    0.0336   &  $\gg$1 s  \\
Ours                        & No     & 6 & No   &  \textbf{32.1502} & \textbf{0.9689} & \textbf{0.0286} &  \textasciitilde35 ms  \\ \bottomrule
\end{tabular}
\end{table*}
\begin{table*}[t]
\centering
\caption{Quantitative comparison of camera pose estimation results. Best results are highlighted in \textbf{bold}, and second-best results are \underline{underlined}. In COLMAP, SP+SG denotes using SuperPoint for feature extraction and SuperGlue for cross-view feature matching. Under the sparse surrounding-camera setup, COLMAP fails to register all cameras in over 40\% of the test scenes; therefore, mean and median RRE/RTE are not reported.}
\label{tab:pose_comparison}
\resizebox{\textwidth}{!}{
\begin{tabular}{c|ccc|ccc|ccc|cccc}
\toprule
\textbf{Method}
& \multicolumn{3}{c|}{\textbf{RRE Acc.}~$\uparrow$}
& \multicolumn{3}{c|}{\textbf{RTE Acc.}~$\uparrow$}
& \multicolumn{3}{c|}{\textbf{AUC}~$\uparrow$}
& \multicolumn{4}{c}{\textbf{Error}~$\downarrow$} \\
\cmidrule(lr){2-4}
\cmidrule(lr){5-7}
\cmidrule(lr){8-10}
\cmidrule(lr){11-14}
& \textbf{@1}  & \textbf{@2}  & \textbf{@5} 
& \textbf{@1}  & \textbf{@2}  & \textbf{@5} 
& \textbf{@5}  & \textbf{@10}  & \textbf{@30} 
& \textbf{Mean RRE}  & \textbf{Mean RTE} 
& \textbf{Med. RRE}  & \textbf{Med. RTE}  \\
\midrule
COLMAP~(SP+SG)~\cite{schonberger2016structure}
& 12.58 & 31.10 & 38.33
& 15.37 & 25.64 & 40.41
& 25.63 & 33.13 & 41.96
& -- & -- & -- & -- \\

VGGT~\cite{vggt}
& 22.14 & 72.80 & 98.29
& 32.51 & 69.08 & 96.28
& 63.82 & 81.38 & 93.79
& 1.72 & 1.74 & 1.52 & 1.41 \\

DA~3~\cite{da3}
& \underline{24.29} & \textbf{83.21} & \textbf{99.94}
& 37.58 & 74.71 & \underline{99.62}
& \underline{67.86} & \underline{83.90} & \underline{94.63}
& \textbf{1.44} & 1.47 & \textbf{1.38} & 1.36 \\

Our model (w/o differentiable BA module)
& 21.83 & 72.10 & 98.53
& \underline{46.40} & \underline{92.38} & 99.37
& 66.74 & 82.90 & 94.25
& 1.73 & \underline{1.14} & 1.54 & \underline{1.05} \\

Our model (w/ differentiable BA module)
& \textbf{25.17} & \underline{77.73} & \underline{99.25}
& \textbf{47.54} & \textbf{93.65} & \textbf{99.81}
& \textbf{69.31} & \textbf{84.49} & \textbf{94.83}
& \underline{1.55} & \textbf{1.07} & \underline{1.43} & \textbf{1.04} \\
\bottomrule
\end{tabular}
}
\end{table*}

\subsection{Evaluation}
\noindent\textbf{Qualitative Evaluation.}
As illustrated in Figure~\ref{fig:comp-main}, our method exhibits clear visual superiority over existing baselines. 
Competing approaches struggle to accurately render complex human structures. 
In the sparse 6-view 360° setting, uncalibrated methods such as NoPoSplat~\cite{noposplat}, AnySplat~\cite{anysplat}, and Depth Anything 3~\cite{da3} struggle with alignment, leading to degraded quality, severe geometric distortions, and boundary artifacts such as fragmented limbs and noisy outlines.
GHG~\cite{kwon2024generalizable} relies on accurate SMPL~\cite{loper2023smpl} estimates; inaccurate SMPL fitting, particularly around the face and body, leads to degraded performance.
Although DoubleField~\cite{shao2022doublefield} produces relatively accurate geometry, its NeRF-based~\cite{mildenhall2020nerf} formulation struggles to capture high-frequency details.
GPS-Gaussian~\cite{gps} struggles to perform reliable stereo matching when the input views exhibit large image disparities, leading to erroneous depth estimates and misaligned leg reconstruction.
Due to the limited number and sparsity of input views, NeuS2~\cite{neus2} also struggles to produce high-quality reconstructions.
In contrast, our approach robustly recovers fine-grained visual details—yielding remarkably crisp facial features and highly accurate extremities—achieving photorealistic rendering quality that faithfully aligns with the ground truth.

\begin{figure}[t!]
    \centering
    \includegraphics[width=1.0\linewidth]{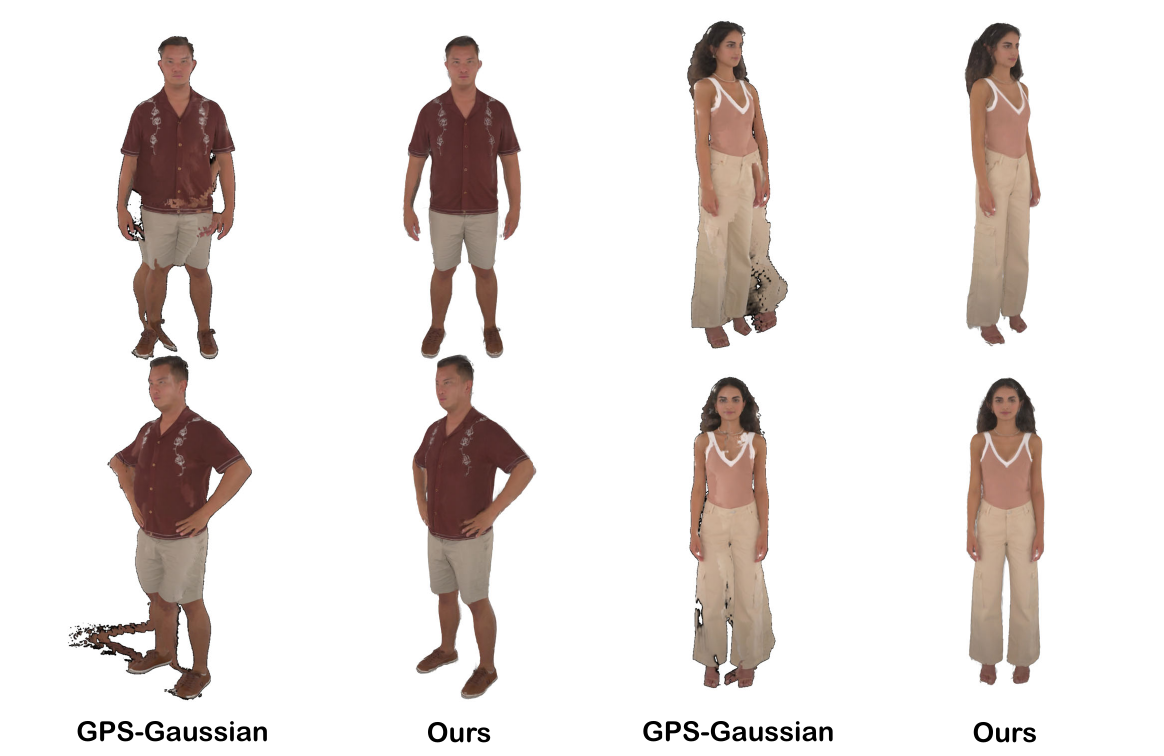}
    \caption{Qualitative comparisons with the person-specific method GPS-Gaussian.}
    \label{fig:supp-comp-gps}
\end{figure}

We provide additional qualitative comparisons with the person-specific method GPS-Gaussian~\cite{gps} on the ActorsHQ~\cite{icsik2023humanrf} dataset, as shown in Fig.~\ref{fig:supp-comp-gps}. Compared with GPS-Gaussian, our method produces cleaner and more stable reconstructions with noticeably higher rendering quality. In particular, our results exhibit fewer floating artifacts and significantly more coherent object boundaries, while GPS-Gaussian often suffers from jittery edges and unstable geometry around fine structures.

\begin{table}[t!]
\centering
\caption{Quantitative comparison of novel view synthesis on the ActorsHQ~\cite{icsik2023humanrf} dataset for human-only methods. }
\label{tab:comp-actorshq}
\begin{tabular}{@{}cccc@{}}
\toprule
Method                        & ~PSNR$\uparrow$~ & ~SSIM$\uparrow$~ & ~LPIPS$\downarrow$~  \\ \midrule
DoubleField~\cite{shao2022doublefield}   &  28.18       &     0.9479     &    0.0386   \\
GHG~\cite{kwon2024generalizable} &  25.52       &     0.9249     &    0.0672   \\
GPS-Gaussian~\cite{gps}    &  23.62       &     0.9527     &    0.0521 \\
Ours                       & \textbf{30.72} & \textbf{0.9573} & \textbf{0.0346}  \\ \bottomrule
\end{tabular}
\end{table}

\noindent\textbf{Quantitative Evaluation.}
As demonstrated in Table~\ref{tab:comp-rendering-main}, our method significantly outperforms all baselines across all evaluation metrics. 
Notably, our approach achieves a substantial margin of over 1.5 dB in PSNR compared to the second-best method (NeuS2~\cite{neus2}), while yielding the lowest LPIPS score (0.0286). 
Furthermore, our model exhibits exceptional efficiency with an inference time of merely \textasciitilde35 ms, more than 4 times faster than competing approaches, achieving state-of-the-art rendering quality without relying on camera parameter input or alignment operations. 

Note that in Table~\ref{tab:comp-rendering-main}, ``Align'' denotes an evaluation-time camera alignment step used for NoPoSplat~\cite{noposplat} and AnySplat~\cite{anysplat} when reporting novel-view synthesis metrics. 
Since the reconstructed Gaussians and camera poses are predicted in a method-specific canonical coordinate system, the target-view camera parameters provided by the dataset cannot be directly used for rendering. 
To enable a fair comparison, we first compute a coarse global similarity alignment between the dataset camera system and the predicted canonical space, and then further refine the target-view pose using a mask-based rendering objective. 

We further evaluate the accuracy of our camera estimation module. 
As shown in Table~\ref{tab:pose_comparison}, our method leads in RTE accuracy and AUC across all reported thresholds, while its RRE performance remains comparable to DA~3.
These results demonstrate that our module can reliably recover camera poses from sparse input views.
Note that rendering quality cannot be directly inferred from pose estimation accuracy, since depth and camera poses are inherently coupled.

\noindent\textbf{Temporal Consistency.}
Beyond per-frame reconstruction and pose accuracy, we evaluate temporal stability by rendering each sequence from a fixed novel viewpoint and computing the motion smoothness and temporal flickering metrics of VBench~\cite{huang2024vbench}.
As shown in Table~\ref{tab:temporal-consistency}, our method achieves the highest scores on both metrics among the compared methods.
The predicted cameras also exhibit low intra-sequence jitter (mean/median): $0.27^\circ/0.24^\circ$ in rotation, $3.3\times10^{-3}/2.8\times10^{-3}$ in translation, and $0.29\%/0.26\%$ in focal length.

\begin{table}[t]
\centering
\small
\caption{Temporal-consistency results. Values are measured from renderings at a fixed novel viewpoint.}
\label{tab:temporal-consistency}
\begin{tabular}{@{\extracolsep{\fill}}ccc@{}}
\toprule
Method & Motion Smooth.$\uparrow$ & Temp. Flicker.$\uparrow$ \\ \midrule
GPS-Gaussian~\cite{gps} & 0.99441 & 0.99202 \\
DA~3~\cite{da3}         & 0.99386 & 0.99277 \\
Ours                    & \textbf{0.99820} & \textbf{0.99669} \\ \bottomrule
\end{tabular}
\end{table}

\subsection{Ablation Study}

\begin{table}[t]
\centering
\caption{Quantitative ablation study on the proposed hybrid feature extraction. In the model without hybrid feature extraction, DINO is given higher-resolution images. }
\label{tab:abl-hybrid-feat}
\resizebox{\columnwidth}{!}{
\begin{tabular}{@{}ccccc@{}}
\toprule
Method             & PSNR$\uparrow$   & SSIM$\uparrow$  & LPIPS$\downarrow$ & Inference Time$\downarrow$ \\ \midrule
w/o Hybrid Feature & 28.2609          & 0.9473          & 0.0557          & 43 ms                                          \\
w/ Hybrid Feature  & \textbf{30.3645} & \textbf{0.9615} & \textbf{0.0490} & \textbf{35 ms}                                 \\ \bottomrule
\end{tabular}
}
\end{table}

\noindent\textbf{Ablation on hybrid feature extraction.}
\begin{figure}[ht!]
    \centering
    \includegraphics[width=1.0\linewidth]{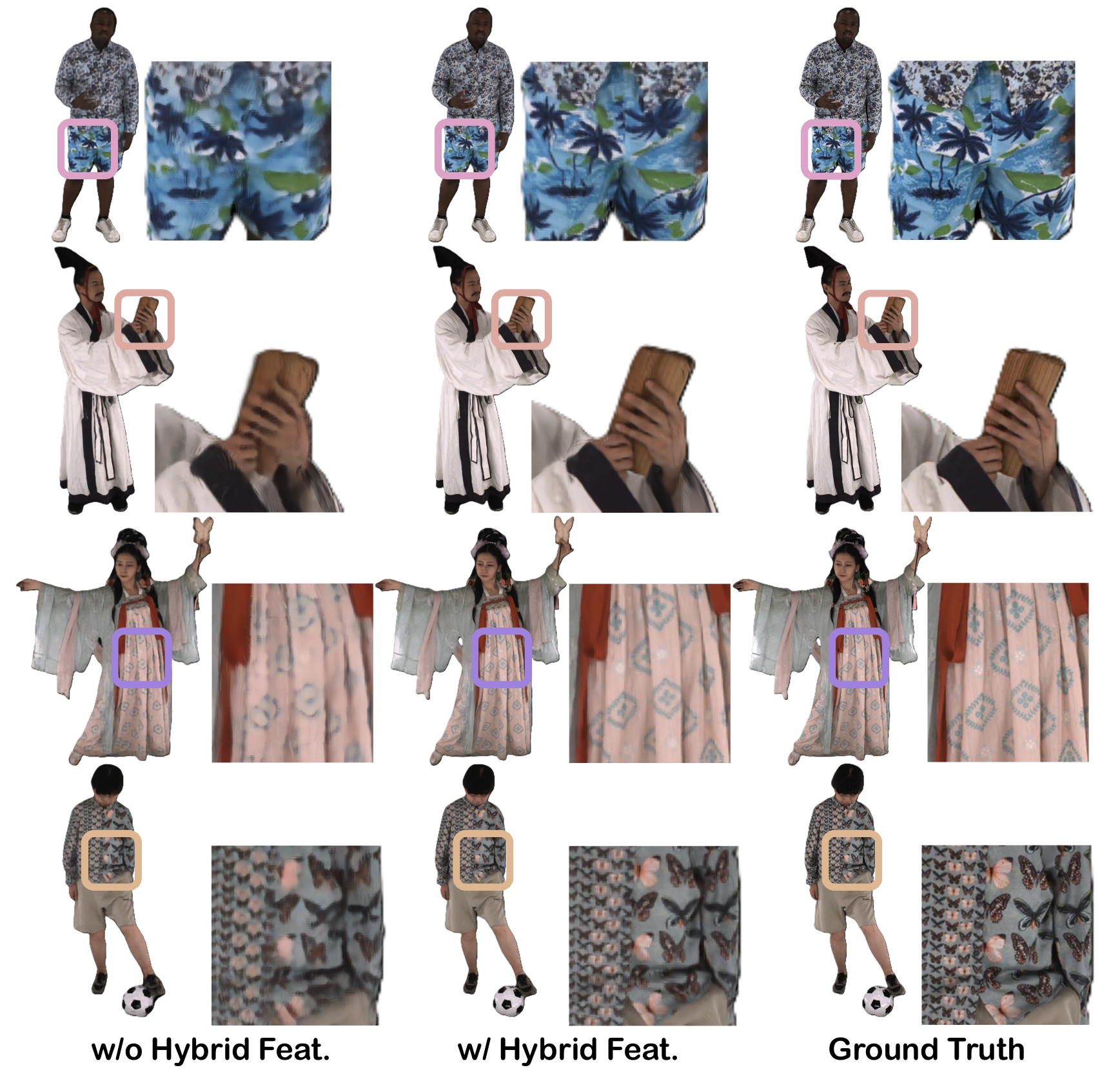}
    \caption{Qualitative ablation study on the proposed hybrid feature extraction. Removing this module (w/o Hybrid Feat.) leads to excessively smoothed and blurred textures.}
    \label{fig:abl-hybrid}
\end{figure}
To validate the efficacy of our proposed hybrid feature extraction strategy, we present a qualitative ablation study in Figure~\ref{fig:abl-hybrid} and Table~\ref{tab:abl-hybrid-feat}. 
For the ablated variant, we removed the CNN and doubled DINO's input resolution. 
Despite the significantly increased computational burden, the network still struggles to fully exploit the spatial information from high-resolution inputs, resulting in noticeably blurry textures. In contrast, integrating the hybrid features allows our model to fully capitalize on the high-resolution images. 

\begin{table}[]
\centering
\caption{Ablation of decoder attention patterns on THuman~\cite{yu2021function4d}, evaluated by Chamfer distance. SA, FA, and CA denote intra-frame self-attention, inter-frame attention, and cross-attention to multi-scale conditioning features, respectively. Each sequence specifies the ordered attention blocks within a decoder stage.}
\label{tab:comp-decoder-abl}
\begin{tabular}{@{}cc@{}}
\toprule
Method                        & ~Chamfer Distance$\downarrow$~ \\ \midrule
W/o hybrid feature extraction &   0.00953       \\
SA + CA                            &   0.00755       \\
SA + FA + CA                           &   0.00552       \\
CA + CA + CA                           &   0.00548       \\
Our model (SA + SA + CA)                   &   \textbf{0.00541}       \\ \bottomrule
\end{tabular}
\end{table}

\noindent\textbf{Ablation of different architectures.}
We study alternative decoder designs to understand the trade-off between efficiency and reconstruction quality (Table~\ref{tab:comp-decoder-abl}). Our final decoder adopts a sparse refinement scheme with intra-view self-attention and cross-attention to multi-scale conditioning features, which provides the best balance in our setting. We also experimented with additional cross-view (inter-frame/inter-view) attention blocks inside the decoder to further mix information across views. However, we find that multi-view interaction captured by the backbone is already sufficient: adding extra cross-view attention in the decoder brings negligible improvement while noticeably increasing runtime. Consequently, we keep the decoder view-independent to preserve real-time performance.

\begin{figure}[t]
    \centering
    \includegraphics[width=1.0\linewidth]{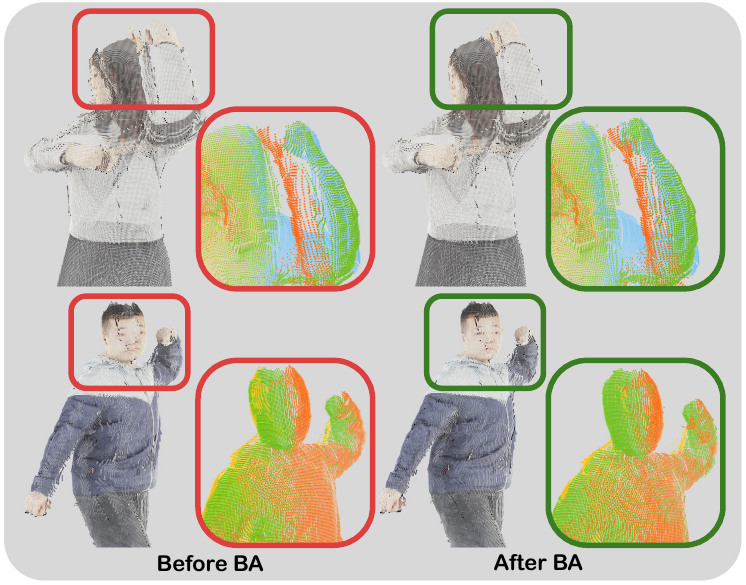}
    \caption{Visualization of the camera BA process. Point maps from different views are shown in different colors. The proposed BA module improves multi-view geometric consistency, which is crucial for human-centric sparse-view reconstruction.}
    \label{fig:abl-ba-vis}
\end{figure}

\noindent\textbf{Ablation on camera BA module.}
To validate the effectiveness of the differentiable camera BA module, we provide a qualitative ablation in Figure~\ref{fig:abl-ba-vis} and a quantitative pose comparison in Table~\ref{tab:pose_comparison}.
Specifically, Figure~\ref{fig:abl-ba-vis} visualizes the point clouds before and after applying camera BA.
Although the point clouds directly predicted by the network are already reasonably aligned, the proposed BA module further refines the camera poses and improves point-cloud alignment.
This demonstrates that our differentiable BA module effectively optimizes camera alignment.

\noindent\textbf{Effect of camera input.}
Tele360 also accepts calibrated camera parameters as model inputs; in that case, it bypasses the camera decoder and BA (Fig.~\ref{fig:pipeline}).
We evaluate this mode by feeding ground-truth cameras under the protocol of Table~\ref{tab:comp-rendering-main}.
As shown in Table~\ref{tab:camera-input-ablation}, this yields only marginal PSNR and LPIPS gains, with unchanged SSIM and runtime, indicating that residual camera-estimation errors have limited impact on reconstruction quality.

\begin{table}[t]
\centering
\small
\caption{Camera-input ablation. Pred. estimates cameras internally; GT input supplies ground-truth cameras to the model and bypasses the camera decoder and BA.}
\label{tab:camera-input-ablation}
\begin{tabular}{@{\extracolsep{\fill}}ccccc@{}}
\toprule
Camera mode & PSNR$\uparrow$ & SSIM$\uparrow$ & LPIPS$\downarrow$ & Time$\downarrow$ \\ \midrule
Pred.    & 32.15 & 0.969 & 0.0286 & \textasciitilde35 ms \\
GT input & 32.59 & 0.969 & 0.0279 & \textasciitilde35 ms \\ \bottomrule
\end{tabular}
\end{table}

\noindent\textbf{Robustness to foreground masks.}
We evaluate the effect of mask quality on a THuman subset by replacing only the foreground masks while keeping the remaining reconstruction and evaluation pipeline unchanged. We compare MatAnyone~2~\cite{yang2026matanyone2}, RVM~\cite{lin2021robust}, and SAM~3~\cite{carion2026sam3} with the text prompt ``person'', together with ground-truth masks.
\begin{table}[t]
\centering
\caption{Robustness to foreground-mask quality on a THuman subset. Depth denotes depth error; CD denotes Chamfer distance; Cam. Rot. and Cam. Trans. denote camera rotation error in degrees and camera translation error, respectively. GT denotes ground-truth masks. }
\label{tab:mask-robustness}
\resizebox{\columnwidth}{!}{
\begin{tabular}{@{}cccccc@{}}
\toprule
Mask & IoU$\uparrow$ & Depth$\downarrow$ & CD$\downarrow$ & Cam. Rot./Trans.$\downarrow$ & PSNR$\uparrow$ \\ \midrule
MatAnyone 2 & 0.9940 & 0.0053 & 0.0033 & 0.916/0.0261 & 27.23 \\
RVM-R50     & 0.9922 & 0.0053 & 0.0033 & 1.013/0.0282 & 27.12 \\
SAM 3       & 0.9796 & 0.0070 & 0.0042 & 2.993/0.0648 & 26.67 \\
GT          & 1.0000 & 0.0053 & 0.0032 & 0.886/0.0254 & 27.34 \\ \bottomrule
\end{tabular}
}
\end{table}
High-quality mattes approach ground-truth performance across geometry, camera, and rendering metrics, whereas lower-quality masks consistently degrade all three, confirming sensitivity to matting quality.

\subsection{Runtime and Sparsity Analysis}

\paragraph{Module-wise Runtime Analysis.} 
Table~\ref{tab:supp-moduletime} reports the module-wise inference time of our model. 
The sparse transformer backbone accounts for the largest portion of the runtime, followed by the prediction heads and the decoder. 
The feature extraction stage remains relatively efficient.
Overall, the full model takes 35.36 ms per frame under the 6-view 2K setting, demonstrating that the total inference time of the model remains compatible with real-time deployment. 
These timings characterize model throughput rather than action-to-display latency; system-level latency is reported in Sec.~\ref{sec:system}.

\begin{table}[t]
\centering
\caption{Module-wise inference time of our model on the DNA-Rendering~\cite{cheng2023dna} dataset under the 6-view 2K input setting, using a single RTX 5090 GPU.}
\label{tab:supp-moduletime}
\begin{tabular}{@{\extracolsep{\fill}}c|ccccc@{}}
\toprule
Module & DINO & CNN  & Backbone & Decoder & Heads \\ \midrule
Runtime (ms) & 5.15 &  2.85  & 12.15  & 6.07  &  9.14  \\ \bottomrule
\end{tabular}
\end{table}

\paragraph{Effect of token sparsity.} 

Table~\ref{tab:supp-sparsity-runtime} analyzes the effect of token sparsity on runtime under different numbers of input views. 
As mentioned above, we extract only the valid foreground tokens according to the predicted mask. In our observations on the DNA-Rendering dataset~\cite{cheng2023dna}, the valid tokens typically account for around 10\% of all image tokens.
To evaluate whether our real-time performance relies on an overly sparse setting, we further enlarge the valid token ratio to 15\% and 20\%, simulating cases where the human occupies a larger image region.
Table~\ref{tab:supp-sparsity-runtime} shows that, as the retained token ratio increases, most of the runtime increase comes from the transformer-based backbone and decoder, consistent with the token-sensitive nature of transformer computation. By comparing the ``Backbone + Decoder'' and ``Total Inference'' timings, we observe that the remaining modules contribute nearly constant overhead and are much less affected by the token ratio. Importantly, both the 4-view and 6-view settings remain efficient even at a 20\% token ratio, showing that our system does not rely on extremely sparse foreground occupancy. 

\paragraph{Gaussian count and streaming bandwidth.}
Since each foreground pixel instantiates one Gaussian, the Gaussian count scales approximately linearly with the number of input views, typically ranging from about 400--600K with four views and 600--900K with six views. The fixed canvas layout makes bandwidth largely independent of Gaussian count: the deployed end-to-end system requires approximately 100~Mbit/s for six-view setting (Sec.~\ref{subsec:trans}).

\paragraph{Analysis of quantization and compression.}
Since our system requires quantization and video coding/decoding during transmission, we further analyze the performance changes after quantization and after video encoding/decoding. As shown in Table~\ref{tab:metric-quant}, our model maintains strong performance even after quantization and coding. 
This robustness mainly arises from the compact data range and smooth variations of the Gaussian maps across different views, which effectively reduce the precision loss introduced during quantization and coding. 

\begin{figure}
    \centering
    \includegraphics[
        width=\columnwidth,
        keepaspectratio
    ]{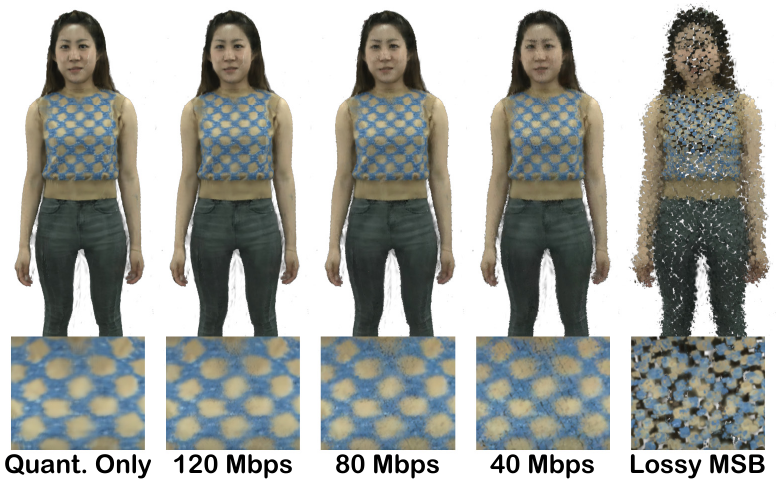}
        \caption{Qualitative effects of Gaussian-map compression. Lower bitrates reduce texture fidelity, while lossy MSB coding causes severe geometric artifacts.}
    \label{fig:codec-visual}
\end{figure}

We also analyze the rate-distortion trade-off, as shown in Table~\ref{tab:bitrate-quality}. Quantization reduces PSNR from 32.65 to 31.10 dB. With lossless MSB coding and an LSB QP of 28, the complete stream requires approximately 80 Mbit/s and achieves 30.57 dB. Increasing the QP reduces the bitrate to approximately 40 Mbit/s at 29.72 dB. After weighing rendering quality against transmission bandwidth, we ultimately selected a QP of 28 for the LSB stream. Note that lossless encoding is used for the MSB because its bandwidth falls within an acceptable range, whereas lossy encoding would significantly degrade quality (resulting in a PSNR drop of at least 5 dB). A visual example is shown in Fig.~\ref{fig:codec-visual}.

\begin{table}[t]
\centering
\caption{Effect of token sparsity on inference time under different numbers of input views. Percentages denote the fraction of image tokens retained; all runtimes are reported in ms.}
\label{tab:supp-sparsity-runtime}
\begin{tabular*}{1.0\linewidth}{@{\extracolsep{\fill}}c|ccc|ccc@{}}
\toprule
\multirow{2}{*}{Number of views} & \multicolumn{3}{c|}{Backbone $+$ Decoder} & \multicolumn{3}{c}{Total Inference} \\
& 10\% & 15\% & 20\% & 10\% & 15\% & 20\% \\
\midrule
4-view & 12.1 & 19.3 & 23.3 & 24.5 & 33.7 & 37.3 \\
6-view & 18.1 & 29.3 & 36.6 & 36.6 & 48.6 & 55.8 \\
\bottomrule
\end{tabular*}
\end{table} 

\begin{table}[]
\centering
\caption{Quantitative effects of Gaussian attribute-map quantization and video compression on rendering quality. Metrics are computed over six data segments totaling 4,500 frames.}
\label{tab:metric-quant}
\begin{tabular}{@{}cccc@{}}
\toprule
Process              & PSNR$\uparrow$ & SSIM$\uparrow$ & LPIPS$\downarrow$ \\ \midrule
Raw Result           & 32.6511        &  0.9720      &   0.0426      \\
w/ Quantization      & 31.1007        &  0.9625      &   0.0533   \\
w/ Quant. \& Codec.  & 30.5731        &  0.9574      &   0.0622         \\ \bottomrule
\end{tabular}
\end{table}

\begin{table}[]
\centering
\caption{Rate--distortion trade-off for H.265 compression of Gaussian attribute maps. The MSB stream is encoded losslessly, while the LSB stream is encoded at different quantization parameters (QPs); total bitrate includes both streams. Rendering PSNR is evaluated over six data segments totaling 4,500 frames.}
\label{tab:bitrate-quality}
\begin{tabular}{@{}ccc@{}}
\toprule
Codec Setting (H.265, yuv444p) & Total Bitrate$\downarrow$ & PSNR$\uparrow$ \\ \midrule
LSB QP=24, MSB Lossless      & 120 Mbps      &  31.02         \\
LSB QP=28, MSB Lossless      & 80 Mbps       &  30.57         \\
LSB QP=32, MSB Lossless      & 60 Mbps       &  30.33         \\ 
LSB QP=36, MSB Lossless      & 45 Mbps       &  30.03         \\
LSB QP=40, MSB Lossless      & 40 Mbps       &  29.72         \\ \bottomrule
\end{tabular}
\end{table}

\begin{figure*}[!tp]
    \centering
    \includegraphics[
        width=\textwidth,
        height=0.95\textheight,
        keepaspectratio
    ]{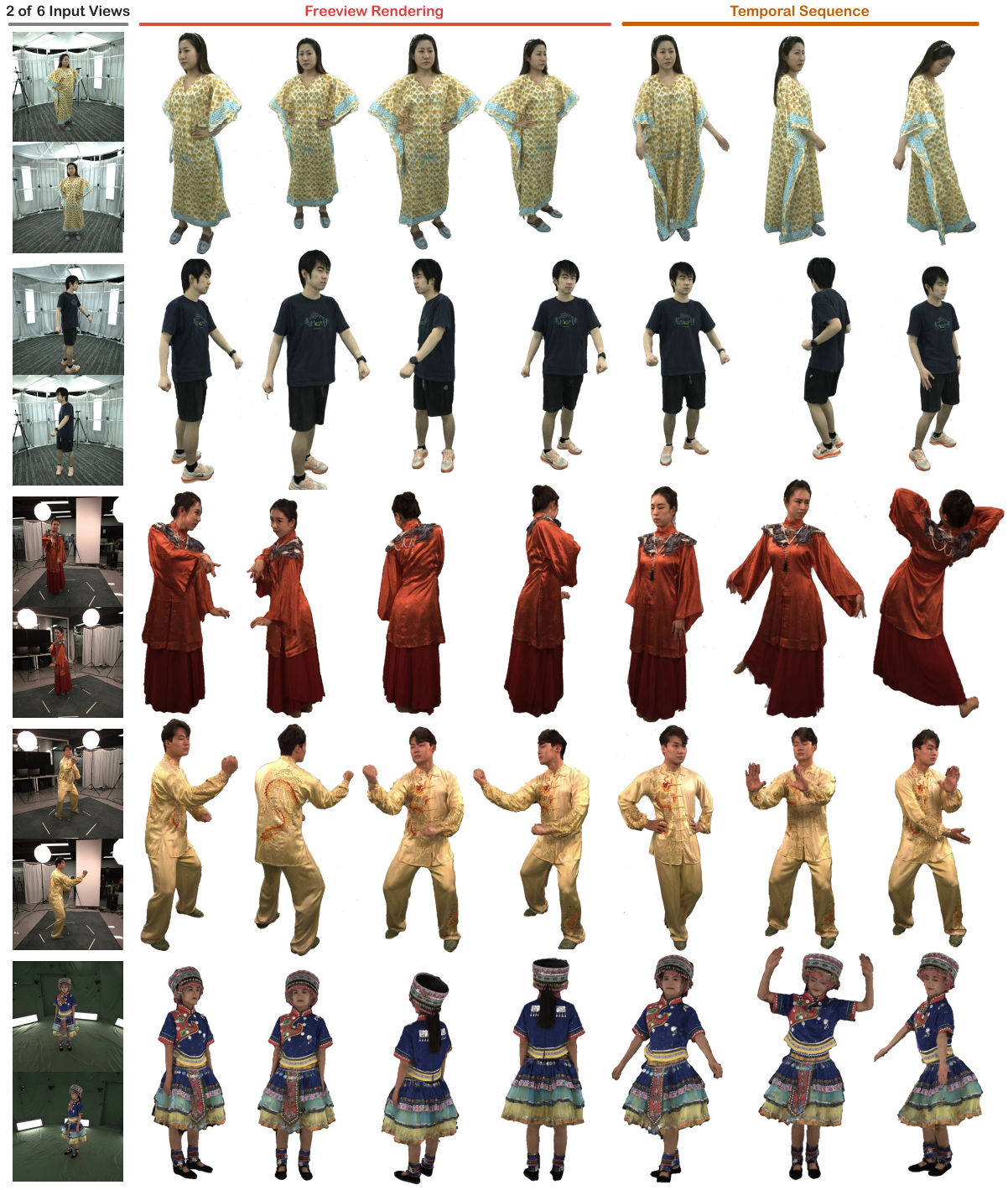}
        \caption{Qualitative results of our method. The first four rows show our collected multi-view videos, and the fifth row shows results on the DNA-Rendering~\cite{cheng2023dna} dataset. Our method is able to recover loose-fitting garments with intricate textures, such as skirts and dresses, as well as a wide variety of body movements. Note that our method requires foreground mattes as additional input, obtained using the open-source Robust Video Matting~\cite{lin2021robust}.}
    \label{fig:comp-in-the-wild}
    \end{figure*}

\subsection{Additional Results on Self-Captured Sequences}
To further demonstrate the generalization capability of our method, we qualitatively evaluate free-viewpoint rendering on self-captured sequences featuring subjects of different genders, diverse clothing, and varied motions.
As shown in Figure~\ref{fig:comp-in-the-wild}, our method still exhibits strong reconstruction capability on diverse self-captured data.

\section{Discussion}
\paragraph{Conclusion.} 
We presented Tele360, the first real-time feed-forward system for dynamic human reconstruction and live free-viewpoint visualization from a sparse set of unposed RGB video streams. Tele360 jointly estimates camera poses and predicts a dynamic 3D Gaussian representation at each time step, enabling 2K input-to-rendering at 25 FPS on a single consumer GPU. The proposed design combines token-level sparsity with global context aggregation, a sparse and efficient transformer-based Gaussian decoder, hybrid CNN–ViT features for high-resolution appearance cues, and Gram-based distillation to transfer multi-view geometry priors. Together with a codec-based streaming pipeline, Tele360 supports 30-FPS interactive client viewing on remote devices, providing a practical step toward deployable human telepresence.

\begin{figure}
    \centering
    \includegraphics[width=1.0\linewidth]{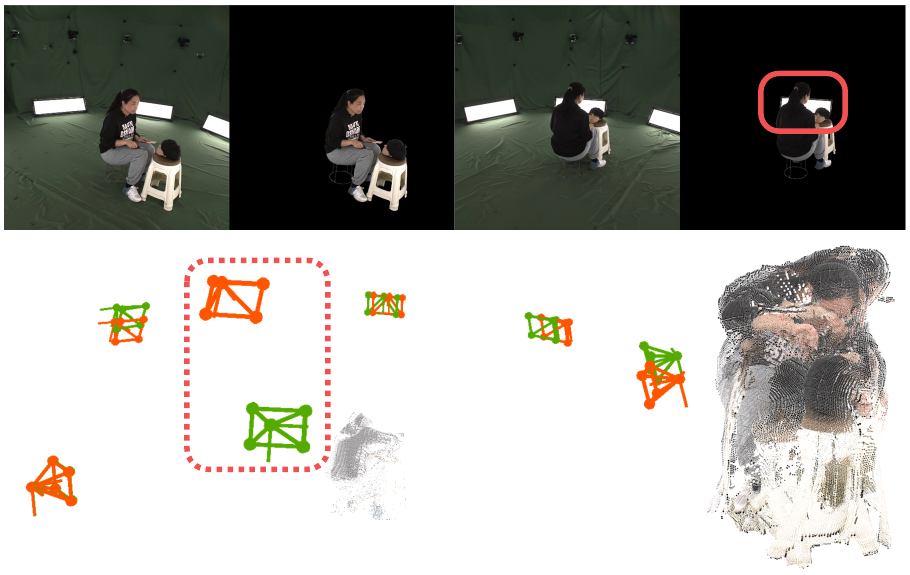}
    \caption{Failure case due to mask inaccuracy. Predicted and ground-truth cameras are marked in red and green, respectively.}
    \label{fig:fail-cam}
\end{figure}

\paragraph{Limitations and Future Work.}
Tele360 relies on foreground masks to enforce token-level sparsity and is therefore sensitive to matting quality, as quantified in Table~\ref{tab:mask-robustness}.
Inaccurate or inconsistent masks can introduce boundary jitter and local artifacts in the predicted Gaussian maps and, in severe cases, degrade camera-pose estimation, as illustrated in Figure~\ref{fig:fail-cam}.
Despite the favorable aggregate VBench scores in Table~\ref{tab:temporal-consistency}, our per-frame formulation does not explicitly enforce temporal coherence; rapid motion, motion blur, or transient matting and pose errors may therefore still produce local flickering.
Moreover, the current appearance representation does not explicitly model view-dependent effects, which may produce inconsistent highlights and artifacts on specular surfaces under novel viewpoints.
Our reconstruction primarily models surfaces supported by the input views, potentially leaving heavily occluded regions incomplete.
Future work will jointly predict foreground segmentation and reconstruction and incorporate temporal aggregation of information across frames, view-dependent appearance modeling, and occlusion-aware priors.

\bibliographystyle{ACM-Reference-Format}
\bibliography{bib}

%%% -*-BibTeX-*-
%%% Do NOT edit. File created by BibTeX with style
%%% ACM-Reference-Format-Journals [18-Jan-2012].

\begin{thebibliography}{97}

%%% ====================================================================
%%% NOTE TO THE USER: you can override these defaults by providing
%%% customized versions of any of these macros before the \bibliography
%%% command.  Each of them MUST provide its own final punctuation,
%%% except for \shownote{}, \showDOI{}, and \showURL{}.  The latter two
%%% do not use final punctuation, in order to avoid confusing it with
%%% the Web address.
%%%
%%% To suppress output of a particular field, define its macro to expand
%%% to an empty string, or better, \unskip, like this:
%%%
%%% \newcommand{\showDOI}[1]{\unskip}   % LaTeX syntax
%%%
%%% \def \showDOI #1{\unskip}           % plain TeX syntax
%%%
%%% ====================================================================

\ifx \showCODEN    \undefined \def \showCODEN     #1{\unskip}     \fi
\ifx \showDOI      \undefined \def \showDOI       #1{#1}\fi
\ifx \showISBNx    \undefined \def \showISBNx     #1{\unskip}     \fi
\ifx \showISBNxiii \undefined \def \showISBNxiii  #1{\unskip}     \fi
\ifx \showISSN     \undefined \def \showISSN      #1{\unskip}     \fi
\ifx \showLCCN     \undefined \def \showLCCN      #1{\unskip}     \fi
\ifx \shownote     \undefined \def \shownote      #1{#1}          \fi
\ifx \showarticletitle \undefined \def \showarticletitle #1{#1}   \fi
\ifx \showURL      \undefined \def \showURL       {\relax}        \fi
% The following commands are used for tagged output and should be
% invisible to TeX
\providecommand\bibfield[2]{#2}
\providecommand\bibinfo[2]{#2}
\providecommand\natexlab[1]{#1}
\providecommand\showeprint[2][]{arXiv:#2}

\bibitem[Bae et~al\mbox{.}(2025)]%
        {bae2024per}
\bibfield{author}{\bibinfo{person}{Jeongmin Bae}, \bibinfo{person}{Seoha Kim},
  \bibinfo{person}{Youngsik Yun}, \bibinfo{person}{Hahyun Lee},
  \bibinfo{person}{Gun Bang}, {and} \bibinfo{person}{Youngjung Uh}.}
  \bibinfo{year}{2025}\natexlab{}.
\newblock \showarticletitle{{Per-Gaussian Embedding-Based Deformation for
  Deformable 3D Gaussian Splatting}}. In \bibinfo{booktitle}{\emph{Computer
  Vision -- ECCV 2024}} \emph{(\bibinfo{series}{Lecture Notes in Computer
  Science}, Vol.~\bibinfo{volume}{15073})}. \bibinfo{publisher}{Springer Nature
  Switzerland}, \bibinfo{pages}{321--335}.
\newblock
\urldef\tempurl%
\url{https://doi.org/10.1007/978-3-031-72633-0_18}
\showDOI{\tempurl}


\bibitem[Bo{\v z}i{\v c} et~al\mbox{.}(2020)]%
        {bozic2020deepdeform}
\bibfield{author}{\bibinfo{person}{Alja{\v z} Bo{\v z}i{\v c}},
  \bibinfo{person}{Michael Zollh{\"o}fer}, \bibinfo{person}{Christian
  Theobalt}, {and} \bibinfo{person}{Matthias Nie{\ss}ner}.}
  \bibinfo{year}{2020}\natexlab{}.
\newblock \showarticletitle{{DeepDeform: Learning Non-Rigid RGB-D
  Reconstruction With Semi-Supervised Data}}. In
  \bibinfo{booktitle}{\emph{Proceedings of the IEEE/CVF Conference on Computer
  Vision and Pattern Recognition}}. \bibinfo{pages}{7002--7012}.
\newblock


\bibitem[Buehler et~al\mbox{.}(2001)]%
        {buehler2001unstructured}
\bibfield{author}{\bibinfo{person}{Chris Buehler}, \bibinfo{person}{Michael
  Bosse}, \bibinfo{person}{Leonard McMillan}, \bibinfo{person}{Steven Gortler},
  {and} \bibinfo{person}{Michael Cohen}.} \bibinfo{year}{2001}\natexlab{}.
\newblock \showarticletitle{{Unstructured Lumigraph Rendering}}. In
  \bibinfo{booktitle}{\emph{Proceedings of the 28th Annual Conference on
  Computer Graphics and Interactive Techniques}}. \bibinfo{pages}{425--432}.
\newblock
\urldef\tempurl%
\url{https://doi.org/10.1145/383259.383309}
\showDOI{\tempurl}


\bibitem[Carion et~al\mbox{.}(2026)]%
        {carion2026sam3}
\bibfield{author}{\bibinfo{person}{Nicolas Carion}, \bibinfo{person}{Laura
  Gustafson}, \bibinfo{person}{Yuan-Ting Hu}, \bibinfo{person}{Shoubhik
  Debnath}, \bibinfo{person}{Ronghang Hu}, \bibinfo{person}{Didac Suris
  Coll-Vinent}, \bibinfo{person}{Chaitanya Ryali},
  \bibinfo{person}{Kalyan~Vasudev Alwala}, \bibinfo{person}{Haitham Khedr},
  \bibinfo{person}{Andrew Huang}, \bibinfo{person}{Jie Lei},
  \bibinfo{person}{Tengyu Ma}, \bibinfo{person}{Baishan Guo},
  \bibinfo{person}{Arpit Kalla}, \bibinfo{person}{Markus Marks},
  \bibinfo{person}{Joseph Greer}, \bibinfo{person}{Meng Wang},
  \bibinfo{person}{Peize Sun}, \bibinfo{person}{Roman R{\"a}dle},
  \bibinfo{person}{Triantafyllos Afouras}, \bibinfo{person}{Effrosyni
  Mavroudi}, \bibinfo{person}{Katherine Xu}, \bibinfo{person}{Tsung-Han Wu},
  \bibinfo{person}{Yu Zhou}, \bibinfo{person}{Liliane Momeni},
  \bibinfo{person}{Rishi Hazra}, \bibinfo{person}{Shuangrui Ding},
  \bibinfo{person}{Sagar Vaze}, \bibinfo{person}{Francois Porcher},
  \bibinfo{person}{Feng Li}, \bibinfo{person}{Siyuan Li},
  \bibinfo{person}{Aishwarya Kamath}, \bibinfo{person}{Ho~Kei Cheng},
  \bibinfo{person}{Piotr Doll{\'a}r}, \bibinfo{person}{Nikhila Ravi},
  \bibinfo{person}{Kate Saenko}, \bibinfo{person}{Pengchuan Zhang}, {and}
  \bibinfo{person}{Christoph Feichtenhofer}.} \bibinfo{year}{2026}\natexlab{}.
\newblock \showarticletitle{SAM 3: Segment Anything with Concepts}. In
  \bibinfo{booktitle}{\emph{International Conference on Learning
  Representations (ICLR)}}, Vol.~\bibinfo{volume}{2026}.
  \bibinfo{pages}{138846--138923}.
\newblock
\urldef\tempurl%
\url{https://proceedings.iclr.cc/paper_files/paper/2026/hash/e0982cbc81401df3430ee1ff780dc7a2-Abstract-Conference.html}
\showURL{%
\tempurl}


\bibitem[Chen et~al\mbox{.}(2025a)]%
        {chen2025taoavatar}
\bibfield{author}{\bibinfo{person}{Jianchuan Chen}, \bibinfo{person}{Jingchuan
  Hu}, \bibinfo{person}{Gaige Wang}, \bibinfo{person}{Zhonghua Jiang},
  \bibinfo{person}{Tiansong Zhou}, \bibinfo{person}{Zhiwen Chen}, {and}
  \bibinfo{person}{Chengfei Lv}.} \bibinfo{year}{2025}\natexlab{a}.
\newblock \showarticletitle{{TaoAvatar: Real-Time Lifelike Full-Body Talking
  Avatars for Augmented Reality via 3D Gaussian Splatting}}. In
  \bibinfo{booktitle}{\emph{Proceedings of the Computer Vision and Pattern
  Recognition Conference}}. \bibinfo{pages}{10723--10734}.
\newblock


\bibitem[Chen et~al\mbox{.}(2024)]%
        {chen2024monogaussianavatar}
\bibfield{author}{\bibinfo{person}{Yufan Chen}, \bibinfo{person}{Lizhen Wang},
  \bibinfo{person}{Qijing Li}, \bibinfo{person}{Hongjiang Xiao},
  \bibinfo{person}{Shengping Zhang}, \bibinfo{person}{Hongxun Yao}, {and}
  \bibinfo{person}{Yebin Liu}.} \bibinfo{year}{2024}\natexlab{}.
\newblock \showarticletitle{{MonoGaussianAvatar: Monocular Gaussian Point-based
  Head Avatar}}. In \bibinfo{booktitle}{\emph{ACM SIGGRAPH 2024 Conference
  Papers}}. Article \bibinfo{articleno}{58}, \bibinfo{numpages}{9}~pages.
\newblock
\urldef\tempurl%
\url{https://doi.org/10.1145/3641519.3657499}
\showDOI{\tempurl}


\bibitem[Chen et~al\mbox{.}(2025b)]%
        {chen2024meshavatar}
\bibfield{author}{\bibinfo{person}{Yushuo Chen}, \bibinfo{person}{Zerong
  Zheng}, \bibinfo{person}{Zhe Li}, \bibinfo{person}{Chao Xu}, {and}
  \bibinfo{person}{Yebin Liu}.} \bibinfo{year}{2025}\natexlab{b}.
\newblock \showarticletitle{{MeshAvatar: Learning High-Quality Triangular Human
  Avatars from Multi-view Videos}}. In \bibinfo{booktitle}{\emph{Computer
  Vision -- ECCV 2024}} \emph{(\bibinfo{series}{Lecture Notes in Computer
  Science}, Vol.~\bibinfo{volume}{15126})}. \bibinfo{publisher}{Springer Nature
  Switzerland}, \bibinfo{pages}{250--269}.
\newblock
\urldef\tempurl%
\url{https://doi.org/10.1007/978-3-031-73113-6_15}
\showDOI{\tempurl}


\bibitem[Cheng et~al\mbox{.}(2023)]%
        {cheng2023dna}
\bibfield{author}{\bibinfo{person}{Wei Cheng}, \bibinfo{person}{Ruixiang Chen},
  \bibinfo{person}{Siming Fan}, \bibinfo{person}{Wanqi Yin},
  \bibinfo{person}{Keyu Chen}, \bibinfo{person}{Zhongang Cai},
  \bibinfo{person}{Jingbo Wang}, \bibinfo{person}{Yang Gao},
  \bibinfo{person}{Zhengming Yu}, \bibinfo{person}{Zhengyu Lin},
  \bibinfo{person}{Daxuan Ren}, \bibinfo{person}{Lei Yang},
  \bibinfo{person}{Ziwei Liu}, \bibinfo{person}{Chen~Change Loy},
  \bibinfo{person}{Chen Qian}, \bibinfo{person}{Wayne Wu},
  \bibinfo{person}{Dahua Lin}, \bibinfo{person}{Bo Dai}, {and}
  \bibinfo{person}{Kwan-Yee Lin}.} \bibinfo{year}{2023}\natexlab{}.
\newblock \showarticletitle{{DNA-Rendering: A Diverse Neural Actor Repository
  for High-Fidelity Human-Centric Rendering}}. In
  \bibinfo{booktitle}{\emph{Proceedings of the IEEE/CVF International
  Conference on Computer Vision}}. \bibinfo{pages}{19982--19993}.
\newblock


\bibitem[Dai et~al\mbox{.}(2025)]%
        {dai20254d}
\bibfield{author}{\bibinfo{person}{Pinxuan Dai}, \bibinfo{person}{Peiquan
  Zhang}, \bibinfo{person}{Zheng Dong}, \bibinfo{person}{Ke Xu},
  \bibinfo{person}{Yifan Peng}, \bibinfo{person}{Dandan Ding},
  \bibinfo{person}{Yujun Shen}, \bibinfo{person}{Yin Yang},
  \bibinfo{person}{Xinguo Liu}, \bibinfo{person}{Rynson W.~H. Lau}, {and}
  \bibinfo{person}{Weiwei Xu}.} \bibinfo{year}{2025}\natexlab{}.
\newblock \showarticletitle{{4D Gaussian Videos with Motion Layering}}.
\newblock \bibinfo{journal}{\emph{ACM Transactions on Graphics}}
  \bibinfo{volume}{44}, \bibinfo{number}{4}, Article \bibinfo{articleno}{124}
  (\bibinfo{year}{2025}), \bibinfo{numpages}{14}~pages.
\newblock
\urldef\tempurl%
\url{https://doi.org/10.1145/3731189}
\showDOI{\tempurl}


\bibitem[Dao(2024)]%
        {dao2023flashattention2}
\bibfield{author}{\bibinfo{person}{Tri Dao}.} \bibinfo{year}{2024}\natexlab{}.
\newblock \showarticletitle{Flash{A}ttention-2: Faster Attention with Better
  Parallelism and Work Partitioning}. In
  \bibinfo{booktitle}{\emph{International Conference on Learning
  Representations (ICLR)}}, Vol.~\bibinfo{volume}{2024}.
  \bibinfo{pages}{35549--35562}.
\newblock


\bibitem[Dosovitskiy et~al\mbox{.}(2021)]%
        {vit}
\bibfield{author}{\bibinfo{person}{Alexey Dosovitskiy}, \bibinfo{person}{Lucas
  Beyer}, \bibinfo{person}{Alexander Kolesnikov}, \bibinfo{person}{Dirk
  Weissenborn}, \bibinfo{person}{Xiaohua Zhai}, \bibinfo{person}{Thomas
  Unterthiner}, \bibinfo{person}{Mostafa Dehghani}, \bibinfo{person}{Matthias
  Minderer}, \bibinfo{person}{Georg Heigold}, \bibinfo{person}{Sylvain Gelly},
  \bibinfo{person}{Jakob Uszkoreit}, {and} \bibinfo{person}{Neil Houlsby}.}
  \bibinfo{year}{2021}\natexlab{}.
\newblock \showarticletitle{{An Image is Worth 16x16 Words: Transformers for
  Image Recognition at Scale}}. In \bibinfo{booktitle}{\emph{International
  Conference on Learning Representations (ICLR)}}.
\newblock
\urldef\tempurl%
\url{https://openreview.net/forum?id=YicbFdNTTy}
\showURL{%
\tempurl}


\bibitem[Dou et~al\mbox{.}(2017)]%
        {dou2017motion2fusion}
\bibfield{author}{\bibinfo{person}{Mingsong Dou}, \bibinfo{person}{Philip
  Davidson}, \bibinfo{person}{Sean~Ryan Fanello}, \bibinfo{person}{Sameh
  Khamis}, \bibinfo{person}{Adarsh Kowdle}, \bibinfo{person}{Christoph
  Rhemann}, \bibinfo{person}{Vladimir Tankovich}, {and}
  \bibinfo{person}{Shahram Izadi}.} \bibinfo{year}{2017}\natexlab{}.
\newblock \showarticletitle{{Motion2Fusion: real-time volumetric performance
  capture}}.
\newblock \bibinfo{journal}{\emph{ACM Transactions on Graphics}}
  \bibinfo{volume}{36}, \bibinfo{number}{6}, Article \bibinfo{articleno}{246}
  (\bibinfo{year}{2017}), \bibinfo{numpages}{16}~pages.
\newblock
\urldef\tempurl%
\url{https://doi.org/10.1145/3130800.3130801}
\showDOI{\tempurl}


\bibitem[Dou et~al\mbox{.}(2016)]%
        {dou2016fusion4d}
\bibfield{author}{\bibinfo{person}{Mingsong Dou}, \bibinfo{person}{Sameh
  Khamis}, \bibinfo{person}{Yury Degtyarev}, \bibinfo{person}{Philip Davidson},
  \bibinfo{person}{Sean~Ryan Fanello}, \bibinfo{person}{Adarsh Kowdle},
  \bibinfo{person}{Sergio Orts-Escolano}, \bibinfo{person}{Christoph Rhemann},
  \bibinfo{person}{David Kim}, \bibinfo{person}{Jonathan Taylor},
  \bibinfo{person}{Pushmeet Kohli}, \bibinfo{person}{Vladimir Tankovich}, {and}
  \bibinfo{person}{Shahram Izadi}.} \bibinfo{year}{2016}\natexlab{}.
\newblock \showarticletitle{{Fusion4D: real-time performance capture of
  challenging scenes}}.
\newblock \bibinfo{journal}{\emph{ACM Transactions on Graphics}}
  \bibinfo{volume}{35}, \bibinfo{number}{4}, Article \bibinfo{articleno}{114}
  (\bibinfo{year}{2016}), \bibinfo{numpages}{13}~pages.
\newblock
\urldef\tempurl%
\url{https://doi.org/10.1145/2897824.2925969}
\showDOI{\tempurl}


\bibitem[Duan et~al\mbox{.}(2024)]%
        {duan20244d}
\bibfield{author}{\bibinfo{person}{Yuanxing Duan}, \bibinfo{person}{Fangyin
  Wei}, \bibinfo{person}{Qiyu Dai}, \bibinfo{person}{Yuhang He},
  \bibinfo{person}{Wenzheng Chen}, {and} \bibinfo{person}{Baoquan Chen}.}
  \bibinfo{year}{2024}\natexlab{}.
\newblock \showarticletitle{{4D-Rotor Gaussian Splatting: Towards Efficient
  Novel View Synthesis for Dynamic Scenes}}. In \bibinfo{booktitle}{\emph{ACM
  SIGGRAPH 2024 Conference Papers}}. Article \bibinfo{articleno}{87},
  \bibinfo{numpages}{11}~pages.
\newblock
\urldef\tempurl%
\url{https://doi.org/10.1145/3641519.3657463}
\showDOI{\tempurl}


\bibitem[Gao et~al\mbox{.}(2025)]%
        {gao20257dgs}
\bibfield{author}{\bibinfo{person}{Zhongpai Gao}, \bibinfo{person}{Benjamin
  Planche}, \bibinfo{person}{Meng Zheng}, \bibinfo{person}{Anwesa Choudhuri},
  \bibinfo{person}{Terrence Chen}, {and} \bibinfo{person}{Ziyan Wu}.}
  \bibinfo{year}{2025}\natexlab{}.
\newblock \showarticletitle{7DGS: Unified Spatial-Temporal-Angular Gaussian
  Splatting}. In \bibinfo{booktitle}{\emph{Proceedings of the IEEE/CVF
  International Conference on Computer Vision (ICCV)}}.
  \bibinfo{pages}{26316--26325}.
\newblock


\bibitem[Gatys et~al\mbox{.}(2016)]%
        {gramloss}
\bibfield{author}{\bibinfo{person}{Leon~A Gatys}, \bibinfo{person}{Alexander~S
  Ecker}, {and} \bibinfo{person}{Matthias Bethge}.}
  \bibinfo{year}{2016}\natexlab{}.
\newblock \showarticletitle{{Image Style Transfer Using Convolutional Neural
  Networks}}. In \bibinfo{booktitle}{\emph{Proceedings of the IEEE conference
  on computer vision and pattern recognition}}. \bibinfo{pages}{2414--2423}.
\newblock


\bibitem[Guan et~al\mbox{.}(2023)]%
        {guan2023metastream}
\bibfield{author}{\bibinfo{person}{Yongjie Guan}, \bibinfo{person}{Xueyu Hou},
  \bibinfo{person}{Nan Wu}, \bibinfo{person}{Bo Han}, {and}
  \bibinfo{person}{Tao Han}.} \bibinfo{year}{2023}\natexlab{}.
\newblock \showarticletitle{{MetaStream: Live Volumetric Content Capture,
  Creation, Delivery, and Rendering in Real Time}}. In
  \bibinfo{booktitle}{\emph{Proceedings of the 29th annual international
  conference on mobile computing and networking}}. Article
  \bibinfo{articleno}{29}, \bibinfo{numpages}{15}~pages.
\newblock
\urldef\tempurl%
\url{https://doi.org/10.1145/3570361.3592530}
\showDOI{\tempurl}


\bibitem[Guo et~al\mbox{.}(2025a)]%
        {guo2025vid2avatarpro}
\bibfield{author}{\bibinfo{person}{Chen Guo}, \bibinfo{person}{Junxuan Li},
  \bibinfo{person}{Yash Kant}, \bibinfo{person}{Yaser Sheikh},
  \bibinfo{person}{Shunsuke Saito}, {and} \bibinfo{person}{Chen Cao}.}
  \bibinfo{year}{2025}\natexlab{a}.
\newblock \showarticletitle{{Vid2Avatar-Pro: Authentic Avatar from Videos in
  the Wild via Universal Prior}}. In \bibinfo{booktitle}{\emph{Proceedings of
  the IEEE/CVF Conference on Computer Vision and Pattern Recognition}}.
  \bibinfo{pages}{5559--5570}.
\newblock


\bibitem[Guo et~al\mbox{.}(2025b)]%
        {guo2024motion}
\bibfield{author}{\bibinfo{person}{Zhiyang Guo}, \bibinfo{person}{Wengang
  Zhou}, \bibinfo{person}{Li Li}, \bibinfo{person}{Min Wang}, {and}
  \bibinfo{person}{Houqiang Li}.} \bibinfo{year}{2025}\natexlab{b}.
\newblock \showarticletitle{{Motion-Aware 3D Gaussian Splatting for Efficient
  Dynamic Scene Reconstruction}}.
\newblock \bibinfo{journal}{\emph{IEEE Transactions on Circuits and Systems for
  Video Technology}} \bibinfo{volume}{35}, \bibinfo{number}{4}
  (\bibinfo{date}{April} \bibinfo{year}{2025}), \bibinfo{pages}{3119--3133}.
\newblock
\urldef\tempurl%
\url{https://doi.org/10.1109/TCSVT.2024.3502257}
\showDOI{\tempurl}


\bibitem[Han et~al\mbox{.}(2023)]%
        {han2023high}
\bibfield{author}{\bibinfo{person}{Sang-Hun Han}, \bibinfo{person}{Min-Gyu
  Park}, \bibinfo{person}{Ju~Hong Yoon}, \bibinfo{person}{Ju-Mi Kang},
  \bibinfo{person}{Young-Jae Park}, {and} \bibinfo{person}{Hae-Gon Jeon}.}
  \bibinfo{year}{2023}\natexlab{}.
\newblock \showarticletitle{{High-Fidelity 3D Human Digitization From Single 2K
  Resolution Images}}. In \bibinfo{booktitle}{\emph{Proceedings of the IEEE/CVF
  Conference on Computer Vision and Pattern Recognition (CVPR)}}.
  \bibinfo{pages}{12869--12879}.
\newblock


\bibitem[Henry et~al\mbox{.}(2020)]%
        {henry2020query}
\bibfield{author}{\bibinfo{person}{Alex Henry}, \bibinfo{person}{Prudhvi~Raj
  Dachapally}, \bibinfo{person}{Shubham~Shantaram Pawar}, {and}
  \bibinfo{person}{Yuxuan Chen}.} \bibinfo{year}{2020}\natexlab{}.
\newblock \showarticletitle{{Query-Key Normalization for Transformers}}. In
  \bibinfo{booktitle}{\emph{Findings of the Association for Computational
  Linguistics: EMNLP 2020}}. \bibinfo{pages}{4246--4253}.
\newblock
\urldef\tempurl%
\url{https://doi.org/10.18653/v1/2020.findings-emnlp.379}
\showDOI{\tempurl}


\bibitem[Hu et~al\mbox{.}(2024)]%
        {hu2024gaussianavatar}
\bibfield{author}{\bibinfo{person}{Liangxiao Hu}, \bibinfo{person}{Hongwen
  Zhang}, \bibinfo{person}{Yuxiang Zhang}, \bibinfo{person}{Boyao Zhou},
  \bibinfo{person}{Boning Liu}, \bibinfo{person}{Shengping Zhang}, {and}
  \bibinfo{person}{Liqiang Nie}.} \bibinfo{year}{2024}\natexlab{}.
\newblock \showarticletitle{{GaussianAvatar: Towards Realistic Human Avatar
  Modeling from a Single Video via Animatable 3D Gaussians}}. In
  \bibinfo{booktitle}{\emph{Proceedings of the IEEE/CVF Conference on Computer
  Vision and Pattern Recognition (CVPR)}}. \bibinfo{pages}{634--644}.
\newblock


\bibitem[Huang et~al\mbox{.}(2024)]%
        {huang2024vbench}
\bibfield{author}{\bibinfo{person}{Ziqi Huang}, \bibinfo{person}{Yinan He},
  \bibinfo{person}{Jiashuo Yu}, \bibinfo{person}{Fan Zhang},
  \bibinfo{person}{Chenyang Si}, \bibinfo{person}{Yuming Jiang},
  \bibinfo{person}{Yuanhan Zhang}, \bibinfo{person}{Tianxing Wu},
  \bibinfo{person}{Qingyang Jin}, \bibinfo{person}{Nattapol Chanpaisit},
  \bibinfo{person}{Yaohui Wang}, \bibinfo{person}{Xinyuan Chen},
  \bibinfo{person}{Limin Wang}, \bibinfo{person}{Dahua Lin},
  \bibinfo{person}{Yu Qiao}, {and} \bibinfo{person}{Ziwei Liu}.}
  \bibinfo{year}{2024}\natexlab{}.
\newblock \showarticletitle{{VBench: Comprehensive Benchmark Suite for Video
  Generative Models}}. In \bibinfo{booktitle}{\emph{Proceedings of the IEEE/CVF
  Conference on Computer Vision and Pattern Recognition (CVPR)}}.
  \bibinfo{pages}{21807--21818}.
\newblock


\bibitem[I{\c{s}}{\i}k et~al\mbox{.}(2023)]%
        {icsik2023humanrf}
\bibfield{author}{\bibinfo{person}{Mustafa I{\c{s}}{\i}k},
  \bibinfo{person}{Martin R{\"u}nz}, \bibinfo{person}{Markos Georgopoulos},
  \bibinfo{person}{Taras Khakhulin}, \bibinfo{person}{Jonathan Starck},
  \bibinfo{person}{Lourdes Agapito}, {and} \bibinfo{person}{Matthias
  Nie{\ss}ner}.} \bibinfo{year}{2023}\natexlab{}.
\newblock \showarticletitle{{HumanRF: High-Fidelity Neural Radiance Fields for
  Humans in Motion}}.
\newblock \bibinfo{journal}{\emph{ACM Transactions on Graphics}}
  \bibinfo{volume}{42}, \bibinfo{number}{4}, Article \bibinfo{articleno}{160}
  (\bibinfo{year}{2023}), \bibinfo{numpages}{12}~pages.
\newblock
\urldef\tempurl%
\url{https://doi.org/10.1145/3592415}
\showDOI{\tempurl}


\bibitem[Jiang et~al\mbox{.}(2025)]%
        {anysplat}
\bibfield{author}{\bibinfo{person}{Lihan Jiang}, \bibinfo{person}{Yucheng Mao},
  \bibinfo{person}{Linning Xu}, \bibinfo{person}{Tao Lu},
  \bibinfo{person}{Kerui Ren}, \bibinfo{person}{Yichen Jin},
  \bibinfo{person}{Xudong Xu}, \bibinfo{person}{Mulin Yu},
  \bibinfo{person}{Jiangmiao Pang}, \bibinfo{person}{Feng Zhao},
  \bibinfo{person}{Dahua Lin}, {and} \bibinfo{person}{Bo Dai}.}
  \bibinfo{year}{2025}\natexlab{}.
\newblock \showarticletitle{{AnySplat: Feed-forward 3D Gaussian Splatting from
  Unconstrained Views}}.
\newblock \bibinfo{journal}{\emph{ACM Transactions on Graphics}}
  \bibinfo{volume}{44}, \bibinfo{number}{6}, Article \bibinfo{articleno}{257}
  (\bibinfo{year}{2025}), \bibinfo{numpages}{16}~pages.
\newblock
\urldef\tempurl%
\url{https://doi.org/10.1145/3763326}
\showDOI{\tempurl}


\bibitem[Jiang et~al\mbox{.}(2026)]%
        {jiang2026hireff}
\bibfield{author}{\bibinfo{person}{Yiming Jiang}, \bibinfo{person}{Hanzhang
  Tu}, \bibinfo{person}{Wenfeng Song}, \bibinfo{person}{Siyou Lin},
  \bibinfo{person}{Liang An}, \bibinfo{person}{Shuai Li},
  \bibinfo{person}{Aimin Hao}, {and} \bibinfo{person}{Yebin Liu}.}
  \bibinfo{year}{2026}\natexlab{}.
\newblock \showarticletitle{{HiReFF: High-Resolution Feedforward Human
  Reconstruction from Uncalibrated Sparse-View Video}}.
\newblock \bibinfo{journal}{\emph{arXiv preprint arXiv:2606.29333}}
  (\bibinfo{year}{2026}).
\newblock
\showeprint[arxiv]{2606.29333}~[cs.CV]
\urldef\tempurl%
\url{https://arxiv.org/abs/2606.29333}
\showURL{%
\tempurl}


\bibitem[Keetha et~al\mbox{.}(2026)]%
        {keetha2025mapanything}
\bibfield{author}{\bibinfo{person}{Nikhil~Varma Keetha},
  \bibinfo{person}{Norman M{\"u}ller}, \bibinfo{person}{Johannes
  Sch{\"o}nberger}, \bibinfo{person}{Lorenzo Porzi}, \bibinfo{person}{Yuchen
  Zhang}, \bibinfo{person}{Tobias Fischer}, \bibinfo{person}{Arno Knapitsch},
  \bibinfo{person}{Duncan Zauss}, \bibinfo{person}{Ethan Weber},
  \bibinfo{person}{Nelson Antunes}, \bibinfo{person}{Jonathon Luiten},
  \bibinfo{person}{Manuel Lopez-Antequera}, \bibinfo{person}{Samuel
  Rota~Bul{\`o}}, \bibinfo{person}{Christian Richardt}, \bibinfo{person}{Deva
  Ramanan}, \bibinfo{person}{Sebastian Scherer}, {and} \bibinfo{person}{Peter
  Kontschieder}.} \bibinfo{year}{2026}\natexlab{}.
\newblock \showarticletitle{MapAnything: Universal Feed-Forward Metric 3D
  Reconstruction}. In \bibinfo{booktitle}{\emph{International Conference on 3D
  Vision (3DV)}}. IEEE, \bibinfo{pages}{499--509}.
\newblock
\urldef\tempurl%
\url{https://doi.org/10.1109/3DV69130.2026.00054}
\showDOI{\tempurl}


\bibitem[Kerbl et~al\mbox{.}(2023)]%
        {3dgs}
\bibfield{author}{\bibinfo{person}{Bernhard Kerbl}, \bibinfo{person}{Georgios
  Kopanas}, \bibinfo{person}{Thomas Leimk{\"u}hler}, {and}
  \bibinfo{person}{George Drettakis}.} \bibinfo{year}{2023}\natexlab{}.
\newblock \showarticletitle{3D Gaussian Splatting for Real-Time Radiance Field
  Rendering}.
\newblock \bibinfo{journal}{\emph{ACM Transactions on Graphics}}
  \bibinfo{volume}{42}, \bibinfo{number}{4}, Article \bibinfo{articleno}{139}
  (\bibinfo{year}{2023}), \bibinfo{numpages}{14}~pages.
\newblock
\urldef\tempurl%
\url{https://doi.org/10.1145/3592433}
\showDOI{\tempurl}


\bibitem[Kim et~al\mbox{.}(2024)]%
        {kim20244d}
\bibfield{author}{\bibinfo{person}{Mijeong Kim}, \bibinfo{person}{Jongwoo Lim},
  {and} \bibinfo{person}{Bohyung Han}.} \bibinfo{year}{2024}\natexlab{}.
\newblock \showarticletitle{4D Gaussian Splatting in the Wild with
  Uncertainty-Aware Regularization}. In \bibinfo{booktitle}{\emph{Advances in
  Neural Information Processing Systems}}, Vol.~\bibinfo{volume}{37}.
  \bibinfo{pages}{129209--129226}.
\newblock
\urldef\tempurl%
\url{https://doi.org/10.52202/079017-4104}
\showDOI{\tempurl}


\bibitem[Kwon et~al\mbox{.}(2025)]%
        {kwon2024generalizable}
\bibfield{author}{\bibinfo{person}{Youngjoong Kwon}, \bibinfo{person}{Baole
  Fang}, \bibinfo{person}{Yixing Lu}, \bibinfo{person}{Haoye Dong},
  \bibinfo{person}{Cheng Zhang}, \bibinfo{person}{Francisco~Vicente Carrasco},
  \bibinfo{person}{Albert Mosella-Montoro}, \bibinfo{person}{Jianjin Xu},
  \bibinfo{person}{Shingo Takagi}, \bibinfo{person}{Daeil Kim},
  \bibinfo{person}{Aayush Prakash}, {and} \bibinfo{person}{Fernando De~la
  Torre}.} \bibinfo{year}{2025}\natexlab{}.
\newblock \showarticletitle{{Generalizable Human Gaussians for Sparse View
  Synthesis}}. In \bibinfo{booktitle}{\emph{Computer Vision -- ECCV 2024}}
  \emph{(\bibinfo{series}{Lecture Notes in Computer Science},
  Vol.~\bibinfo{volume}{15136})}. \bibinfo{publisher}{Springer Nature
  Switzerland}, \bibinfo{pages}{451--468}.
\newblock
\urldef\tempurl%
\url{https://doi.org/10.1007/978-3-031-73229-4_26}
\showDOI{\tempurl}


\bibitem[Labe et~al\mbox{.}(2025)]%
        {labe2024dgd}
\bibfield{author}{\bibinfo{person}{Isaac Labe}, \bibinfo{person}{Noam
  Issachar}, \bibinfo{person}{Itai Lang}, {and} \bibinfo{person}{Sagie
  Benaim}.} \bibinfo{year}{2025}\natexlab{}.
\newblock \showarticletitle{{DGD: Dynamic 3D Gaussians Distillation}}. In
  \bibinfo{booktitle}{\emph{Computer Vision -- ECCV 2024}}
  \emph{(\bibinfo{series}{Lecture Notes in Computer Science},
  Vol.~\bibinfo{volume}{15126})}. \bibinfo{publisher}{Springer Nature
  Switzerland}, \bibinfo{pages}{361--378}.
\newblock
\urldef\tempurl%
\url{https://doi.org/10.1007/978-3-031-73113-6_21}
\showDOI{\tempurl}


\bibitem[Lee et~al\mbox{.}(2024)]%
        {lee2024fully}
\bibfield{author}{\bibinfo{person}{Junoh Lee}, \bibinfo{person}{Changyeon Won},
  \bibinfo{person}{Hyunjun Jung}, \bibinfo{person}{Inhwan Bae}, {and}
  \bibinfo{person}{Hae-Gon Jeon}.} \bibinfo{year}{2024}\natexlab{}.
\newblock \showarticletitle{Fully Explicit Dynamic Gaussian Splatting}. In
  \bibinfo{booktitle}{\emph{Advances in Neural Information Processing
  Systems}}, Vol.~\bibinfo{volume}{37}. \bibinfo{pages}{5384--5409}.
\newblock
\urldef\tempurl%
\url{https://doi.org/10.52202/079017-0174}
\showDOI{\tempurl}


\bibitem[Leroy et~al\mbox{.}(2025)]%
        {mast3r}
\bibfield{author}{\bibinfo{person}{Vincent Leroy}, \bibinfo{person}{Yohann
  Cabon}, {and} \bibinfo{person}{J{\'e}r{\^o}me Revaud}.}
  \bibinfo{year}{2025}\natexlab{}.
\newblock \showarticletitle{{Grounding Image Matching in 3D with MASt3R}}. In
  \bibinfo{booktitle}{\emph{Computer Vision -- ECCV 2024}}
  \emph{(\bibinfo{series}{Lecture Notes in Computer Science},
  Vol.~\bibinfo{volume}{15130})}. \bibinfo{publisher}{Springer Nature
  Switzerland}, \bibinfo{pages}{71--91}.
\newblock
\urldef\tempurl%
\url{https://doi.org/10.1007/978-3-031-73220-1_5}
\showDOI{\tempurl}


\bibitem[Li et~al\mbox{.}(2022)]%
        {li2022streaming}
\bibfield{author}{\bibinfo{person}{Lingzhi Li}, \bibinfo{person}{Zhen Shen},
  \bibinfo{person}{Zhongshu Wang}, \bibinfo{person}{Li Shen}, {and}
  \bibinfo{person}{Ping Tan}.} \bibinfo{year}{2022}\natexlab{}.
\newblock \showarticletitle{Streaming Radiance Fields for 3D Video Synthesis}.
  In \bibinfo{booktitle}{\emph{Advances in Neural Information Processing
  Systems}}, Vol.~\bibinfo{volume}{35}. \bibinfo{pages}{13485--13498}.
\newblock
\urldef\tempurl%
\url{https://doi.org/10.52202/068431-0980}
\showDOI{\tempurl}


\bibitem[Li et~al\mbox{.}(2024a)]%
        {li2024spacetime}
\bibfield{author}{\bibinfo{person}{Zhan Li}, \bibinfo{person}{Zhang Chen},
  \bibinfo{person}{Zhong Li}, {and} \bibinfo{person}{Yi Xu}.}
  \bibinfo{year}{2024}\natexlab{a}.
\newblock \showarticletitle{{Spacetime Gaussian Feature Splatting for Real-Time
  Dynamic View Synthesis}}. In \bibinfo{booktitle}{\emph{Proceedings of the
  IEEE/CVF Conference on Computer Vision and Pattern Recognition (CVPR)}}.
  \bibinfo{pages}{8508--8520}.
\newblock


\bibitem[Li et~al\mbox{.}(2021)]%
        {li2021nsff}
\bibfield{author}{\bibinfo{person}{Zhengqi Li}, \bibinfo{person}{Simon
  Niklaus}, \bibinfo{person}{Noah Snavely}, {and} \bibinfo{person}{Oliver
  Wang}.} \bibinfo{year}{2021}\natexlab{}.
\newblock \showarticletitle{{Neural Scene Flow Fields for Space-Time View
  Synthesis of Dynamic Scenes}}. In \bibinfo{booktitle}{\emph{Proceedings of
  the IEEE/CVF Conference on Computer Vision and Pattern Recognition}}.
  \bibinfo{pages}{6498--6508}.
\newblock


\bibitem[Li et~al\mbox{.}(2024b)]%
        {li2024animatable}
\bibfield{author}{\bibinfo{person}{Zhe Li}, \bibinfo{person}{Zerong Zheng},
  \bibinfo{person}{Lizhen Wang}, {and} \bibinfo{person}{Yebin Liu}.}
  \bibinfo{year}{2024}\natexlab{b}.
\newblock \showarticletitle{{Animatable Gaussians: Learning Pose-dependent
  Gaussian Maps for High-fidelity Human Avatar Modeling}}. In
  \bibinfo{booktitle}{\emph{Proceedings of the IEEE/CVF Conference on Computer
  Vision and Pattern Recognition}}. \bibinfo{pages}{19711--19722}.
\newblock


\bibitem[Liang et~al\mbox{.}(2025)]%
        {liang2025gaufre}
\bibfield{author}{\bibinfo{person}{Yiqing Liang}, \bibinfo{person}{Numair
  Khan}, \bibinfo{person}{Zhengqin Li}, \bibinfo{person}{Thu~H. Nguyen-Phuoc},
  \bibinfo{person}{Douglas Lanman}, \bibinfo{person}{James Tompkin}, {and}
  \bibinfo{person}{Lei Xiao}.} \bibinfo{year}{2025}\natexlab{}.
\newblock \showarticletitle{{GauFRe: Gaussian Deformation Fields for Real-Time
  Dynamic Novel View Synthesis}}. In \bibinfo{booktitle}{\emph{Proceedings of
  the IEEE/CVF Winter Conference on Applications of Computer Vision (WACV)}}.
  IEEE, \bibinfo{pages}{2642--2652}.
\newblock


\bibitem[Liao et~al\mbox{.}(2026)]%
        {liao2026sharptimegs}
\bibfield{author}{\bibinfo{person}{Zhanfeng Liao}, \bibinfo{person}{Jiajun
  Zhang}, \bibinfo{person}{Hanzhang Tu}, \bibinfo{person}{Zhixi Wang},
  \bibinfo{person}{Yunqi Gao}, \bibinfo{person}{Hongwen Zhang}, {and}
  \bibinfo{person}{Yebin Liu}.} \bibinfo{year}{2026}\natexlab{}.
\newblock \showarticletitle{SharpTimeGS: Sharp and Stable Dynamic Gaussian
  Splatting via Lifespan Modulation}. In \bibinfo{booktitle}{\emph{Proceedings
  of the IEEE/CVF Conference on Computer Vision and Pattern Recognition
  (CVPR)}}. \bibinfo{pages}{11798--11807}.
\newblock


\bibitem[Lin et~al\mbox{.}(2026)]%
        {da3}
\bibfield{author}{\bibinfo{person}{Haotong Lin}, \bibinfo{person}{Sili Chen},
  \bibinfo{person}{Jun~Hao Liew}, \bibinfo{person}{Donny~Y. Chen},
  \bibinfo{person}{Zhenyu Li}, \bibinfo{person}{Yang Zhao},
  \bibinfo{person}{Sida Peng}, \bibinfo{person}{Hengkai Guo},
  \bibinfo{person}{Xiaowei Zhou}, \bibinfo{person}{Guang Shi},
  \bibinfo{person}{Jiashi Feng}, {and} \bibinfo{person}{Bingyi Kang}.}
  \bibinfo{year}{2026}\natexlab{}.
\newblock \showarticletitle{Depth Anything 3: Recovering the Visual Space from
  Any Views}. In \bibinfo{booktitle}{\emph{International Conference on Learning
  Representations (ICLR)}}, Vol.~\bibinfo{volume}{2026}.
  \bibinfo{pages}{141261--141285}.
\newblock
\urldef\tempurl%
\url{https://proceedings.iclr.cc/paper_files/paper/2026/hash/e4cd50120b6d7e8daff1749d6bbaa889-Abstract-Conference.html}
\showURL{%
\tempurl}


\bibitem[Lin et~al\mbox{.}(2022a)]%
        {lin2022efficient}
\bibfield{author}{\bibinfo{person}{Haotong Lin}, \bibinfo{person}{Sida Peng},
  \bibinfo{person}{Zhen Xu}, \bibinfo{person}{Yunzhi Yan},
  \bibinfo{person}{Qing Shuai}, \bibinfo{person}{Hujun Bao}, {and}
  \bibinfo{person}{Xiaowei Zhou}.} \bibinfo{year}{2022}\natexlab{a}.
\newblock \showarticletitle{{Efficient Neural Radiance Fields for Interactive
  Free-viewpoint Video}}. In \bibinfo{booktitle}{\emph{SIGGRAPH Asia 2022
  Conference Papers}}. Article \bibinfo{articleno}{39},
  \bibinfo{numpages}{9}~pages.
\newblock
\urldef\tempurl%
\url{https://doi.org/10.1145/3550469.3555376}
\showDOI{\tempurl}


\bibitem[Lin et~al\mbox{.}(2022b)]%
        {lin2021robust}
\bibfield{author}{\bibinfo{person}{Shanchuan Lin}, \bibinfo{person}{Linjie
  Yang}, \bibinfo{person}{Imran Saleemi}, {and} \bibinfo{person}{Soumyadip
  Sengupta}.} \bibinfo{year}{2022}\natexlab{b}.
\newblock \showarticletitle{Robust High-Resolution Video Matting With Temporal
  Guidance}. In \bibinfo{booktitle}{\emph{Proceedings of the IEEE/CVF Winter
  Conference on Applications of Computer Vision (WACV)}}.
  \bibinfo{pages}{238--247}.
\newblock


\bibitem[Liu et~al\mbox{.}(2025)]%
        {qingming2025modgs}
\bibfield{author}{\bibinfo{person}{Qingming Liu}, \bibinfo{person}{Yuan Liu},
  \bibinfo{person}{Jiepeng Wang}, \bibinfo{person}{Xianqiang Lyu},
  \bibinfo{person}{Peng Wang}, \bibinfo{person}{Wenping Wang}, {and}
  \bibinfo{person}{Junhui Hou}.} \bibinfo{year}{2025}\natexlab{}.
\newblock \showarticletitle{{MoDGS: Dynamic Gaussian Splatting from
  Casually-captured Monocular Videos with Depth Priors}}. In
  \bibinfo{booktitle}{\emph{International Conference on Learning
  Representations (ICLR)}}, Vol.~\bibinfo{volume}{2025}.
  \bibinfo{pages}{97048--97074}.
\newblock


\bibitem[Liu et~al\mbox{.}(2026)]%
        {liu2024dynamics}
\bibfield{author}{\bibinfo{person}{Zhening Liu}, \bibinfo{person}{Yingdong Hu},
  \bibinfo{person}{Xinjie Zhang}, \bibinfo{person}{Rui Song},
  \bibinfo{person}{Jiawei Shao}, \bibinfo{person}{Zehong Lin}, {and}
  \bibinfo{person}{Jun Zhang}.} \bibinfo{year}{2026}\natexlab{}.
\newblock \showarticletitle{{Dynamics-Aware Gaussian Splatting Streaming Toward
  Fast On-the-Fly 4D Reconstruction}}.
\newblock \bibinfo{journal}{\emph{IEEE Transactions on Visualization and
  Computer Graphics}} \bibinfo{volume}{32}, \bibinfo{number}{7}
  (\bibinfo{date}{July} \bibinfo{year}{2026}), \bibinfo{pages}{6810--6824}.
\newblock
\urldef\tempurl%
\url{https://doi.org/10.1109/TVCG.2026.3688730}
\showDOI{\tempurl}


\bibitem[Loper et~al\mbox{.}(2015)]%
        {loper2023smpl}
\bibfield{author}{\bibinfo{person}{Matthew Loper}, \bibinfo{person}{Naureen
  Mahmood}, \bibinfo{person}{Javier Romero}, \bibinfo{person}{Gerard
  Pons-Moll}, {and} \bibinfo{person}{Michael~J. Black}.}
  \bibinfo{year}{2015}\natexlab{}.
\newblock \showarticletitle{{SMPL}: A Skinned Multi-Person Linear Model}.
\newblock \bibinfo{journal}{\emph{ACM Transactions on Graphics (Proceedings of
  SIGGRAPH Asia)}} \bibinfo{volume}{34}, \bibinfo{number}{6}
  (\bibinfo{date}{October} \bibinfo{year}{2015}),
  \bibinfo{pages}{248:1--248:16}.
\newblock
\urldef\tempurl%
\url{https://doi.org/10.1145/2816795.2818013}
\showDOI{\tempurl}


\bibitem[Lu et~al\mbox{.}(2024)]%
        {lu20243d}
\bibfield{author}{\bibinfo{person}{Zhicheng Lu}, \bibinfo{person}{Xiang Guo},
  \bibinfo{person}{Le Hui}, \bibinfo{person}{Tianrui Chen},
  \bibinfo{person}{Min Yang}, \bibinfo{person}{Xiao Tang},
  \bibinfo{person}{Feng Zhu}, {and} \bibinfo{person}{Yuchao Dai}.}
  \bibinfo{year}{2024}\natexlab{}.
\newblock \showarticletitle{{3D Geometry-Aware Deformable Gaussian Splatting
  for Dynamic View Synthesis}}. In \bibinfo{booktitle}{\emph{Proceedings of the
  IEEE/CVF Conference on Computer Vision and Pattern Recognition (CVPR)}}.
  \bibinfo{pages}{8900--8910}.
\newblock


\bibitem[Luiten et~al\mbox{.}(2024)]%
        {luiten2024dynamic}
\bibfield{author}{\bibinfo{person}{Jonathon Luiten}, \bibinfo{person}{Georgios
  Kopanas}, \bibinfo{person}{Bastian Leibe}, {and} \bibinfo{person}{Deva
  Ramanan}.} \bibinfo{year}{2024}\natexlab{}.
\newblock \showarticletitle{Dynamic 3D Gaussians: Tracking by Persistent
  Dynamic View Synthesis}. In \bibinfo{booktitle}{\emph{International
  Conference on 3D Vision (3DV)}}. \bibinfo{pages}{800--809}.
\newblock
\urldef\tempurl%
\url{https://doi.org/10.1109/3DV62453.2024.00044}
\showDOI{\tempurl}


\bibitem[Maaz et~al\mbox{.}(2023)]%
        {edgenext}
\bibfield{author}{\bibinfo{person}{Muhammad Maaz}, \bibinfo{person}{Abdelrahman
  Shaker}, \bibinfo{person}{Hisham Cholakkal}, \bibinfo{person}{Salman Khan},
  \bibinfo{person}{Syed~Waqas Zamir}, \bibinfo{person}{Rao~Muhammad Anwer},
  {and} \bibinfo{person}{Fahad Shahbaz~Khan}.} \bibinfo{year}{2023}\natexlab{}.
\newblock \showarticletitle{{EdgeNeXt: Efficiently Amalgamated CNN-Transformer
  Architecture for Mobile Vision Applications}}. In
  \bibinfo{booktitle}{\emph{Computer Vision -- ECCV 2022 Workshops}}
  \emph{(\bibinfo{series}{Lecture Notes in Computer Science},
  Vol.~\bibinfo{volume}{13807})}. \bibinfo{publisher}{Springer Nature
  Switzerland}, \bibinfo{pages}{3--20}.
\newblock
\urldef\tempurl%
\url{https://doi.org/10.1007/978-3-031-25082-8_1}
\showDOI{\tempurl}


\bibitem[Mildenhall et~al\mbox{.}(2020)]%
        {mildenhall2020nerf}
\bibfield{author}{\bibinfo{person}{Ben Mildenhall}, \bibinfo{person}{Pratul~P.
  Srinivasan}, \bibinfo{person}{Matthew Tancik}, \bibinfo{person}{Jonathan~T.
  Barron}, \bibinfo{person}{Ravi Ramamoorthi}, {and} \bibinfo{person}{Ren Ng}.}
  \bibinfo{year}{2020}\natexlab{}.
\newblock \showarticletitle{NeRF: Representing Scenes as Neural Radiance Fields
  for View Synthesis}. In \bibinfo{booktitle}{\emph{Computer Vision -- ECCV
  2020}} \emph{(\bibinfo{series}{Lecture Notes in Computer Science},
  Vol.~\bibinfo{volume}{12346})}. \bibinfo{publisher}{Springer International
  Publishing}, \bibinfo{pages}{405--421}.
\newblock
\urldef\tempurl%
\url{https://doi.org/10.1007/978-3-030-58452-8_24}
\showDOI{\tempurl}


\bibitem[Newcombe et~al\mbox{.}(2015)]%
        {newcombe2015dynamicfusion}
\bibfield{author}{\bibinfo{person}{Richard~A Newcombe}, \bibinfo{person}{Dieter
  Fox}, {and} \bibinfo{person}{Steven~M Seitz}.}
  \bibinfo{year}{2015}\natexlab{}.
\newblock \showarticletitle{{DynamicFusion: Reconstruction and Tracking of
  Non-Rigid Scenes in Real-Time}}. In \bibinfo{booktitle}{\emph{Proceedings of
  the IEEE conference on computer vision and pattern recognition}}.
  \bibinfo{pages}{343--352}.
\newblock


\bibitem[Orts-Escolano et~al\mbox{.}(2016)]%
        {orts2016holoportation}
\bibfield{author}{\bibinfo{person}{Sergio Orts-Escolano},
  \bibinfo{person}{Christoph Rhemann}, \bibinfo{person}{Sean Fanello},
  \bibinfo{person}{Wayne Chang}, \bibinfo{person}{Adarsh Kowdle},
  \bibinfo{person}{Yury Degtyarev}, \bibinfo{person}{David Kim},
  \bibinfo{person}{Philip~L. Davidson}, \bibinfo{person}{Sameh Khamis},
  \bibinfo{person}{Mingsong Dou}, \bibinfo{person}{Vladimir Tankovich},
  \bibinfo{person}{Charles Loop}, \bibinfo{person}{Qin Cai},
  \bibinfo{person}{Philip~A. Chou}, \bibinfo{person}{Sarah Mennicken},
  \bibinfo{person}{Julien Valentin}, \bibinfo{person}{Vivek Pradeep},
  \bibinfo{person}{Shenlong Wang}, \bibinfo{person}{Sing~Bing Kang},
  \bibinfo{person}{Pushmeet Kohli}, \bibinfo{person}{Yuliya Lutchyn},
  \bibinfo{person}{Cem Keskin}, {and} \bibinfo{person}{Shahram Izadi}.}
  \bibinfo{year}{2016}\natexlab{}.
\newblock \showarticletitle{{Holoportation: virtual 3D teleportation in
  real-time}}. In \bibinfo{booktitle}{\emph{Proceedings of the 29th annual
  symposium on user interface software and technology}}.
  \bibinfo{pages}{741--754}.
\newblock
\urldef\tempurl%
\url{https://doi.org/10.1145/2984511.2984517}
\showDOI{\tempurl}


\bibitem[Qian et~al\mbox{.}(2024)]%
        {qian2024gaussianavatars}
\bibfield{author}{\bibinfo{person}{Shenhan Qian}, \bibinfo{person}{Tobias
  Kirschstein}, \bibinfo{person}{Liam Schoneveld}, \bibinfo{person}{Davide
  Davoli}, \bibinfo{person}{Simon Giebenhain}, {and} \bibinfo{person}{Matthias
  Nie{\ss}ner}.} \bibinfo{year}{2024}\natexlab{}.
\newblock \showarticletitle{{GaussianAvatars: Photorealistic Head Avatars with
  Rigged 3D Gaussians}}. In \bibinfo{booktitle}{\emph{Proceedings of the
  IEEE/CVF Conference on Computer Vision and Pattern Recognition (CVPR)}}.
  \bibinfo{pages}{20299--20309}.
\newblock


\bibitem[Ranftl et~al\mbox{.}(2021)]%
        {dpt}
\bibfield{author}{\bibinfo{person}{Ren{\'e} Ranftl}, \bibinfo{person}{Alexey
  Bochkovskiy}, {and} \bibinfo{person}{Vladlen Koltun}.}
  \bibinfo{year}{2021}\natexlab{}.
\newblock \showarticletitle{{Vision Transformers for Dense Prediction}}. In
  \bibinfo{booktitle}{\emph{Proceedings of the IEEE/CVF international
  conference on computer vision}}. \bibinfo{pages}{12179--12188}.
\newblock


\bibitem[Sch{\"o}nberger and Frahm(2016)]%
        {schonberger2016structure}
\bibfield{author}{\bibinfo{person}{Johannes~L. Sch{\"o}nberger} {and}
  \bibinfo{person}{Jan-Michael Frahm}.} \bibinfo{year}{2016}\natexlab{}.
\newblock \showarticletitle{{Structure-From-Motion Revisited}}. In
  \bibinfo{booktitle}{\emph{Proceedings of the IEEE Conference on Computer
  Vision and Pattern Recognition (CVPR)}}. \bibinfo{pages}{4104--4113}.
\newblock


\bibitem[Shao et~al\mbox{.}(2022)]%
        {shao2022doublefield}
\bibfield{author}{\bibinfo{person}{Ruizhi Shao}, \bibinfo{person}{Hongwen
  Zhang}, \bibinfo{person}{He Zhang}, \bibinfo{person}{Mingjia Chen},
  \bibinfo{person}{Yan-Pei Cao}, \bibinfo{person}{Tao Yu}, {and}
  \bibinfo{person}{Yebin Liu}.} \bibinfo{year}{2022}\natexlab{}.
\newblock \showarticletitle{{DoubleField: Bridging the Neural Surface and
  Radiance Fields for High-Fidelity Human Reconstruction and Rendering}}. In
  \bibinfo{booktitle}{\emph{Proceedings of the IEEE/CVF Conference on Computer
  Vision and Pattern Recognition}}. \bibinfo{pages}{15872--15882}.
\newblock


\bibitem[Shaw et~al\mbox{.}(2025)]%
        {shaw2024swings}
\bibfield{author}{\bibinfo{person}{Richard Shaw}, \bibinfo{person}{Michal
  Nazarczuk}, \bibinfo{person}{Jifei Song}, \bibinfo{person}{Arthur Moreau},
  \bibinfo{person}{Sibi Catley-Chandar}, \bibinfo{person}{Helisa Dhamo}, {and}
  \bibinfo{person}{Eduardo P{\'e}rez-Pellitero}.}
  \bibinfo{year}{2025}\natexlab{}.
\newblock \showarticletitle{{SWinGS: Sliding Windows for Dynamic 3D Gaussian
  Splatting}}. In \bibinfo{booktitle}{\emph{Computer Vision -- ECCV 2024}}
  \emph{(\bibinfo{series}{Lecture Notes in Computer Science},
  Vol.~\bibinfo{volume}{15113})}. \bibinfo{publisher}{Springer Nature
  Switzerland}, \bibinfo{pages}{37--54}.
\newblock
\urldef\tempurl%
\url{https://doi.org/10.1007/978-3-031-73001-6_3}
\showDOI{\tempurl}


\bibitem[Sim{\'e}oni et~al\mbox{.}(2026)]%
        {simeoni2025dinov3}
\bibfield{author}{\bibinfo{person}{Oriane Sim{\'e}oni}, \bibinfo{person}{Huy~V.
  Vo}, \bibinfo{person}{Maximilian Seitzer}, \bibinfo{person}{Federico
  Baldassarre}, \bibinfo{person}{Maxime Oquab}, \bibinfo{person}{Cijo Jose},
  \bibinfo{person}{Vasil Khalidov}, \bibinfo{person}{Marc Szafraniec},
  \bibinfo{person}{Seung~Eun Yi}, \bibinfo{person}{Michael Ramamonjisoa},
  \bibinfo{person}{Francisco Massa}, \bibinfo{person}{Daniel HAZIZA},
  \bibinfo{person}{Luca Wehrstedt}, \bibinfo{person}{Jianyuan Wang},
  \bibinfo{person}{Timoth{\'e}e Darcet}, \bibinfo{person}{Th{\'e}o Moutakanni},
  \bibinfo{person}{Leonel Sentana}, \bibinfo{person}{Claire Roberts},
  \bibinfo{person}{Andrea Vedaldi}, \bibinfo{person}{Jamie Tolan},
  \bibinfo{person}{John Brandt}, \bibinfo{person}{Camille Couprie},
  \bibinfo{person}{Julien Mairal}, \bibinfo{person}{Herve Jegou},
  \bibinfo{person}{Patrick Labatut}, {and} \bibinfo{person}{Piotr Bojanowski}.}
  \bibinfo{year}{2026}\natexlab{}.
\newblock \showarticletitle{{DINO}v3}.
\newblock \bibinfo{journal}{\emph{Transactions on Machine Learning Research}}
  (\bibinfo{year}{2026}).
\newblock
\showISSN{2835-8856}
\urldef\tempurl%
\url{https://openreview.net/forum?id=2NlGyqNjns}
\showURL{%
\tempurl}
\newblock
\shownote{Featured Certification}.


\bibitem[Slavcheva et~al\mbox{.}(2017)]%
        {slavcheva2017killingfusion}
\bibfield{author}{\bibinfo{person}{Miroslava Slavcheva},
  \bibinfo{person}{Maximilian Baust}, \bibinfo{person}{Daniel Cremers}, {and}
  \bibinfo{person}{Slobodan Ilic}.} \bibinfo{year}{2017}\natexlab{}.
\newblock \showarticletitle{{KillingFusion: Non-Rigid 3D Reconstruction Without
  Correspondences}}. In \bibinfo{booktitle}{\emph{Proceedings of the IEEE
  conference on computer vision and pattern recognition}}.
  \bibinfo{pages}{1386--1395}.
\newblock


\bibitem[Slavcheva et~al\mbox{.}(2018)]%
        {slavcheva2018sobolevfusion}
\bibfield{author}{\bibinfo{person}{Miroslava Slavcheva},
  \bibinfo{person}{Maximilian Baust}, {and} \bibinfo{person}{Slobodan Ilic}.}
  \bibinfo{year}{2018}\natexlab{}.
\newblock \showarticletitle{{SobolevFusion: 3D Reconstruction of Scenes
  Undergoing Free Non-Rigid Motion}}. In \bibinfo{booktitle}{\emph{Proceedings
  of the IEEE conference on computer vision and pattern recognition}}.
  \bibinfo{pages}{2646--2655}.
\newblock


\bibitem[Su et~al\mbox{.}(2024)]%
        {su2024roformer}
\bibfield{author}{\bibinfo{person}{Jianlin Su}, \bibinfo{person}{Murtadha
  Ahmed}, \bibinfo{person}{Yu Lu}, \bibinfo{person}{Shengfeng Pan},
  \bibinfo{person}{Wen Bo}, {and} \bibinfo{person}{Yunfeng Liu}.}
  \bibinfo{year}{2024}\natexlab{}.
\newblock \showarticletitle{{RoFormer: Enhanced transformer with Rotary
  Position Embedding}}.
\newblock \bibinfo{journal}{\emph{Neurocomputing}}  \bibinfo{volume}{568}
  (\bibinfo{year}{2024}), \bibinfo{pages}{127063}.
\newblock
\urldef\tempurl%
\url{https://doi.org/10.1016/j.neucom.2023.127063}
\showDOI{\tempurl}


\bibitem[Sun et~al\mbox{.}(2024)]%
        {sun20243dgstream}
\bibfield{author}{\bibinfo{person}{Jiakai Sun}, \bibinfo{person}{Han Jiao},
  \bibinfo{person}{Guangyuan Li}, \bibinfo{person}{Zhanjie Zhang},
  \bibinfo{person}{Lei Zhao}, {and} \bibinfo{person}{Wei Xing}.}
  \bibinfo{year}{2024}\natexlab{}.
\newblock \showarticletitle{{3DGStream: On-the-Fly Training of 3D Gaussians for
  Efficient Streaming of Photo-Realistic Free-Viewpoint Videos}}. In
  \bibinfo{booktitle}{\emph{Proceedings of the IEEE/CVF Conference on Computer
  Vision and Pattern Recognition}}. \bibinfo{pages}{20675--20685}.
\newblock


\bibitem[Touvron et~al\mbox{.}(2021)]%
        {touvron2021going}
\bibfield{author}{\bibinfo{person}{Hugo Touvron}, \bibinfo{person}{Matthieu
  Cord}, \bibinfo{person}{Alexandre Sablayrolles}, \bibinfo{person}{Gabriel
  Synnaeve}, {and} \bibinfo{person}{Herv{\'e} J{\'e}gou}.}
  \bibinfo{year}{2021}\natexlab{}.
\newblock \showarticletitle{{Going Deeper With Image Transformers}}. In
  \bibinfo{booktitle}{\emph{Proceedings of the IEEE/CVF international
  conference on computer vision}}. \bibinfo{pages}{32--42}.
\newblock


\bibitem[Tu et~al\mbox{.}(2024)]%
        {telealoha}
\bibfield{author}{\bibinfo{person}{Hanzhang Tu}, \bibinfo{person}{Ruizhi Shao},
  \bibinfo{person}{Xue Dong}, \bibinfo{person}{Shunyuan Zheng},
  \bibinfo{person}{Hao Zhang}, \bibinfo{person}{Lili Chen},
  \bibinfo{person}{Meili Wang}, \bibinfo{person}{Wenyu Li},
  \bibinfo{person}{Siyan Ma}, \bibinfo{person}{Shengping Zhang},
  \bibinfo{person}{Boyao Zhou}, {and} \bibinfo{person}{Yebin Liu}.}
  \bibinfo{year}{2024}\natexlab{}.
\newblock \showarticletitle{{Tele-Aloha: A Telepresence System with Low-budget
  and High-authenticity Using Sparse RGB Cameras}}. In
  \bibinfo{booktitle}{\emph{ACM SIGGRAPH 2024 Conference Papers}}. Article
  \bibinfo{articleno}{116}, \bibinfo{numpages}{12}~pages.
\newblock
\urldef\tempurl%
\url{https://doi.org/10.1145/3641519.3657491}
\showDOI{\tempurl}


\bibitem[{Twindom}(2026)]%
        {twindom}
\bibfield{author}{\bibinfo{person}{{Twindom}}.}
  \bibinfo{year}{2026}\natexlab{}.
\newblock \bibinfo{title}{Twindom}.
\newblock \bibinfo{howpublished}{Online}.
\newblock
\urldef\tempurl%
\url{https://web.twindom.com/}
\showURL{%
\tempurl}
\newblock
\shownote{Accessed August 25, 2026}.


\bibitem[Wang et~al\mbox{.}(2025a)]%
        {vggt}
\bibfield{author}{\bibinfo{person}{Jianyuan Wang}, \bibinfo{person}{Minghao
  Chen}, \bibinfo{person}{Nikita Karaev}, \bibinfo{person}{Andrea Vedaldi},
  \bibinfo{person}{Christian Rupprecht}, {and} \bibinfo{person}{David
  Novotny}.} \bibinfo{year}{2025}\natexlab{a}.
\newblock \showarticletitle{{VGGT: Visual Geometry Grounded Transformer}}. In
  \bibinfo{booktitle}{\emph{Proceedings of the Computer Vision and Pattern
  Recognition Conference}}. \bibinfo{pages}{5294--5306}.
\newblock


\bibitem[Wang et~al\mbox{.}(2023b)]%
        {wang2023neural2}
\bibfield{author}{\bibinfo{person}{Liao Wang}, \bibinfo{person}{Qiang Hu},
  \bibinfo{person}{Qihan He}, \bibinfo{person}{Ziyu Wang},
  \bibinfo{person}{Jingyi Yu}, \bibinfo{person}{Tinne Tuytelaars},
  \bibinfo{person}{Lan Xu}, {and} \bibinfo{person}{Minye Wu}.}
  \bibinfo{year}{2023}\natexlab{b}.
\newblock \showarticletitle{{Neural Residual Radiance Fields for Streamably
  Free-Viewpoint Videos}}. In \bibinfo{booktitle}{\emph{Proceedings of the
  IEEE/CVF Conference on Computer Vision and Pattern Recognition}}.
  \bibinfo{pages}{76--87}.
\newblock


\bibitem[Wang et~al\mbox{.}(2024b)]%
        {wang2024videorf}
\bibfield{author}{\bibinfo{person}{Liao Wang}, \bibinfo{person}{Kaixin Yao},
  \bibinfo{person}{Chengcheng Guo}, \bibinfo{person}{Zhirui Zhang},
  \bibinfo{person}{Qiang Hu}, \bibinfo{person}{Jingyi Yu}, \bibinfo{person}{Lan
  Xu}, {and} \bibinfo{person}{Minye Wu}.} \bibinfo{year}{2024}\natexlab{b}.
\newblock \showarticletitle{{VideoRF: Rendering Dynamic Radiance Fields as 2D
  Feature Video Streams}}. In \bibinfo{booktitle}{\emph{Proceedings of the
  IEEE/CVF Conference on Computer Vision and Pattern Recognition}}.
  \bibinfo{pages}{470--481}.
\newblock


\bibitem[Wang et~al\mbox{.}(2024c)]%
        {v3}
\bibfield{author}{\bibinfo{person}{Penghao Wang}, \bibinfo{person}{Zhirui
  Zhang}, \bibinfo{person}{Liao Wang}, \bibinfo{person}{Kaixin Yao},
  \bibinfo{person}{Siyuan Xie}, \bibinfo{person}{Jingyi Yu},
  \bibinfo{person}{Minye Wu}, {and} \bibinfo{person}{Lan Xu}.}
  \bibinfo{year}{2024}\natexlab{c}.
\newblock \showarticletitle{{$V^3$}: Viewing Volumetric Videos on Mobiles via
  Streamable 2D Dynamic Gaussians}.
\newblock \bibinfo{journal}{\emph{ACM Transactions on Graphics (TOG)}}
  \bibinfo{volume}{43}, \bibinfo{number}{6}, Article \bibinfo{articleno}{187}
  (\bibinfo{year}{2024}), \bibinfo{numpages}{13}~pages.
\newblock
\urldef\tempurl%
\url{https://doi.org/10.1145/3687935}
\showDOI{\tempurl}


\bibitem[Wang et~al\mbox{.}(2021)]%
        {wang2021ibrnet}
\bibfield{author}{\bibinfo{person}{Qianqian Wang}, \bibinfo{person}{Zhicheng
  Wang}, \bibinfo{person}{Kyle Genova}, \bibinfo{person}{Pratul~P. Srinivasan},
  \bibinfo{person}{Howard Zhou}, \bibinfo{person}{Jonathan~T. Barron},
  \bibinfo{person}{Ricardo Martin-Brualla}, \bibinfo{person}{Noah Snavely},
  {and} \bibinfo{person}{Thomas Funkhouser}.} \bibinfo{year}{2021}\natexlab{}.
\newblock \showarticletitle{{IBRNet: Learning Multi-View Image-Based
  Rendering}}. In \bibinfo{booktitle}{\emph{Proceedings of the IEEE/CVF
  Conference on Computer Vision and Pattern Recognition}}.
  \bibinfo{pages}{4690--4699}.
\newblock


\bibitem[Wang et~al\mbox{.}(2024a)]%
        {wang2024dust3r}
\bibfield{author}{\bibinfo{person}{Shuzhe Wang}, \bibinfo{person}{Vincent
  Leroy}, \bibinfo{person}{Yohann Cabon}, \bibinfo{person}{Boris Chidlovskii},
  {and} \bibinfo{person}{J{\'e}r{\^o}me Revaud}.}
  \bibinfo{year}{2024}\natexlab{a}.
\newblock \showarticletitle{{DUSt3R: Geometric 3D Vision Made Easy}}. In
  \bibinfo{booktitle}{\emph{Proceedings of the IEEE/CVF conference on computer
  vision and pattern recognition}}. \bibinfo{pages}{20697--20709}.
\newblock


\bibitem[Wang et~al\mbox{.}(2023a)]%
        {neus2}
\bibfield{author}{\bibinfo{person}{Yiming Wang}, \bibinfo{person}{Qin Han},
  \bibinfo{person}{Marc Habermann}, \bibinfo{person}{Kostas Daniilidis},
  \bibinfo{person}{Christian Theobalt}, {and} \bibinfo{person}{Lingjie Liu}.}
  \bibinfo{year}{2023}\natexlab{a}.
\newblock \showarticletitle{NeuS2: Fast Learning of Neural Implicit Surfaces
  for Multi-view Reconstruction}. In \bibinfo{booktitle}{\emph{Proceedings of
  the IEEE/CVF International Conference on Computer Vision (ICCV)}}.
  \bibinfo{pages}{3295--3306}.
\newblock


\bibitem[Wang et~al\mbox{.}(2025b)]%
        {wang2025freetimegs}
\bibfield{author}{\bibinfo{person}{Yifan Wang}, \bibinfo{person}{Peishan Yang},
  \bibinfo{person}{Zhen Xu}, \bibinfo{person}{Jiaming Sun},
  \bibinfo{person}{Zhanhua Zhang}, \bibinfo{person}{Yong Chen},
  \bibinfo{person}{Hujun Bao}, \bibinfo{person}{Sida Peng}, {and}
  \bibinfo{person}{Xiaowei Zhou}.} \bibinfo{year}{2025}\natexlab{b}.
\newblock \showarticletitle{FreeTimeGS: Free Gaussian Primitives at Anytime
  Anywhere for Dynamic Scene Reconstruction}. In
  \bibinfo{booktitle}{\emph{Proceedings of the IEEE/CVF Conference on Computer
  Vision and Pattern Recognition (CVPR)}}. \bibinfo{pages}{21750--21760}.
\newblock


\bibitem[Wang et~al\mbox{.}(2026)]%
        {pi3}
\bibfield{author}{\bibinfo{person}{Yifan Wang}, \bibinfo{person}{Jianjun Zhou},
  \bibinfo{person}{Haoyi Zhu}, \bibinfo{person}{Wenzheng Chang},
  \bibinfo{person}{Yang Zhou}, \bibinfo{person}{Zizun Li},
  \bibinfo{person}{Junyi Chen}, \bibinfo{person}{Jiangmiao Pang},
  \bibinfo{person}{Chunhua Shen}, {and} \bibinfo{person}{Tong He}.}
  \bibinfo{year}{2026}\natexlab{}.
\newblock \showarticletitle{{$\pi^3$}: Permutation-Equivariant Visual Geometry
  Learning}. In \bibinfo{booktitle}{\emph{International Conference on Learning
  Representations (ICLR)}}, Vol.~\bibinfo{volume}{2026}.
  \bibinfo{pages}{10481--10497}.
\newblock
\urldef\tempurl%
\url{https://proceedings.iclr.cc/paper_files/paper/2026/hash/11a09e0aaa74867c6b0719c639fc09f8-Abstract-Conference.html}
\showURL{%
\tempurl}


\bibitem[Wang et~al\mbox{.}(2004)]%
        {wang2004image}
\bibfield{author}{\bibinfo{person}{Zhou Wang}, \bibinfo{person}{Alan~C Bovik},
  \bibinfo{person}{Hamid~R Sheikh}, {and} \bibinfo{person}{Eero~P Simoncelli}.}
  \bibinfo{year}{2004}\natexlab{}.
\newblock \showarticletitle{Image quality assessment: from error visibility to
  structural similarity}.
\newblock \bibinfo{journal}{\emph{IEEE Transactions on Image Processing}}
  \bibinfo{volume}{13}, \bibinfo{number}{4} (\bibinfo{year}{2004}),
  \bibinfo{pages}{600--612}.
\newblock


\bibitem[Wu et~al\mbox{.}(2024)]%
        {wu2024tetrirf}
\bibfield{author}{\bibinfo{person}{Minye Wu}, \bibinfo{person}{Zehao Wang},
  \bibinfo{person}{Georgios Kouros}, {and} \bibinfo{person}{Tinne Tuytelaars}.}
  \bibinfo{year}{2024}\natexlab{}.
\newblock \showarticletitle{{TeTriRF: Temporal Tri-Plane Radiance Fields for
  Efficient Free-Viewpoint Video}}. In \bibinfo{booktitle}{\emph{Proceedings of
  the IEEE/CVF conference on computer vision and pattern recognition}}.
  \bibinfo{pages}{6487--6496}.
\newblock


\bibitem[Xiang et~al\mbox{.}(2023)]%
        {xiang2023drivable}
\bibfield{author}{\bibinfo{person}{Donglai Xiang}, \bibinfo{person}{Fabian
  Prada}, \bibinfo{person}{Zhe Cao}, \bibinfo{person}{Kaiwen Guo},
  \bibinfo{person}{Chenglei Wu}, \bibinfo{person}{Jessica Hodgins}, {and}
  \bibinfo{person}{Timur Bagautdinov}.} \bibinfo{year}{2023}\natexlab{}.
\newblock \showarticletitle{{Drivable Avatar Clothing: Faithful Full-Body
  Telepresence with Dynamic Clothing Driven by Sparse RGB-D Input}}. In
  \bibinfo{booktitle}{\emph{SIGGRAPH Asia 2023 Conference Papers}}. Article
  \bibinfo{articleno}{24}, \bibinfo{numpages}{11}~pages.
\newblock
\urldef\tempurl%
\url{https://doi.org/10.1145/3610548.3618136}
\showDOI{\tempurl}


\bibitem[Xiao et~al\mbox{.}(2025)]%
        {spatialtrackerv2}
\bibfield{author}{\bibinfo{person}{Yuxi Xiao}, \bibinfo{person}{Jianyuan Wang},
  \bibinfo{person}{Nan Xue}, \bibinfo{person}{Nikita Karaev},
  \bibinfo{person}{Yuri Makarov}, \bibinfo{person}{Bingyi Kang},
  \bibinfo{person}{Xing Zhu}, \bibinfo{person}{Hujun Bao},
  \bibinfo{person}{Yujun Shen}, {and} \bibinfo{person}{Xiaowei Zhou}.}
  \bibinfo{year}{2025}\natexlab{}.
\newblock \showarticletitle{SpatialTrackerV2: Advancing 3D Point Tracking with
  Explicit Camera Motion}. In \bibinfo{booktitle}{\emph{Proceedings of the
  IEEE/CVF International Conference on Computer Vision (ICCV)}}.
  \bibinfo{pages}{6726--6737}.
\newblock


\bibitem[Xu et~al\mbox{.}(2025)]%
        {xu2025depthsplat}
\bibfield{author}{\bibinfo{person}{Haofei Xu}, \bibinfo{person}{Songyou Peng},
  \bibinfo{person}{Fangjinhua Wang}, \bibinfo{person}{Hermann Blum},
  \bibinfo{person}{Daniel Barath}, \bibinfo{person}{Andreas Geiger}, {and}
  \bibinfo{person}{Marc Pollefeys}.} \bibinfo{year}{2025}\natexlab{}.
\newblock \showarticletitle{{DepthSplat: Connecting Gaussian Splatting and
  Depth}}. In \bibinfo{booktitle}{\emph{Proceedings of the Computer Vision and
  Pattern Recognition Conference}}. \bibinfo{pages}{16453--16463}.
\newblock


\bibitem[Xu et~al\mbox{.}(2024a)]%
        {xu2024grid4d}
\bibfield{author}{\bibinfo{person}{Jiawei Xu}, \bibinfo{person}{Zexin Fan},
  \bibinfo{person}{Jian Yang}, {and} \bibinfo{person}{Jin Xie}.}
  \bibinfo{year}{2024}\natexlab{a}.
\newblock \showarticletitle{Grid4D: 4D Decomposed Hash Encoding for
  High-Fidelity Dynamic Gaussian Splatting}. In
  \bibinfo{booktitle}{\emph{Advances in Neural Information Processing
  Systems}}, Vol.~\bibinfo{volume}{37}. \bibinfo{pages}{123787--123811}.
\newblock
\urldef\tempurl%
\url{https://doi.org/10.52202/079017-3934}
\showDOI{\tempurl}


\bibitem[Xu et~al\mbox{.}(2023)]%
        {xu2023avatarmav}
\bibfield{author}{\bibinfo{person}{Yuelang Xu}, \bibinfo{person}{Lizhen Wang},
  \bibinfo{person}{Xiaochen Zhao}, \bibinfo{person}{Hongwen Zhang}, {and}
  \bibinfo{person}{Yebin Liu}.} \bibinfo{year}{2023}\natexlab{}.
\newblock \showarticletitle{AvatarMAV: Fast 3D Head Avatar Reconstruction Using
  Motion-Aware Neural Voxels}. In \bibinfo{booktitle}{\emph{ACM SIGGRAPH 2023
  Conference Proceedings}}. \bibinfo{publisher}{Association for Computing
  Machinery}, \bibinfo{address}{New York, NY, USA}, Article
  \bibinfo{articleno}{47}, \bibinfo{numpages}{10}~pages.
\newblock
\urldef\tempurl%
\url{https://doi.org/10.1145/3588432.3591567}
\showDOI{\tempurl}


\bibitem[Xu et~al\mbox{.}(2024b)]%
        {4k4d}
\bibfield{author}{\bibinfo{person}{Zhen Xu}, \bibinfo{person}{Sida Peng},
  \bibinfo{person}{Haotong Lin}, \bibinfo{person}{Guangzhao He},
  \bibinfo{person}{Jiaming Sun}, \bibinfo{person}{Yujun Shen},
  \bibinfo{person}{Hujun Bao}, {and} \bibinfo{person}{Xiaowei Zhou}.}
  \bibinfo{year}{2024}\natexlab{b}.
\newblock \showarticletitle{{4K4D: Real-Time 4D View Synthesis at 4K
  Resolution}}. In \bibinfo{booktitle}{\emph{Proceedings of the IEEE/CVF
  Conference on Computer Vision and Pattern Recognition (CVPR)}}.
  \bibinfo{pages}{20029--20040}.
\newblock


\bibitem[Xu et~al\mbox{.}(2024c)]%
        {xu2024representing}
\bibfield{author}{\bibinfo{person}{Zhen Xu}, \bibinfo{person}{Yinghao Xu},
  \bibinfo{person}{Zhiyuan Yu}, \bibinfo{person}{Sida Peng},
  \bibinfo{person}{Jiaming Sun}, \bibinfo{person}{Hujun Bao}, {and}
  \bibinfo{person}{Xiaowei Zhou}.} \bibinfo{year}{2024}\natexlab{c}.
\newblock \showarticletitle{{Representing Long Volumetric Video with Temporal
  Gaussian Hierarchy}}.
\newblock \bibinfo{journal}{\emph{ACM Transactions on Graphics}}
  \bibinfo{volume}{43}, \bibinfo{number}{6}, Article \bibinfo{articleno}{171}
  (\bibinfo{year}{2024}), \bibinfo{numpages}{18}~pages.
\newblock
\urldef\tempurl%
\url{https://doi.org/10.1145/3687919}
\showDOI{\tempurl}


\bibitem[Yan et~al\mbox{.}(2025)]%
        {yan2025instant}
\bibfield{author}{\bibinfo{person}{Jinbo Yan}, \bibinfo{person}{Rui Peng},
  \bibinfo{person}{Zhiyan Wang}, \bibinfo{person}{Luyang Tang},
  \bibinfo{person}{Jiayu Yang}, \bibinfo{person}{Jie Liang},
  \bibinfo{person}{Jiahao Wu}, {and} \bibinfo{person}{Ronggang Wang}.}
  \bibinfo{year}{2025}\natexlab{}.
\newblock \showarticletitle{{Instant Gaussian Stream: Fast and Generalizable
  Streaming of Dynamic Scene Reconstruction via Gaussian Splatting}}. In
  \bibinfo{booktitle}{\emph{Proceedings of the Computer Vision and Pattern
  Recognition Conference}}. \bibinfo{pages}{16520--16531}.
\newblock


\bibitem[Yang et~al\mbox{.}(2026)]%
        {yang2026matanyone2}
\bibfield{author}{\bibinfo{person}{Peiqing Yang}, \bibinfo{person}{Shangchen
  Zhou}, \bibinfo{person}{Kai Hao}, {and} \bibinfo{person}{Qingyi Tao}.}
  \bibinfo{year}{2026}\natexlab{}.
\newblock \showarticletitle{MatAnyone 2: Scaling Video Matting via a Learned
  Quality Evaluator}. In \bibinfo{booktitle}{\emph{Proceedings of the IEEE/CVF
  Conference on Computer Vision and Pattern Recognition (CVPR)}}.
  \bibinfo{pages}{37476--37485}.
\newblock


\bibitem[Yang et~al\mbox{.}(2024a)]%
        {yang2024deformable}
\bibfield{author}{\bibinfo{person}{Ziyi Yang}, \bibinfo{person}{Xinyu Gao},
  \bibinfo{person}{Wen Zhou}, \bibinfo{person}{Shaohui Jiao},
  \bibinfo{person}{Yuqing Zhang}, {and} \bibinfo{person}{Xiaogang Jin}.}
  \bibinfo{year}{2024}\natexlab{a}.
\newblock \showarticletitle{Deformable 3D Gaussians for High-Fidelity Monocular
  Dynamic Scene Reconstruction}. In \bibinfo{booktitle}{\emph{Proceedings of
  the IEEE/CVF Conference on Computer Vision and Pattern Recognition (CVPR)}}.
  \bibinfo{pages}{20331--20341}.
\newblock


\bibitem[Yang et~al\mbox{.}(2024b)]%
        {yang2024real}
\bibfield{author}{\bibinfo{person}{Zeyu Yang}, \bibinfo{person}{Hongye Yang},
  \bibinfo{person}{Zijie Pan}, {and} \bibinfo{person}{Li Zhang}.}
  \bibinfo{year}{2024}\natexlab{b}.
\newblock \showarticletitle{{Real-time Photorealistic Dynamic Scene
  Representation and Rendering with 4D Gaussian Splatting}}. In
  \bibinfo{booktitle}{\emph{The Twelfth International Conference on Learning
  Representations}}, Vol.~\bibinfo{volume}{2024}. \bibinfo{pages}{9142--9159}.
\newblock
\urldef\tempurl%
\url{https://openreview.net/forum?id=WhgB5sispV}
\showURL{%
\tempurl}


\bibitem[Ye et~al\mbox{.}(2025)]%
        {noposplat}
\bibfield{author}{\bibinfo{person}{Botao Ye}, \bibinfo{person}{Sifei Liu},
  \bibinfo{person}{Haofei Xu}, \bibinfo{person}{Xueting Li},
  \bibinfo{person}{Marc Pollefeys}, \bibinfo{person}{Ming-Hsuan Yang}, {and}
  \bibinfo{person}{Songyou Peng}.} \bibinfo{year}{2025}\natexlab{}.
\newblock \showarticletitle{No Pose, No Problem: Surprisingly Simple 3D
  Gaussian Splats from Sparse Unposed Images}. In \bibinfo{booktitle}{\emph{The
  Thirteenth International Conference on Learning Representations}},
  Vol.~\bibinfo{volume}{2025}. \bibinfo{pages}{54009--54033}.
\newblock


\bibitem[Yu et~al\mbox{.}(2017)]%
        {yu2017bodyfusion}
\bibfield{author}{\bibinfo{person}{Tao Yu}, \bibinfo{person}{Kaiwen Guo},
  \bibinfo{person}{Feng Xu}, \bibinfo{person}{Yuan Dong},
  \bibinfo{person}{Zhaoqi Su}, \bibinfo{person}{Jianhui Zhao},
  \bibinfo{person}{Jianguo Li}, \bibinfo{person}{Qionghai Dai}, {and}
  \bibinfo{person}{Yebin Liu}.} \bibinfo{year}{2017}\natexlab{}.
\newblock \showarticletitle{{BodyFusion: Real-Time Capture of Human Motion and
  Surface Geometry Using a Single Depth Camera}}. In
  \bibinfo{booktitle}{\emph{Proceedings of the IEEE international conference on
  computer vision}}. \bibinfo{pages}{910--919}.
\newblock


\bibitem[Yu et~al\mbox{.}(2021)]%
        {yu2021function4d}
\bibfield{author}{\bibinfo{person}{Tao Yu}, \bibinfo{person}{Zerong Zheng},
  \bibinfo{person}{Kaiwen Guo}, \bibinfo{person}{Pengpeng Liu},
  \bibinfo{person}{Qionghai Dai}, {and} \bibinfo{person}{Yebin Liu}.}
  \bibinfo{year}{2021}\natexlab{}.
\newblock \showarticletitle{{Function4D: Real-Time Human Volumetric Capture
  From Very Sparse Consumer RGBD Sensors}}. In
  \bibinfo{booktitle}{\emph{Proceedings of the IEEE/CVF conference on computer
  vision and pattern recognition}}. \bibinfo{pages}{5746--5756}.
\newblock


\bibitem[Yu et~al\mbox{.}(2018)]%
        {yu2018doublefusion}
\bibfield{author}{\bibinfo{person}{Tao Yu}, \bibinfo{person}{Zerong Zheng},
  \bibinfo{person}{Kaiwen Guo}, \bibinfo{person}{Jianhui Zhao},
  \bibinfo{person}{Qionghai Dai}, \bibinfo{person}{Hao Li},
  \bibinfo{person}{Gerard Pons-Moll}, {and} \bibinfo{person}{Yebin Liu}.}
  \bibinfo{year}{2018}\natexlab{}.
\newblock \showarticletitle{{DoubleFusion: Real-Time Capture of Human
  Performances With Inner Body Shapes From a Single Depth Sensor}}. In
  \bibinfo{booktitle}{\emph{Proceedings of the IEEE conference on computer
  vision and pattern recognition}}. \bibinfo{pages}{7287--7296}.
\newblock


\bibitem[Zhang et~al\mbox{.}(2021)]%
        {zhang2021stnerf}
\bibfield{author}{\bibinfo{person}{Jiakai Zhang}, \bibinfo{person}{Xinhang
  Liu}, \bibinfo{person}{Xinyi Ye}, \bibinfo{person}{Fuqiang Zhao},
  \bibinfo{person}{Yanshun Zhang}, \bibinfo{person}{Minye Wu},
  \bibinfo{person}{Yingliang Zhang}, \bibinfo{person}{Lan Xu}, {and}
  \bibinfo{person}{Jingyi Yu}.} \bibinfo{year}{2021}\natexlab{}.
\newblock \showarticletitle{{Editable Free-viewpoint Video Using a Layered
  Neural Representation}}.
\newblock \bibinfo{journal}{\emph{ACM Transactions on Graphics}}
  \bibinfo{volume}{40}, \bibinfo{number}{4}, Article \bibinfo{articleno}{149}
  (\bibinfo{year}{2021}), \bibinfo{numpages}{18}~pages.
\newblock
\urldef\tempurl%
\url{https://doi.org/10.1145/3450626.3459756}
\showDOI{\tempurl}


\bibitem[Zhang et~al\mbox{.}(2018)]%
        {zhang2018the}
\bibfield{author}{\bibinfo{person}{Richard Zhang}, \bibinfo{person}{Phillip
  Isola}, \bibinfo{person}{Alexei~A. Efros}, \bibinfo{person}{Eli Shechtman},
  {and} \bibinfo{person}{Oliver Wang}.} \bibinfo{year}{2018}\natexlab{}.
\newblock \showarticletitle{The Unreasonable Effectiveness of Deep Features as
  a Perceptual Metric}. In \bibinfo{booktitle}{\emph{Proceedings of the IEEE
  Conference on Computer Vision and Pattern Recognition (CVPR)}}.
  \bibinfo{pages}{586--595}.
\newblock


\bibitem[Zhao et~al\mbox{.}(2022)]%
        {zhao2022humannerf}
\bibfield{author}{\bibinfo{person}{Fuqiang Zhao}, \bibinfo{person}{Wei Yang},
  \bibinfo{person}{Jiakai Zhang}, \bibinfo{person}{Pei Lin},
  \bibinfo{person}{Yingliang Zhang}, \bibinfo{person}{Jingyi Yu}, {and}
  \bibinfo{person}{Lan Xu}.} \bibinfo{year}{2022}\natexlab{}.
\newblock \showarticletitle{{HumanNeRF: Efficiently Generated Human Radiance
  Field From Sparse Inputs}}. In \bibinfo{booktitle}{\emph{Proceedings of the
  IEEE/CVF Conference on Computer Vision and Pattern Recognition}}.
  \bibinfo{pages}{7743--7753}.
\newblock


\bibitem[Zheng et~al\mbox{.}(2024)]%
        {gps}
\bibfield{author}{\bibinfo{person}{Shunyuan Zheng}, \bibinfo{person}{Boyao
  Zhou}, \bibinfo{person}{Ruizhi Shao}, \bibinfo{person}{Boning Liu},
  \bibinfo{person}{Shengping Zhang}, \bibinfo{person}{Liqiang Nie}, {and}
  \bibinfo{person}{Yebin Liu}.} \bibinfo{year}{2024}\natexlab{}.
\newblock \showarticletitle{{GPS-Gaussian: Generalizable Pixel-wise 3D Gaussian
  Splatting for Real-time Human Novel View Synthesis}}. In
  \bibinfo{booktitle}{\emph{Proceedings of the IEEE/CVF conference on computer
  vision and pattern recognition}}. \bibinfo{pages}{19680--19690}.
\newblock


\bibitem[Zhou et~al\mbox{.}(2025)]%
        {zhou2024gps}
\bibfield{author}{\bibinfo{person}{Boyao Zhou}, \bibinfo{person}{Shunyuan
  Zheng}, \bibinfo{person}{Hanzhang Tu}, \bibinfo{person}{Ruizhi Shao},
  \bibinfo{person}{Boning Liu}, \bibinfo{person}{Shengping Zhang},
  \bibinfo{person}{Liqiang Nie}, {and} \bibinfo{person}{Yebin Liu}.}
  \bibinfo{year}{2025}\natexlab{}.
\newblock \showarticletitle{GPS-Gaussian+: Generalizable Pixel-wise 3D Gaussian
  Splatting for Real-Time Human-Scene Rendering from Sparse Views}.
\newblock \bibinfo{journal}{\emph{IEEE Transactions on Pattern Analysis and
  Machine Intelligence}} (\bibinfo{year}{2025}), \bibinfo{pages}{1--16}.
\newblock
\urldef\tempurl%
\url{https://doi.org/10.1109/TPAMI.2025.3561248}
\showDOI{\tempurl}


\bibitem[Zhou et~al\mbox{.}(2023)]%
        {zhou2023live4d}
\bibfield{author}{\bibinfo{person}{Yifeng Zhou}, \bibinfo{person}{Shuheng
  Wang}, \bibinfo{person}{Wenfa Li}, \bibinfo{person}{Chao Zhang},
  \bibinfo{person}{Li Rao}, \bibinfo{person}{Pu Cheng}, \bibinfo{person}{Yi
  Xu}, \bibinfo{person}{Jinle Ke}, \bibinfo{person}{Wenduo Feng},
  \bibinfo{person}{Wen Zhou}, \bibinfo{person}{Hao Xu}, \bibinfo{person}{Yukang
  Gao}, \bibinfo{person}{Yang Ding}, \bibinfo{person}{Weixuan Tang}, {and}
  \bibinfo{person}{Shaohui Jiao}.} \bibinfo{year}{2023}\natexlab{}.
\newblock \showarticletitle{{Live4D: A Real-time Capture System for Streamable
  Volumetric Video}}. In \bibinfo{booktitle}{\emph{SIGGRAPH Asia 2023 Technical
  Communications}}. \bibinfo{publisher}{Association for Computing Machinery},
  \bibinfo{address}{New York, NY, USA}, Article \bibinfo{articleno}{23},
  \bibinfo{numpages}{4}~pages.
\newblock
\urldef\tempurl%
\url{https://doi.org/10.1145/3610543.3626178}
\showDOI{\tempurl}


\bibitem[Zhu et~al\mbox{.}(2024)]%
        {zhu2024motiongs}
\bibfield{author}{\bibinfo{person}{Ruijie Zhu}, \bibinfo{person}{Yanzhe Liang},
  \bibinfo{person}{Hanzhi Chang}, \bibinfo{person}{Jiacheng Deng},
  \bibinfo{person}{Jiahao Lu}, \bibinfo{person}{Wenfei Yang},
  \bibinfo{person}{Tianzhu Zhang}, {and} \bibinfo{person}{Yongdong Zhang}.}
  \bibinfo{year}{2024}\natexlab{}.
\newblock \showarticletitle{{MotionGS: Exploring Explicit Motion Guidance for
  Deformable 3D Gaussian Splatting}}. In \bibinfo{booktitle}{\emph{Advances in
  Neural Information Processing Systems}}, Vol.~\bibinfo{volume}{37}.
  \bibinfo{pages}{101790--101817}.
\newblock


\end{thebibliography}

\end{document}